\documentclass{article}

\usepackage{arxiv}

\usepackage[utf8]{inputenc} 
\usepackage[T1]{fontenc}    
\usepackage{hyperref}       
\usepackage{url}            
\usepackage{booktabs}       
\usepackage{amsfonts}       
\usepackage{nicefrac}       
\usepackage{microtype}      
\usepackage{amsmath}
\usepackage{amssymb}
\usepackage{subcaption}
\usepackage{graphicx}
\usepackage{doi}

\usepackage[backend=bibtex]{biblatex}
\usepackage{placeins}
\title{Reducing Redundancy in Retrieval-Augmented Generation through Chunk Filtering}

\author{
Daria Berdyugina$^{1}$ \quad
Anaëlle Cohen$^{1}$ \quad
Yohann Rioual$^{1}$ \quad \\
$^{1}$AI Lab -- VO2 Group
}

\date{}

\hypersetup{
pdftitle={Chunking},
pdfauthor={Berdyugina D., Cohen A., Rioual Y.},
pdfkeywords={RAG, NLP, Information Retrieval, Chunking},
}

\begin{document}

\maketitle

\begin{abstract}
Standard Retrieval-Augmented Generation (RAG) chunking methods often create excessive redundancy, increasing storage costs and slowing retrieval. This study explores chunk filtering strategies, such as semantic, topic-based, and named-entity-based methods in order to reduce the indexed corpus while preserving retrieval quality. Experiments are conducted on multiple corpora. Retrieval performance is evaluated using a token-based framework based on precision, recall, and intersection-over-union metrics. Results indicate that entity-based filtering can reduce vector index size by approximately 25\% to 36\% while maintaining high retrieval quality close to the baseline. These findings suggest that redundancy introduced during chunking can be effectively reduced through lightweight filtering, improving the efficiency of retrieval-oriented components in RAG pipelines.
\end{abstract}


\section{Introduction}
\label{sec:intro}

\subsection{Context}

Retrieval-Augmented Generation (RAG) is used as an effective method for improving the reliability and factual grounding of large language models (LLMs). In a RAG architecture, a generative language model is paired with an information retrieval system that searches a document corpus for passages relevant to a user query. These passages are then passed to the language model as contextual information in order to allow generating responses based on the external knowledge rather than rely on the external parameters. 

This architecture offers several advantages. It helps reduce hallucinations, allows the integration of domain-specific or recently updated knowledge without without a necessity to retrain a model, and improves the traceability of generated answers. As a result, RAG systems are used as question answering, document assistance, and knowledge management systems.

\subsection{Motivations}

One of a key components of any RAG pipeline is the preprocessing of document corpus. In particular, documents must be segmented into smaller textual units before the indexation. This process is named \emph{chunking}. It consists of dividing documents into manageable passages called \emph{chunks}.

Chunking addresses several structural constraints of retrieval systems. First, it permits large documents to be indexed and efficiently stored in vector databases. Second, it improves retrieval precision. The system matches a query with the most relevant passage instead of an entire document. Finally, chunking ensures that retrieved content remains compatible with the limited context window of LLMs.

However, chunking strategies present some issues. Many approaches rely on overlapping chunks in order to preserve contextual continuity between them. This technique increases the likelihood that relevant information will be captured during retrieval. At the same time, overlapping introduces redundancy in the indexed corpus. As a result, the number of stored chunks grows significantly and this growth can negatively impact indexing costs, retrieval efficiency, and memory usage.

\subsection{Goals}

This work aims to investigate whether part of the redundancy introduced by common chunking strategies can be reduced by applying the filtering mechanisms during the preprocessing stage. More specifically, we explore an approach that investigate the combination of filtering techniques with chunking strategies. Such combinations aim decrease the quantity of indexed chunks while maintaining effective retrieval performance.

The central hypothesis of this study is that redundant chunks can be partially eliminated through adapted filtering methods without significantly degrading the quality of retrieved information. Such an approach could improve the efficiency of the indexing and retrieval components of RAG pipelines by reducing storage and computation requirements and preserving most of the information needed for downstream use.

The contributions of this work are threefold:
\begin{itemize}
	\item an analysis of redundancy introduced by common chunking strategies,
	\item the proposal of a filtering-based preprocessing approach to reduce unnecessary chunks,
	\item an empirical evaluation of lexical, semantic, and structure-aware filtering strategies on retrieval quality and index compactness in a RAG-oriented retrieval setting.
\end{itemize}

\section{Theoretical Background and Related Work}
\label{sec:background}

\subsection{Retrieval-Augmented Generation: Principles, Advantages, and Limitations}

\begin{figure}[!htbp]
	\includegraphics[width=\linewidth]{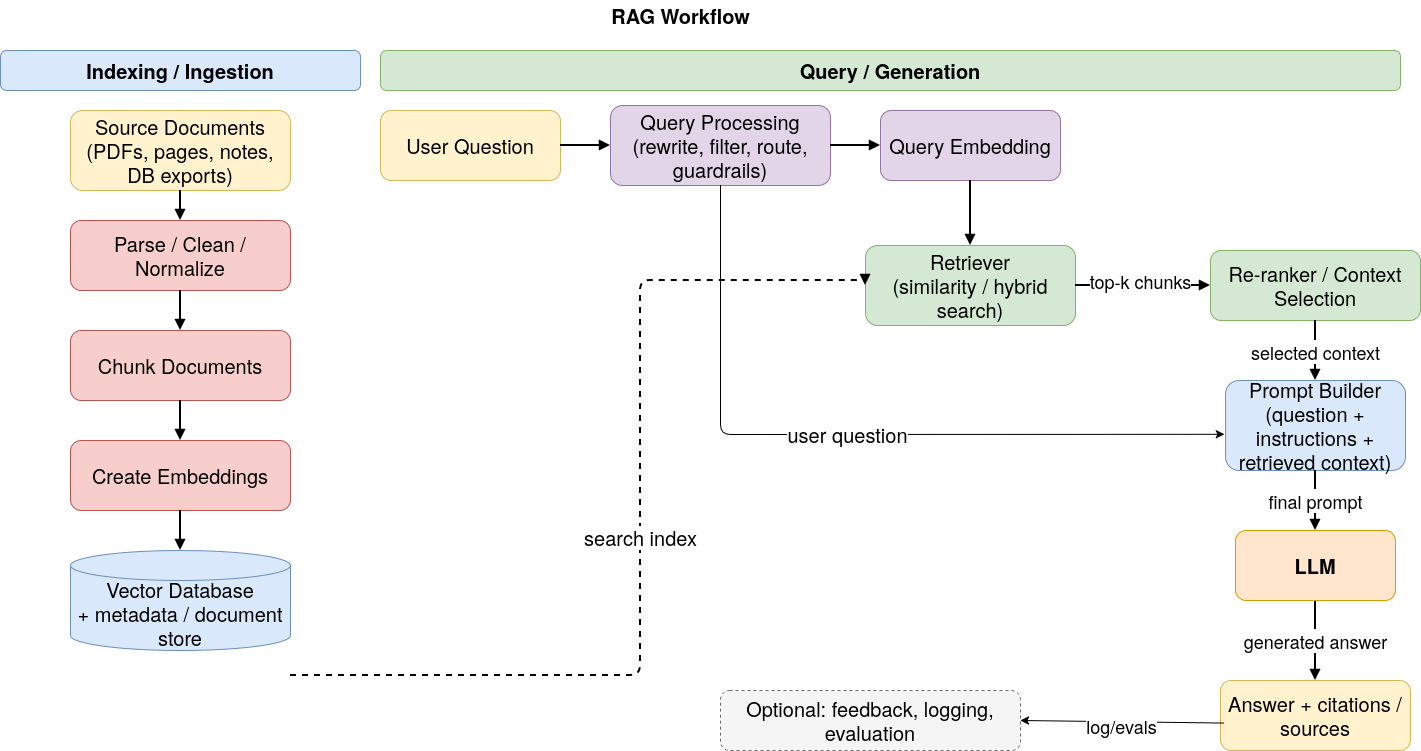}
	\caption{RAG workflow}
	\label{fig:rag_wf}
\end{figure}

Retrieval-Augmented Generation (RAG) integrates a large language model (LLM) with an external retrieval system. This combination permits to generate responses based on a collection of documents, rather than depending on the model's internal memory. It is unnecessary to store all documents directly inside the parameters of the model. A RAG system is capable to retrieve pieces of information from a document corpus. Then, the retrieved relevant information is provided to the model during generation during generation (fig.~\ref{fig:rag_wf}). This general framework was formalized by Lewis et al.~\cite{lewis_retrieval-augmented_2021} and has since become a major paradigm for building question-answering and document-assistance systems.

The main focus of recent studies has been on optimizing retrieval-oriented components of RAG pipelines. Some approaches improve document representation before retrieval, for instance through hypothetical document generation (HyDE), which rewrites the query into a pseudo-document better aligned with relevant passages \cite{gao-etal-2023-precise}. Others modify how chunks are represented or contextualized, as in late chunking \cite{guenther2025latechunkingcontextualchunk} and contextual retrieval \cite{anthropic2024contextualretrieval}, which aim to preserve surrounding document context when building retrieval units. Hierarchical retrieval methods such as RAPTOR \cite{sarthi2024raptorrecursiveabstractiveprocessing} further address long-document reasoning by combining chunk-level retrieval with summary-based navigation over document trees. In parallel, reranking-based approaches, including recent context-ranking methods such as RankRAG \cite{yu2024rankragunifyingcontextranking}, optimize the ordering of retrieved chunks after the initial retrieval stage.

The present work is complementary to these approaches. Rather than modifying the query, enriching chunk representations, or reranking retrieved passages, it focuses on redundancy reduction at the chunk level and studies how different filtering signals interact with different chunking strategies before downstream generation.

Alternative retrieval-augmented architectures have also been proposed, including REALM \cite{guu2020realm}, Fusion-in-Decoder models for open-domain question answering \cite{izacard-grave-2021-leveraging}, and large-scale retrieval-based language models such as RETRO \cite{borgeaud2022retro}. More recent approaches further explore adaptive retrieval and self-reflective generation strategies \cite{asai2024selfrag}.

Conventionally, a RAG pipeline can be decomposed into three stages. The first stage is \emph{indexation}. The system is preparing the document corpus for retrieval. Sources documents pass throw preprocessing. Then these documents are segmented into smaller units called \emph{chunks}. Afterwards, the chunks are transformed into numerical representations using an embedding model. The embeddings permit to capture the semantic meaning of the text. Moreover, it allows the system to compare chunks based on their semantic similarity. These vectors are then stored in an index - a vector database. This type of a storage enables efficient similarity search. 

The second stage is \emph{retrieval}. When a user submits a query, this query is converted into an embedding vector using the same embedding model. Then, the system searches the index to find the $k$ chunks whose vectors are the most similar to the query vector. These retrieved chunks are supposed to contain the most relevant information for answering the query.

The third stage is \emph{generation}. The retrieved chunks are inserted into the prompt provided to the LLM. Then, the model produces a response using the user query and the retrieved context. 
The model then produces a response conditioned on both the user query and the retrieved context. Thus, the generation process is guided by external information and do not rely on the model's internal knowledge.

Although the evaluation of RAG pipelines are based on the quality og generated answers. However, the effectiveness of generation stage is strongly dependent on the nature, diversity, and compactness of the retrieved context. For this reason, retrieval-oriented preprocessing can be studied as a distinct optimization problem, even when end-to-end generation is not directly evaluated.

This architecture offers several important advantages. First, it helps reduce hallucinations because the generated response is grounded in real documents. Second, it gives the possibility to integrate domain-specific or recently knowledge without retraining the language model. Updating the document collection is often sufficient to update the system’s knowledge. Third, RAG systems improve transparency, since the retrieved chunks can be shown to users as supporting evidence for the generated answer.

Despite these advantages, RAG systems also present several limitations. One major limitation is their strong dependence on the quality of the retrieval stage. If the retrieved chunks are irrelevant, incomplete, or redundant, the language model receives poor context and the generated answer will also be inadequate. This situation is described by the principle "garbage in, garbage out."

Another limitation is related to the finite context window of LLMs. Only a limited amount of text can be provided to the model at generation time. In the sutuation where too many chunks are retrieved, some of them may not fit into the context window. The redundant chunks are discarded. Even when they do fit, the model may struggle to use all the information effectively. A phenomenon known as \emph{lost in the middle} \cite{salvatore_lost_2025} has been observed, where information located in the middle of a long context becomes less influential than information placed near the beginning or the end.

Finally, the documents segmentation method can affect the retrieval performance. The segmentation could break the natural structure of the text (for example, coreference) or separate related sentences. Thus, important information may become harder to retrieve. For this reason, document segmentation is not only a technical preprocessing step but also an important design choice that directly influences the effectiveness of the entire RAG pipeline.

\subsection{Document Representation and Chunking Strategies}

Chunking is a most important operation in RAG pipelines. This operation defines the level of detail at which information is represented, indexed and retrieved. A chunk should be small enough to fit within a generation pipeline of and LLM. At the same time, a chunk should be large enough to preserve semantic coherence and contextual meaning.

Chunking therefore reflects two types of constraints. The first is a technical constraint related to the finite context window of embedding models and LLMs. Large documents cannot be processed as a whole and must be divided into smaller units. The second is a semantic constraint. Chunks should ideally correspond to meaningful textual segments (paragraphs or coherent groups of sentences).

Concerning the retrieval part, the precision of semantic matching depends on chunk size. Large chunks often contain more contextual information, which can help preserve coherence. However, they may also contain both relevant and irrelevant content. As a result, their embedding representation becomes less specific and retrieval accuracy may decrease. Conversely, smaller chunks provide finer-grained retrieval units. They allow the system to identify more precise passages that directly answer a query. However, excessively small chunks fragment the text. This fragmentation removes the context needed to correctly interpret the information.

Passage-level retrieval has been studied in open-domain question answering systems. Some dense retrieval models, for example, Dense Passage Retrieval (DPR) demonstrate that dividing documents into passages s	ignificantly improves retrieval accuracy \cite{karpukhin_dense_2020}. Subsequent work further investigated the effect of passage granularity and document segmentation on dense retrieval performance \cite{xiong2021approximate}. These studies highlight that the segmentation of documents into retrieval units is a critical element for modern retrieval-augmented architectures.

Recent work has proposed more contextualized alternatives to standard chunking. Late chunking delays the pooling operation until after long-context encoding, so that chunk embeddings retain information from surrounding text rather than being computed from isolated segments \cite{guenther2025latechunkingcontextualchunk}. Contextual retrieval similarly enriches chunk representations by incorporating document-level context before retrieval, for example through contextual embeddings and contextual BM25 \cite{anthropic2024contextualretrieval}. These approaches aim to improve the quality of retrieval units without explicitly reducing index size. In contrast, the present study focuses on the complementary problem of identifying and removing redundant chunks once a chunking strategy has been applied.

Several chunking strategies have been proposed in the literature. Each of them is offering different trade-offs between simplicity, computational cost, and semantic coherence.

The most basic approach is \emph{fixed-size splitting}. In this strategy, the documents are divided into segments of a predefined number of characters or tokens. This approach is simple and computationally efficient. However, it ignores linguistic structure and may break sentences across chunk boundaries. This breakage is potentially degrading the quality of the resulting embeddings. For this reason, fixed-size chunking is often used as a baseline rather than as an optimal strategy.

A more advanced method is \emph{recursive splitting}. This strategy tends to preserve the natural structure of the document. In this strategy, the text is split using larger separators such as sections, paragraphs, or sentences. If the resulting segments are still too long, they are further divided into smaller units. Recursive chunking is widely used in practice. This method provides a good compromise between structural coherence and computational efficiency.

More recent approaches rely on \emph{semantic chunking}. These methods do not use predefined separators. Instead, is analyzes the semantic similarity between adjacent sentences. When a significant semantic shift is detected, a new chunk boundary is created. This approach produces segments that are more semantically coherent. Thus, it improves retrieval quality. However, it also requires additional computation because embeddings must be calculated during the segmentation process.

Many chunking strategies use an overlapping technique in order to balance precision and coherence. Overlapping consists on sharing a portion of the content between consecutive chunks. This technique helps preserve continuity between chunks and reduces the risk that important information will be split across chunk boundaries. The overlapping could improve retrieval robustness. However, it also introduces significant redundancy. A large portion of the indexed text may appear multiple times across overlapping chunks. This redundancy increases the size of the vector index. Moreover, it leads to multiple similar chunks retrieved for the same query. As a result, the system may waste part of the available context window on duplicated or highly similar information.

Overall, the literature shows that chunk size, overlap ratio, and segmentation logic influence the precision of retrieval and the efficiency of the indexing process. The choice of chunking strategies has therefore become an important research topic in RAG systems.

\subsection{Redundancy and Efficiency in Vector Retrieval Systems}

The chunking improves retrieval granularity. However, it also introduces redundancy into the indexed corpus. Overlapping segmentation strategies are particularly prone to this problem because of the presence of  large portions of text in multiple chunks. More generally, redundancy can take several forms in a vector database.

Redundancy in large text collections has been studied in the context of dataset deduplication and information retrieval efficiency. As it is shown in \cite{lee-etal-2022-deduplicating}., removing near-duplicate documents can significantly improve storage efficiency and retrieval quality. Similar observations have been reported in large-scale retrieval systems. In these systems, redundant passages may reduce the diversity of retrieved contexts \cite{Min2022RethinkingTR}.

First, duplicate documents may lead to identical or nearly identical chunks in index. Second, aggressive overlap between consecutive chunks lead to the presence of a repeated content in the database. Third, boilerplate text such as headers, footers, page numbers, or legal notices may be replicated across many chunks.

This redundancy has several negative consequences. From a storage perspective, redundant chunks increase the size of the vector database and the associated computational costs. Concerning the retrieval, redundancy may reduce the diversity of retrieved results. In a situation of an existence of nearly identical chunks in the database, nearest-neighbour search may return several versions of the same information. And this lead to reducing of the diversity of the retrieved context. Finally, redundant information can also degrade generation quality by occupying valuable space within the language model's context window and by distracting the model with repeated evidence instead of complementary evidence.

One of the most important research topic in RAG systems has therefore become the redundancy in vector indexes. The challenge is not only to remove repeated information. It is also necessary not to harm retrieval recall or deteriorate the informational coverage of the corpus.

\subsection{Filtering Mechanisms for Redundancy}

In order to overcome the problem of redundant chunks, an additional filtering stage can be implemented in order to reduce the number of stored chunks. The goal of this filtering step is not only to remove segments with no additional information. It also aimss to preserve the diversity of the indexed corpus.

Recent studies have explored hybrid retrieval pipelines combining semantic similarity with additional structural signals such as entity information or document clustering in order to improve retrieval diversity and reduce redundancy \cite{gao_retrieval-augmented_2024}. These approaches suggest that combining multiple complementary signals could improve the filtering strategies.

To detect redundant chunks, several approaches have been proposed. In the present work, redundancy is not defined only through lexical similarity. Several complementary signals are considered: semantic similarity, topic similarity, and named-entity overlap. These methods capture different aspects of the information in the text. This fact allows the filtering process to identify chunks that are highly correlated.

\paragraph{Locality-sensitive hashing and lexical deduplication.}

Locality-sensitive hashing (LSH) methods represents a baseline for near-duplicate detection in large text collections. In particular, MinHash estimates Jaccard similarity between sets. LSH makes it possible to retrieve candidate near-duplicates without pairwise comparison. These methods are widely used in web-scale deduplication and corpus cleaning because they are simple, fast, and robust to small lexical variations \cite{broder1997resemblance,broder2000identifying,leskovec2020mining}. In the context of chunk filtering, MinHash-LSH constitutes an important lexical baseline against which embedding-based and structure-aware approaches can be compared.

\paragraph{Semantic similarity.}

One common strategy relies on semantic similarity between chunk embeddings. In natural language processing (NLP) systems, texts are represented as dense vectors. This representation is possible by the use of models such as Sentence Transformers \cite{reimers_sentence-bert_2019}. Thus, the similarity between two chunks can be measured applying metrics such as cosine similarity.

Let $f(\cdot)$ denote the embedding model that maps a text segment into a dense vector space:

\[
\mathbf{e}_c = f(c), \qquad \mathbf{e}_q = f(q),
\]

where $c$ is a chunk and $q$ is a query.

The semantic similarity between a query and a chunk is measured using cosine similarity:

\[
\operatorname{sim}(q,c) =
\frac{\mathbf{e}_q \cdot \mathbf{e}_c}
{\|\mathbf{e}_q\| \, \|\mathbf{e}_c\|}.
\]

Chunks that are highly similar in the embedding space are likely to contain overlapping information. Filtering based on semantic similarity can therefore remove redundant segments and preserve the overall informational coverage of the corpus. Recent work \cite{sarthi2024raptorrecursiveabstractiveprocessing} has also explored graph-based retrieval structures built on semantic similarity relations. This shows that redundancy and relevance can be studied not only at the level of isolated chunks. It can be studied also at the level of chunk-to-chunk structure.

\paragraph{Topic modeling.}

Topic modeling represents another way to analyze redundancy at a high semantic level. Conventional topic modeling methods (Latent Dirichlet Allocation (LDA) \cite{blei_latent_2003}) represent documents as set of latent topics inferred from word distributions. More recent approaches such as BERTopic \cite{grootendorst_bertopic_2022} combine transformer-based embeddings with dimensionality reduction and clustering in order to capture deeper semantic relations between texts.

Applied to chunks, topic modeling provides a representation of the themes discussed in each segment. Two chunks may therefore be associated with the same topic even if they do not share many surface words. This property makes topic-based filtering useful for identifying thematic redundancy across a corpus and reducing the number of indexed segments while maintaining coverage of the main subjects.

\paragraph{Named Entity Recognition (NER).}

NER is used to identify entities such as persons, organizations, locations, dates, or domain-specific concepts within a text. Recent NER systems, such as those implemented in spaCy \cite{honnibal_spacy_2020}, use neural models trained on annotated corpora. This allows to detect these entities automatically.

In the context of document chunking, named entities can serve as structural signals. These signals could describe the informational content of a chunk. Chunks that share many entities may refer to the same events, actors, or concepts. Analyzing entity overlap therefore provides an additional indicator of potential redundancy between segments. Entity-based signals are useful in technical, factual, or domain-specific corpora. In such corpus, the central information is often organized around recurring entities.

Overall, these filtering mechanisms define a progressive view of redundancy. Semantic similarity captures direct textual resemblance. Topic modeling captures shared thematic content. NER captures structural relationships through shared entities. By combining these signals, it becomes possible to reduce the number of indexed chunks and preserve the diversity of information needed for effective retrieval.

The central challenge is therefore to find a balance between reducing the number of stored chunks and maintaining the recall required to retrieve relevant information. This trade-off between efficiency and retrieval performance is at the core of the present study.

\begin{figure}[t]
	\includegraphics[width=\linewidth]{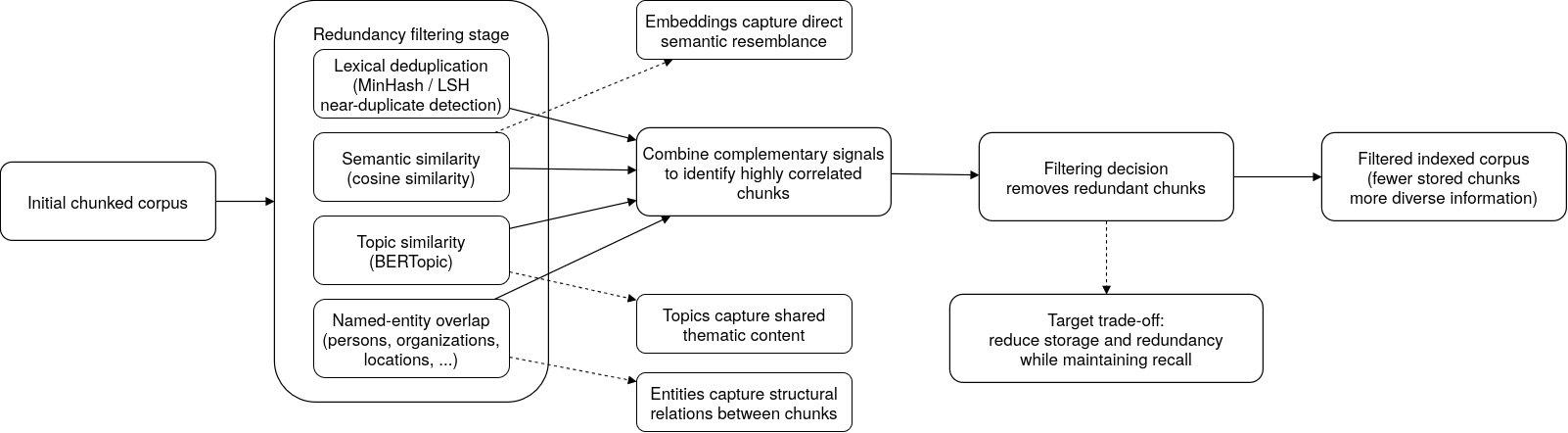}
	\caption{Combining multiple signals to remove redundant chunks and maintain diverse, useful information.}
\end{figure}

\section{Methodology}
\label{sec:methodology}

\subsection{Experimental Objective and Hypothesis}

The objective of this study is to evaluate whether the redundancy introduced by chunking strategies in a Retrieval-Augmented Generation (RAG) pipeline can be reduced through adapted filtering mechanisms without causing a substantial loss in retrieval quality.

More specifically, the study investigates the trade-off between two goals: reducing the size of the vector index and preserving the ability of the system to retrieve the information needed to answer a query. The central idea is that some chunking strategies, especially those using overlap or producing semantically close segments, may generate many chunks that are partially redundant. The system may become more efficient and maintain acceptable performance if the redundant chunks can be removed.

The central hypothesis of this work is therefore the following: \emph{part of the redundancy introduced by common chunking strategies can be eliminated by suitable filtering methods without significantly degrading retrieval performance}. Under this hypothesis, it should be possible to reduce the number of indexed chunks. Moreover, the useful information for downstream generation should be preserved.

\subsection{Data and Corpora Used}

In order to evaluate this hypothesis, the study relies on several heterogeneous corpora covering different writing styles, document structures, and domains. We test the approaches on different corpora because chunking and filtering strategies may behave differently depending on the type of text being processed.

The first group of corpora follows the experimental setting used in the Chroma chunking study \cite{noauthor_chroma-corechroma_2025}. It includes five datasets.

\textbf{State of the Union Address 2024} \cite{noauthor_state_nodate} is a well-structured political speech transcript. It provides relatively clean and homogeneous text. This fact makes it useful for studying chunking behavior in a controlled setting.

\textbf{Wikitext} \cite{merity_pointer_2016} is derived from high-quality Wikipedia articles. It contains long-form expository text. Moreover, the its structure and its topic variation are richer than the State of the Union corpus.

\textbf{Chatlogs} \cite{ding_enhancing_2023}, based on the UltraChat 200k dataset, contains dialogue-style content. In this study, the raw text, including surrounding formatting artifacts, is preserved in order to better reflect realistic production data.

\textbf{Finance} \cite{chen_convfinqa_2022}, derived from ConvFinQA, contains financial question-answering material with numerical reasoning and longer contextual dependencies. This corpus is useful for testing whether filtering remains effective in documents where key information is distributed across several segments.

\textbf{Pubmed} \cite{noauthor_pmcopen_access_2023} is extracted from biomedical and life sciences articles. It represents a technical domain with dense information and domain-specific terminology.

In addition to these corpora, we use two external evaluation datasets.

\textbf{SQuAD 1.1 dev} \cite{noauthor_rajpurkarsquad_2021} is used as a standard benchmark for extractive question answering. It provides questions paired with reference passages from Wikipedia. This fact makes it suitable for evaluating whether retrieved chunks cover the expected information.

\textbf{WebFAQ Retrieval} \cite{dinzinger_webfaq_2025} is used to extend the evaluation to a more heterogeneous and multilingual retrieval setting. In this study, only long documents are retained, in order to focus on cases where chunking and redundancy reduction are genuinely important. This choice makes the evaluation more relevant for realistic RAG scenarios involving large documents, although it also means that the results should be interpreted as applying mainly to long-document settings.

Overall, the selected corpora provide a broad test bed including structured speeches, encyclopedic articles, raw conversational text, technical financial data, biomedical writing, and large web-derived documents. This diversity makes it possible to study the robustness of chunk filtering strategies across different retrieval conditions.

\subsection{Chunking Configurations}

To study the interaction between segmentation and filtering, several chunking strategies are evaluated.

The first family corresponds to \textbf{fixed token splitting}, which serves as the baseline. Documents are divided into windows of fixed size. Different chunk sizes are tested. This variation allows to study the effect of granularity. Overlap is also varied in order to measure the effect of duplicated context.

The second family corresponds to \textbf{recursive token splitting}. In this strategy, the system first tries to preserve larger textual units (paragraphs or sentences). It splits the text when necessary. This method permits to follow the initial structure of the document (better than fixed-size chunking).

The third family corresponds to \textbf{semantic chunking}. These methods segment documents according to semantic transitions rather than fixed boundaries. Several variants are considered: cluster-based semantic chunking and chunking based on local semantic breakpoints. These methods aim to produce more coherent chunks. Although they may also generate variable chunk lengths.

Using several chunking configurations is essential for the present study, since the effect of filtering may depend strongly on the type of redundancy produced by the segmentation strategy.

\subsection{Embedding Models and Vector Indexing}

The experiments rely on dense retrieval in an embedding space. Each chunk and each query is encoded into a dense vector representation. Then retrieval is performed by nearest-neighbour search in a vector database.

In the broader study, two families of embedding models are considered: Sentence Transformers, using \texttt{all-MiniLM-L6-v2} \cite{reimers_sentence-bert_2019}, and BGE, using \texttt{BAAI / bge-m3} \cite{chen_m3-embedding_2025}. However, the main filtering experiments are conducted with \texttt{all-MiniLM-L6-v2}, which is chosen as the reference model because it offers a good balance between speed, reproducibility, and computational cost. The vector index is implemented using ChromaDB \cite{noauthor_chroma-corechroma_2025}.

The indexing and retrieval settings are kept constant within each experimental series. This helps to isolate the effects of chunking and filtering. 

\subsection{Filtering Strategies Evaluated}

The filtering stage is applied after chunk generation. Starting from the full set of chunks produced by a given chunker, the system removes chunks considered redundant according to one or more filtering signals.

Several filtering strategies are evaluated.

\textbf{ExactNorm.}  
As a first lexical baseline, exact deduplication is applied after text normalization. Two chunks are considered duplicates when their normalized textual forms are identical. This baseline captures exact repetition but cannot detect approximate duplicates.

\textbf{MinHash-LSH.}  
As a second lexical baseline, we evaluate a near-duplicate detection method based on MinHash and locality-sensitive hashing (LSH) \cite{broder1997resemblance,broder2000identifying}. Each chunk is represented as a set of lexical units. MinHash signatures are used to approximate Jaccard similarity between chunks. Candidate duplicates are retrieved through LSH, and chunks whose estimated lexical similarity exceeds a threshold are removed. Unlike embedding-based filtering, this baseline targets lexical near-duplicates rather than semantic or structural redundancy.

These baselines provide reference points for assessing whether semantic and entity-aware filtering strategies offer benefits beyond standard lexical deduplication.

\textbf{Random filtering.}  
As a control baseline, we evaluate random chunk removal. For each configuration, a proportion of chunks is removed uniformly at random. This allow to match the reduction ratio achieved by a filtering method. This baseline is used to determine whether the observed trade-offs are due to the filtering signal itself or simply to the reduction of index size.

\textbf{Similarity only.}  
This strategy relies only on semantic similarity between chunk embeddings. If two chunks are above a given similarity threshold, one of them is removed. This is the most direct form of semantic deduplication.

Given two chunks $c_i$ and $c_j$, they are considered redundant if
\[
\operatorname{sim}(c_i,c_j) \geq \tau,
\]
where $\tau$ is a similarity threshold.

The filtered index $\mathcal{C}' \subseteq \mathcal{C}$ is obtained by removing chunks that are judged redundant according to the filtering rule.

\textbf{Similarity + Topics.}  
In this configuration, semantic similarity is combined with topic information. Chunks are first associated with thematic groups, and redundancy is mainly assessed within the same topic. This reduces the risk of removing chunks that are semantically close but the subject discussed is different.

In the topic-aware setting, redundancy is evaluated only when two chunks belong to the same topic:
\[
\operatorname{Redundant}(c_i,c_j) =
\bigl[\operatorname{sim}(c_i,c_j) \geq \tau \bigr]
\land
\bigl[t(c_i) = t(c_j)\bigr],
\]
where $t(c)$ denotes the topic assigned to chunk $c$.

\textbf{Similarity + NER.}  
This strategy combines semantic similarity with named-entity information. Semantic redundancy is only considered stronger when chunks also share entity-level structure, such as common persons, organizations, places, or dates.

In the entity-aware setting, redundancy is conditioned on entity overlap:
\[
\operatorname{Redundant}(c_i,c_j) =
\bigl[\operatorname{sim}(c_i,c_j) \geq \tau \bigr]
\land
\bigl[E(c_i) \cap E(c_j) \neq \varnothing\bigr],
\]
where $E(c)$ is the set of named entities extracted from chunk $c$.

\textbf{Similarity + Topics + NER.}  
This hybrid setting combines all three signals: semantic similarity, thematic grouping, and named-entity overlap. It is the most selective configuration and is designed to capture several types of redundancy at the same time.

\textbf{NER Exact.}  
In this configuration, two chunks are considered redundant when they contain the same set of named entities. This strategy focuses on strict structural duplication and not on global semantic similarity. 

For exact entity-based filtering, two chunks are considered redundant if
\[
E(c_i) = E(c_j).
\]

This behavior does not necessarily imply a loss of useful information for retrieval, since alternative chunks may still provide overlapping or equivalent content.

\textbf{NER Half.}  
In this variant, a chunk may be removed when at least 50\% of its named entities overlap with another chunk. This criterion is less strict than exact matching and aims to capture partial informational redundancy.

For partial entity overlap, redundancy is defined as
\[
\frac{|E(c_i) \cap E(c_j)|}{|E(c_i)|} \geq 0.5.
\]

These different strategies make it possible to compare pure semantic filtering, structure-aware filtering, and hybrid filtering in a common evaluation framework.

A methodological limitation of entity-based filtering consists on the fact that it can only operate on chunks with at least one recognized named entity. Chunks that are purely descriptive, procedural, or narrative may therefore remain unaffected by \emph{NER Exact} and \emph{NER Half}. As a consequence, the effective scope of these filters depends on the entity density. In domains where named entities are frequent (finance or biomedical) entity-based filtering may affect a larger portion of the index than in corpora containing fewer explicit entities.

\subsection{Retrieval and Evaluation Protocol}

All experiments use dense retrieval. Queries are encoded in the same embedding space as the chunks, and the system retrieves the top-$k$ nearest neighbours from the vector index. In all conditions, $k$ is fixed to 5 in order to ensure comparability across experiments.

For a given query $q$, the retriever returns the set of chunks with the highest similarity scores:
\[
\mathcal{R}_k(q) = \operatorname{TopK}_{c \in \mathcal{C}} \; \operatorname{sim}(q,c),
\]
where $\mathcal{C}$ is the indexed set of chunks.

The objective of the evaluation is not to assess end-to-end answer generation by a language model, but to measure the effect of chunk filtering on the retrieval layer of a RAG pipeline. More precisely, the study examines whether the retrieved context still contains the information which is needed to answer a query after filtering. The results should therefore be interpreted as evidence about retrieval-oriented preprocessing, not as a direct measurement of final answer quality.

To do this, the study uses a token-coverage evaluation protocol. For each question, two token sets are defined. The first is the set of tokens contained in the reference passage. The second is the set of tokens contained in the union of the retrieved top-5 chunks. Based on these two sets, three metrics are computed:

\begin{itemize}
	\item \textbf{Precision}: the proportion of retrieved tokens that belong to the reference passage;
	\item \textbf{Recall}: the proportion of reference tokens covered by the retrieved chunks;
	\item \textbf{IoU (Jaccard)}: the ratio between token intersection and token union.
\end{itemize}

For each query, let $T_e$ denote the set of tokens in the reference passage, and let $T_r$ denote the set of tokens contained in the retrieved top-$k$ chunks.

Precision is defined as
\[
\mathrm{Precision} = \frac{|T_e \cap T_r|}{|T_r|}.
\]

Recall is defined as
\[
\mathrm{Recall} = \frac{|T_e \cap T_r|}{|T_e|}.
\]

Intersection over Union (IoU), also known as Jaccard similarity, is defined as
\[
\mathrm{IoU} = \frac{|T_e \cap T_r|}{|T_e \cup T_r|}.
\]

Two evaluation variants are considered. In the \emph{raw} setting, tokens are lowercased. In the \emph{preprocessed} setting, English stopwords are removed, tokens are lemmatized. This operation reduces the effect of superficial lexical variation.

\begin{figure}[!htbp]
	\includegraphics[width=\linewidth]{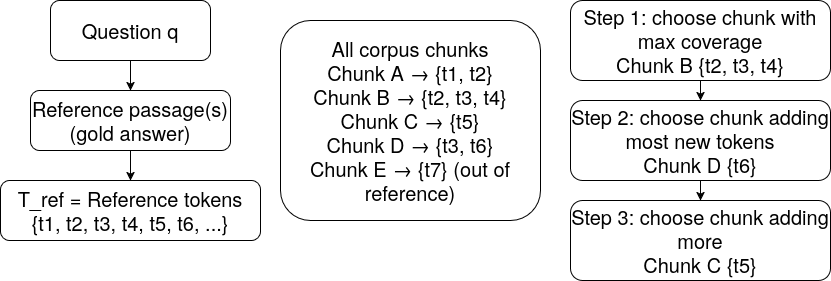}
	\caption{Greedy oracle chunk selection based on token coverage.}
	\label{fig:oracle_selection}
\end{figure}

In addition, an oracle (fig.~\ref{fig:oracle_selection}) upper bound is introduced in order to estimate the best possible retrieval quality achievable with the available chunks. For each question, the oracle greedily selects up to five chunks. This allows to maximize the coverage of reference tokens. This measure allows for the distinguishing of losses caused by the retrieval model from losses caused by fragmentation or filtering. In other words, it indicates whether the required information is still present in the filtered index, even when the retriever does not rank the best chunks first.

\FloatBarrier

\FloatBarrier

Let $\mathcal{S}_k \subseteq \mathcal{C}$ denote a set of at most $k$ chunks. The oracle selects the subset maximizing the coverage of reference tokens:
\[
\mathcal{S}_k^\star
=
\arg\max_{\mathcal{S}_k \subseteq \mathcal{C},\, |\mathcal{S}_k| \leq k}
\left|
T_e \cap \bigcup_{c \in \mathcal{S}_k} T(c)
\right|,
\]
where $T(c)$ denotes the set of tokens contained in chunk $c$.

The token-coverage metrics used in this study should be interpreted as indicators for information accessibility. It does not measures directly the answerability. Although lexical overlap cannot guarantee that a language model produces a correct answer. However, it provides a model-independent way to verify that the required information remains in the retrieved context after filtering. This helps isolate the effect of chunk filtering from other factors (prompting or model issues).

\subsection{Baseline and Comparative Setting}

All results are compared to a reference baseline corresponding to a standard RAG pipeline without chunk filtering. This baseline uses chunking, indexing of all generated chunks, dense retrieval in ChromaDB, and top-5 retrieval.

This reference setting provides a clear comparison point for all filtering strategies. Improvements in index size or declines in retrieval quality can be assessed relative to a standard production-like setup.

\subsection{Methodological Scope}

The study is intentionally focused on retrieval and document-level preprocessing. It does not directly evaluate the final quality of the generated answers produced by a language model, and therefore does not claim to measure end-to-end RAG performance. Instead, it investigates an upstream question: how chunk filtering affects index compactness and the ability of the retrieval stage to preserve access to relevant information.

Similarly, no reranking model is introduced, since reranking would add another decision layer and make it harder to isolate the effect of chunking and filtering alone. The purpose of this methodology is therefore not to optimize a complete RAG system end to end, but to isolate and analyze one specific component of the pipeline: the interaction between chunk filtering, chunking strategy, and retrieval quality.

As a consequence, the reported results should be interpreted as retrieval-level evidence about the usefulness of filtering mechanisms in RAG-oriented pipelines. It remains to be investigated if these improvements translate to downstream generation..

\subsection{Methodological Scope}

The study is intentionally focused on retrieval and document-level preprocessing. It does not directly evaluate the final quality of the generated answers produced by a language model, and therefore does not claim to measure end-to-end RAG performance. Instead, it investigates an upstream question: how chunk filtering affects index compactness and the ability of the retrieval stage to preserve access to relevant information.

Similarly, no reranking model is introduced, since reranking would add another decision layer and make it harder to isolate the effect of chunking and filtering alone. The purpose of this methodology is therefore not to optimize a complete RAG system end to end, but to isolate and analyze one specific component of the pipeline: the interaction between chunk filtering, chunking strategy, and retrieval quality.

As a consequence, the reported results should be interpreted as retrieval-level evidence about the usefulness of filtering mechanisms in RAG-oriented pipelines. Whether these gains transfer directly to downstream generation remains an important question for future work.

\section{Results and Discussion}

\subsection{Chroma Corpora with BGE Embeddings}
\label{sec:results-chroma-bge}

This first benchmark treat the Chroma-based chunking setting using the \texttt{BAAI/bge-m3} embedding model and the token-based evaluation protocol introduced in Section~\ref{sec:methodology}. The objective is to analyze how different filtering strategies influence the size of the index and the quality of retrieval. Across all experiments, the retrieval pipeline is kept fixed. We preserve the chunking configuration, embedding model, top-$k$ retrieval parameter, and evaluation method. We apply only the filtering strategy.

As a first comparison point, we also evaluate lexical and control baselines, namely \emph{ExactNorm}, \emph{MinHash-LSH}, and \emph{Random filtering}. ExactNorm and MinHash-LSH capture exact or near-exact lexical duplication, while random filtering removes chunks without using any content-based signal. These baselines allow to separate the effect of structured redundancy reduction from the effect of index reduction.

\subsubsection{Similarity-based filters}
\begin{figure}[!htbp]
    \centering

    \begin{subfigure}{0.48\textwidth}
        \includegraphics[width=\linewidth]{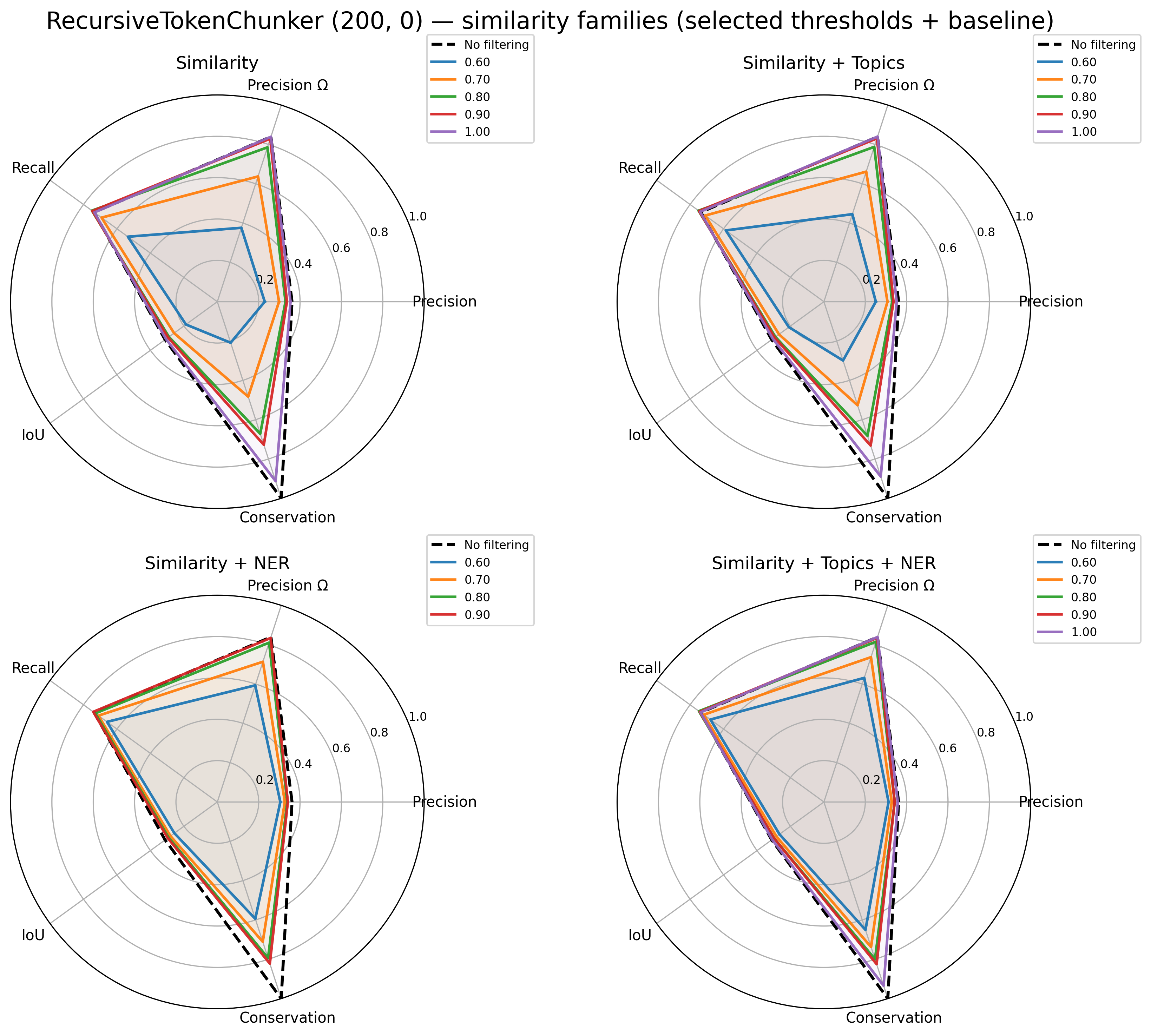}
        \caption{}
    \end{subfigure}
    \hfill
    \begin{subfigure}{0.48\textwidth}
        \includegraphics[width=\linewidth]{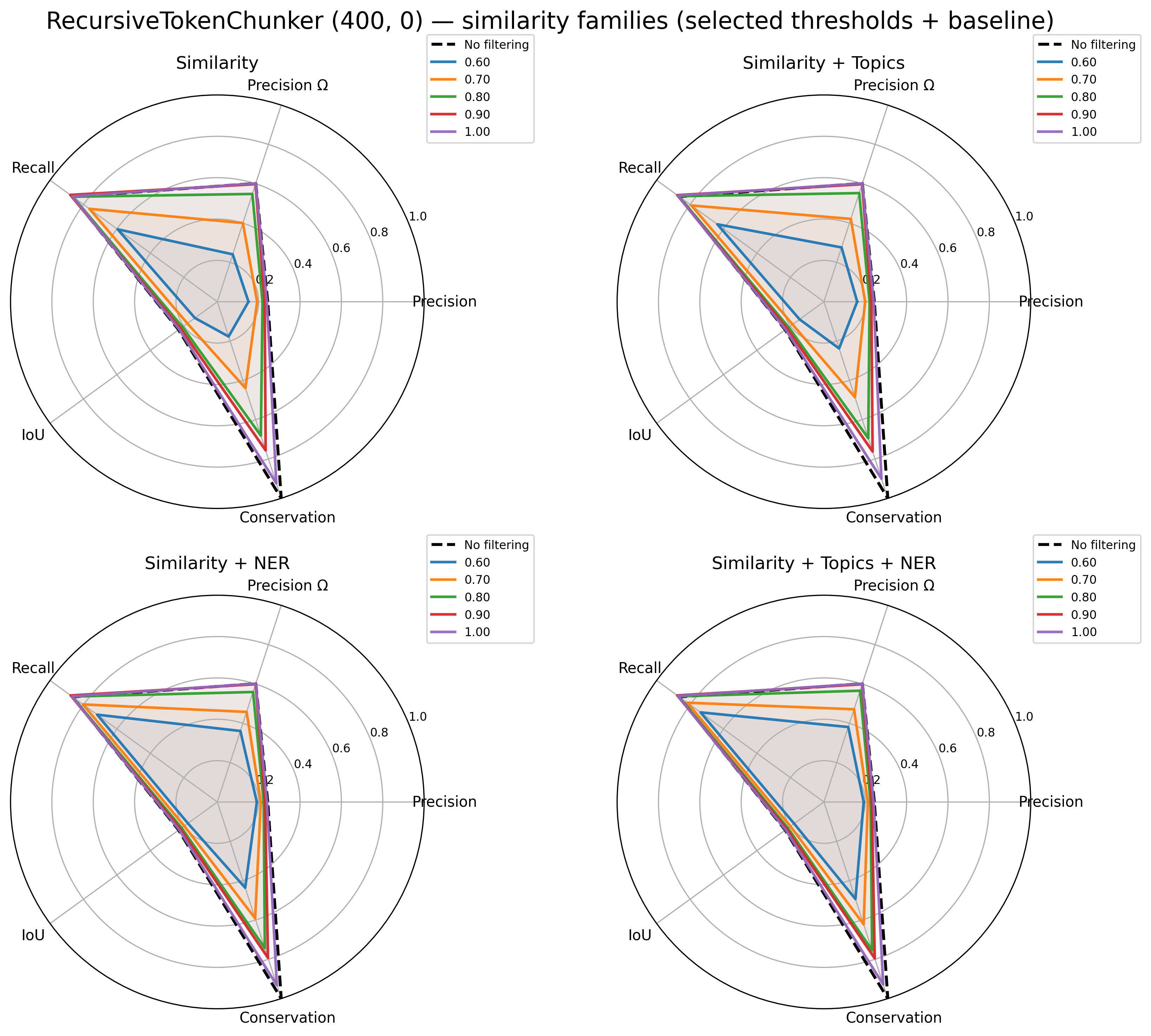}
        \caption{}
    \end{subfigure}

    \vspace{0.5em}

    \begin{subfigure}{0.48\textwidth}
        \includegraphics[width=\linewidth]{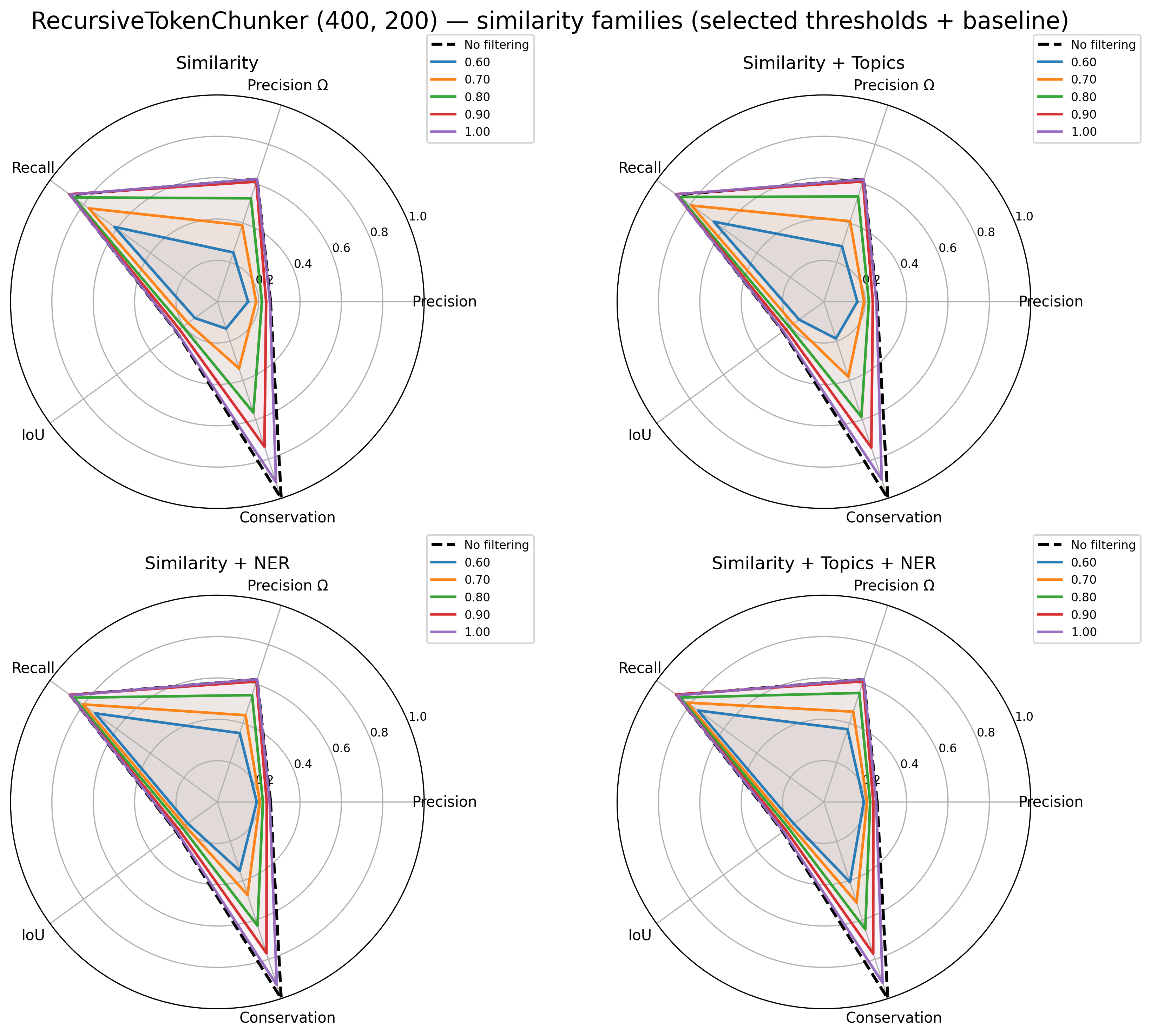}
        \caption{}
    \end{subfigure}
    \hfill
    \begin{subfigure}{0.48\textwidth}
        \includegraphics[width=\linewidth]{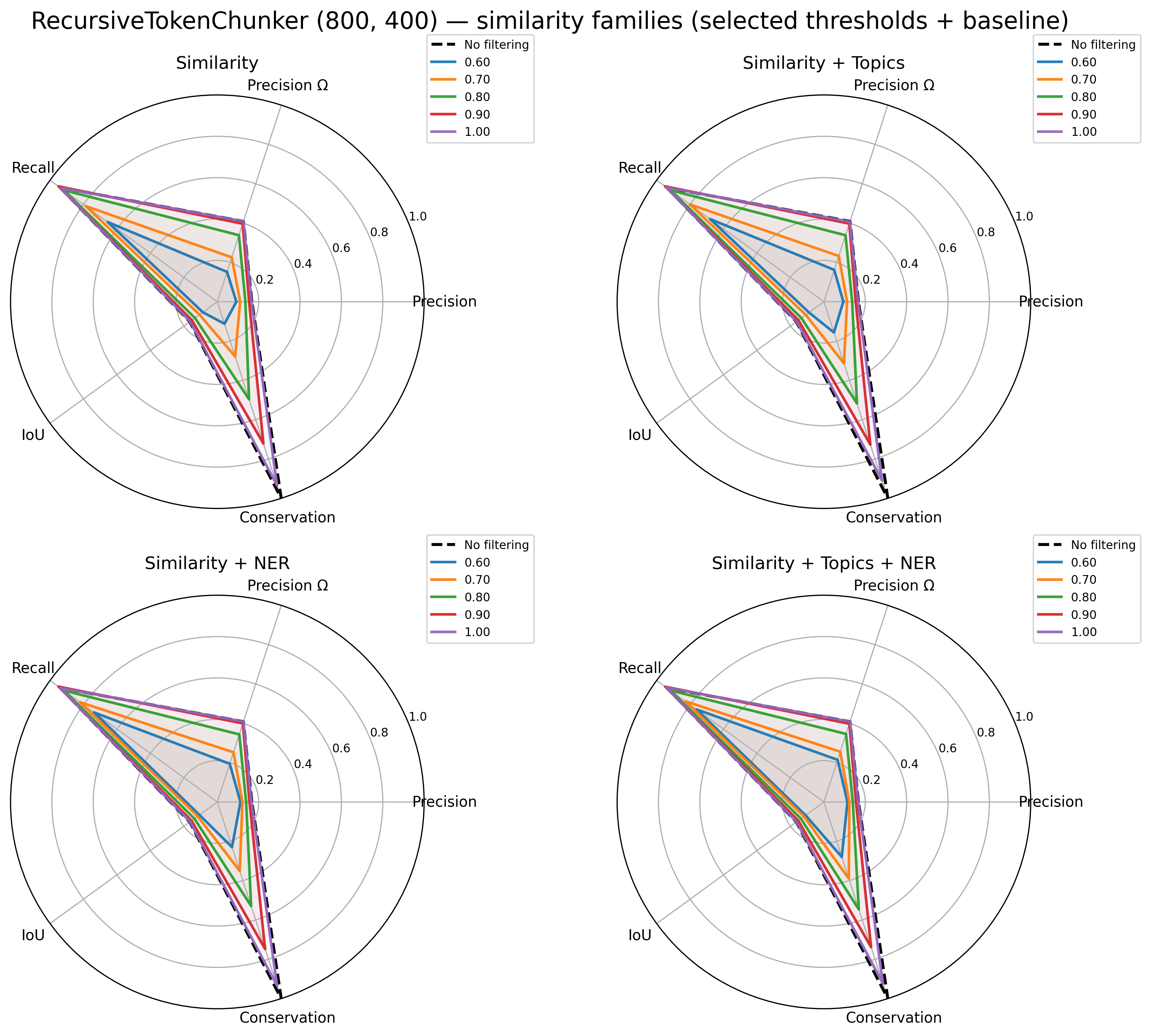}
        \caption{}
    \end{subfigure}

    \caption{RecursiveTokenChunker and similarity-based filtres for Chroma corpus}
    \label{chroma:rec-sim}
\end{figure}

\begin{figure}[!htbp]
    \centering

    \begin{subfigure}{0.48\textwidth}
        \includegraphics[width=\linewidth]{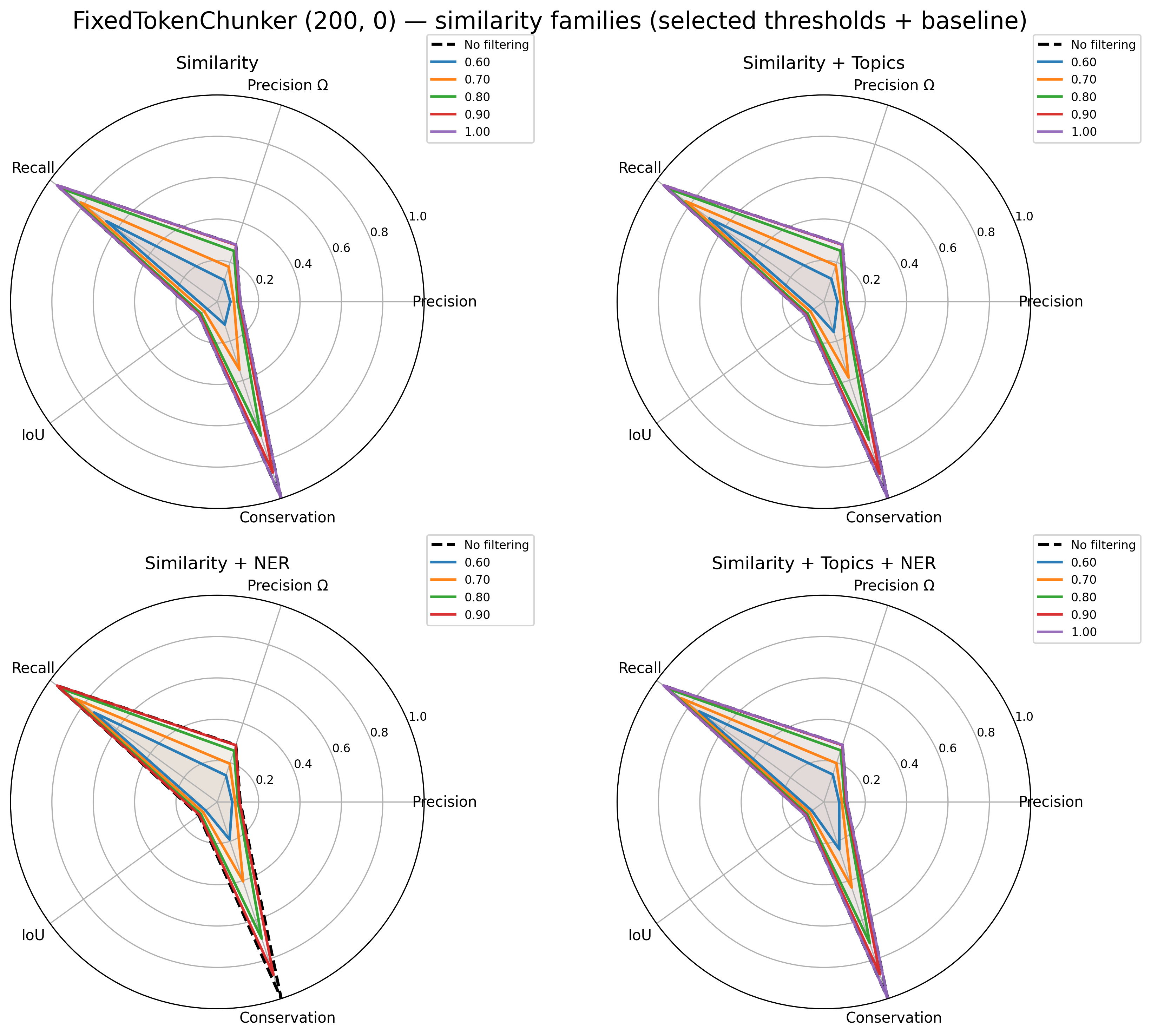}
        \caption{}
    \end{subfigure}
    \hfill
    \begin{subfigure}{0.48\textwidth}
        \includegraphics[width=\linewidth]{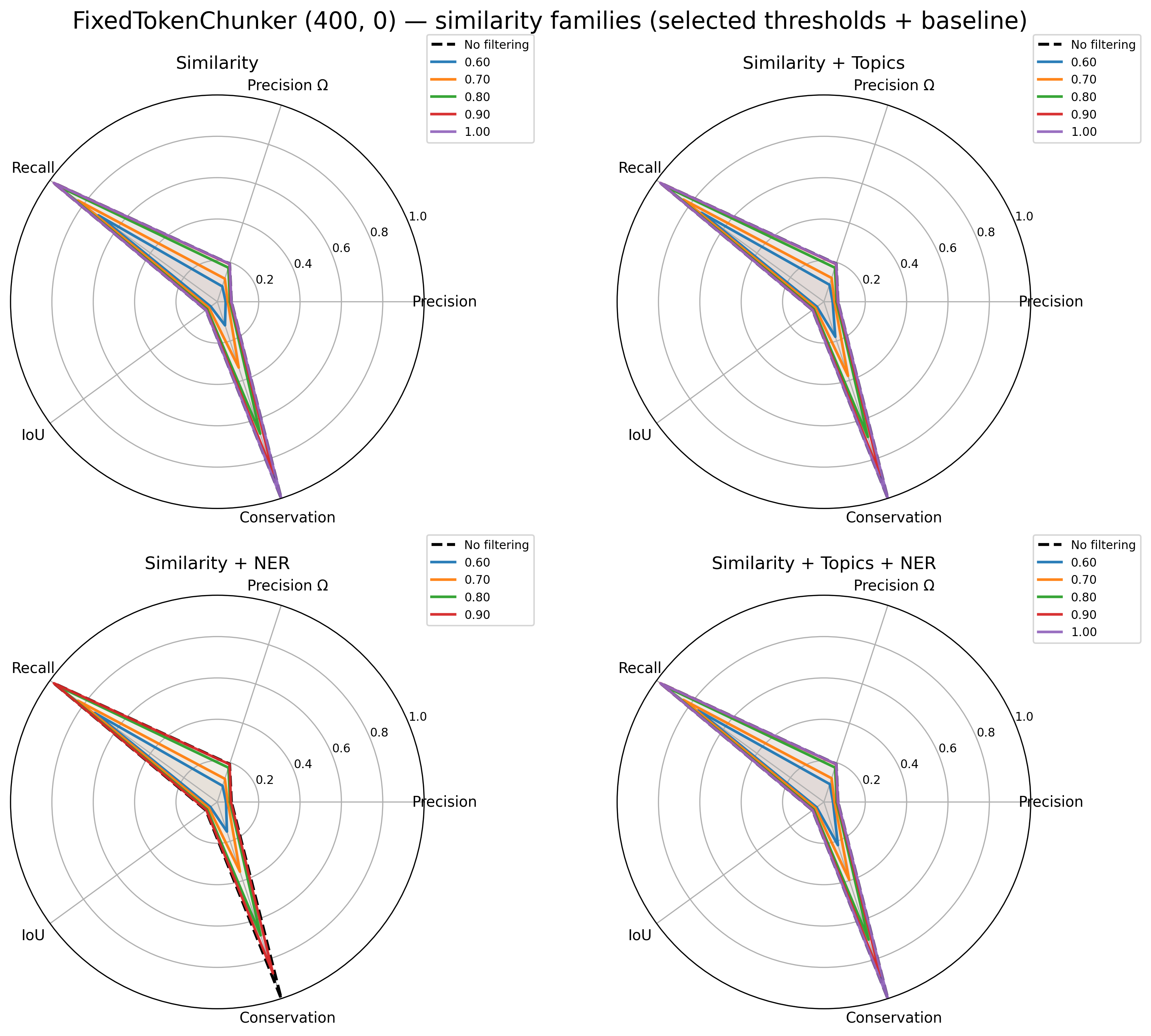}
        \caption{}
    \end{subfigure}

    \vspace{0.5em}

    \begin{subfigure}{0.48\textwidth}
        \includegraphics[width=\linewidth]{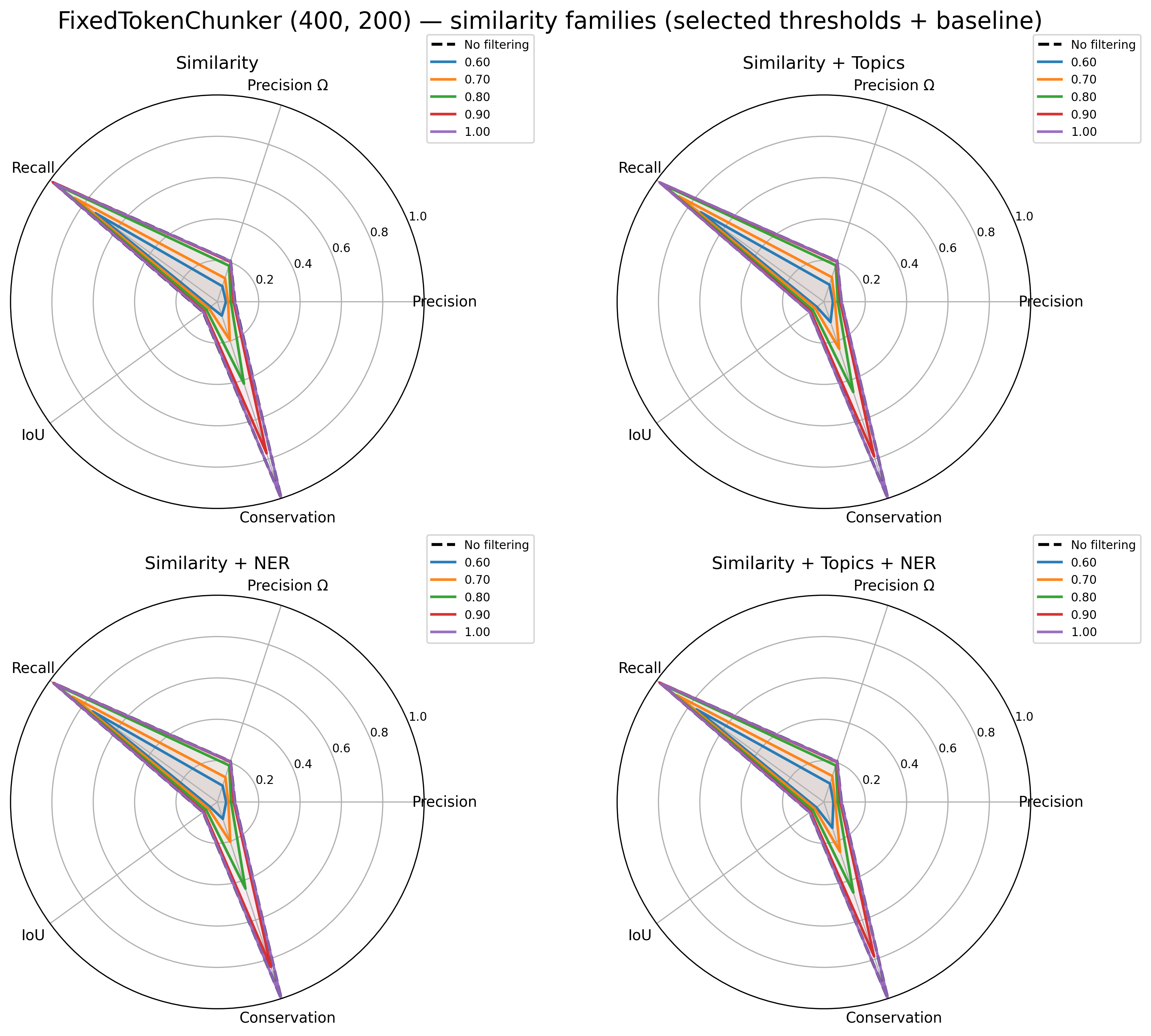}
        \caption{}
    \end{subfigure}
    \hfill
    \begin{subfigure}{0.48\textwidth}
        \includegraphics[width=\linewidth]{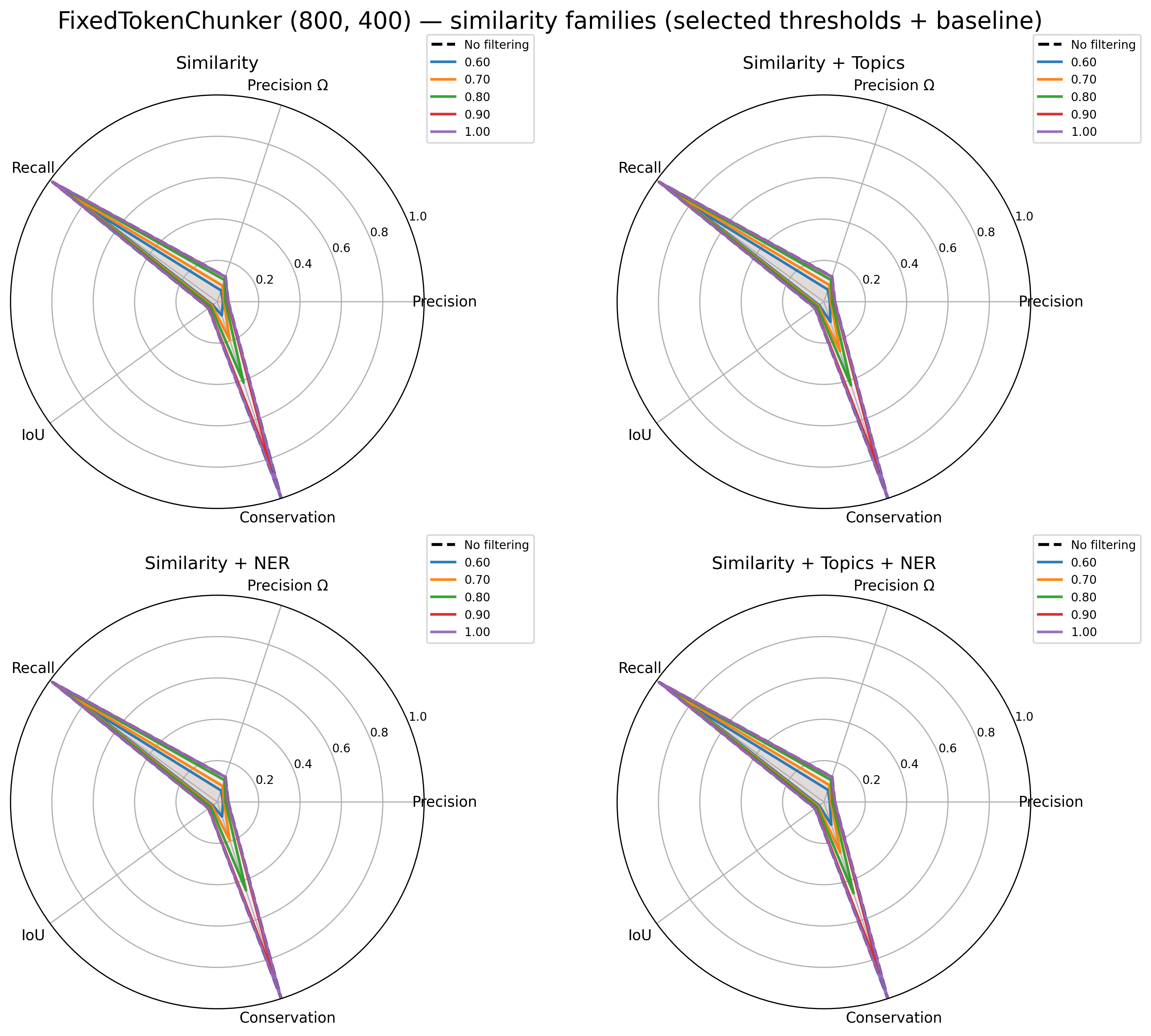}
        \caption{}
    \end{subfigure}

    \caption{FixedTokenChunker and similarity-based filtres for Chroma corpus}
    \label{chroma:fix-sim}
\end{figure}

\begin{figure}[!htbp]
    \centering

    \begin{subfigure}{0.48\textwidth}
        \includegraphics[width=\linewidth]{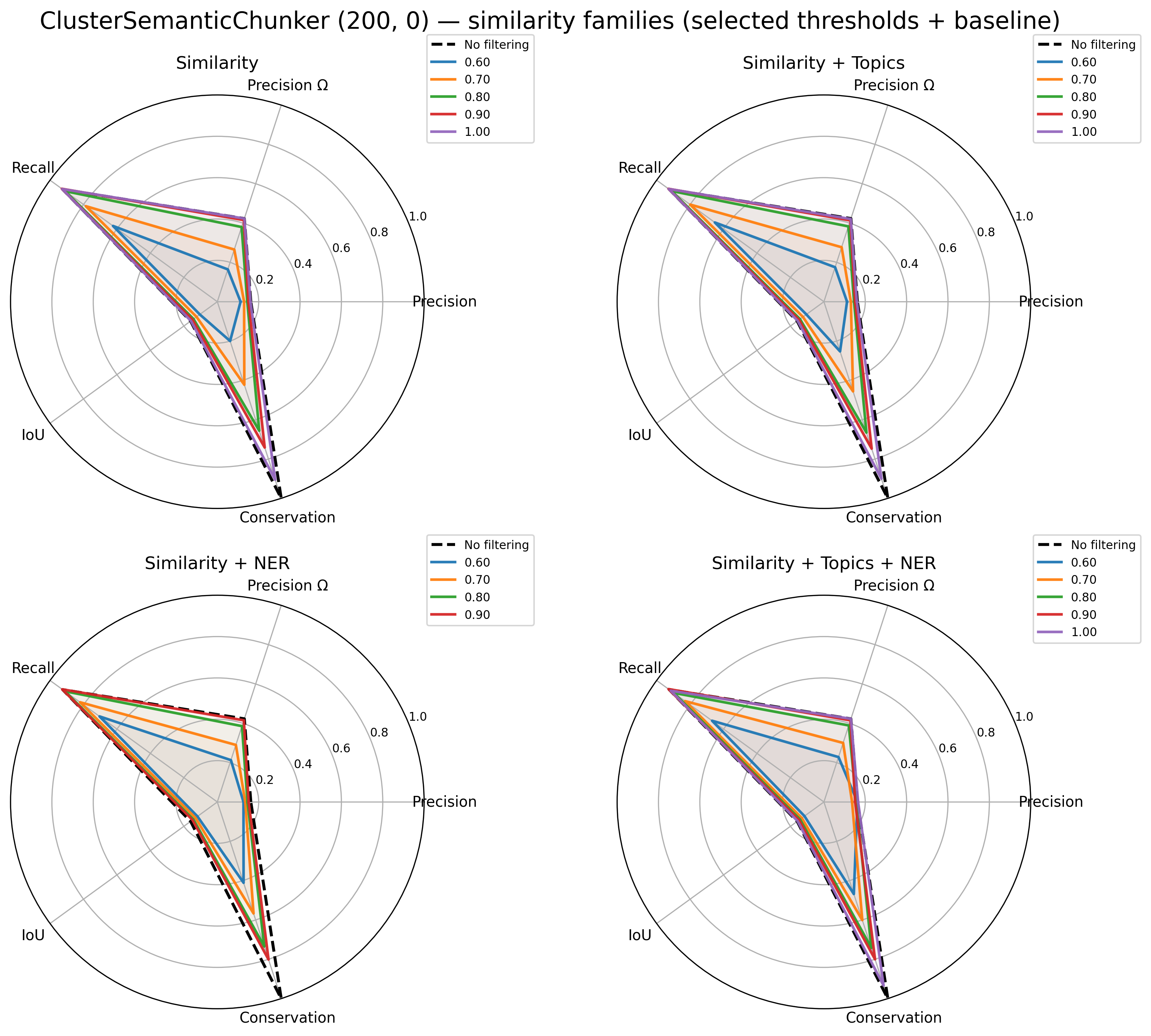}
        \caption{}
    \end{subfigure}
    \hfill
    \begin{subfigure}{0.48\textwidth}
        \includegraphics[width=\linewidth]{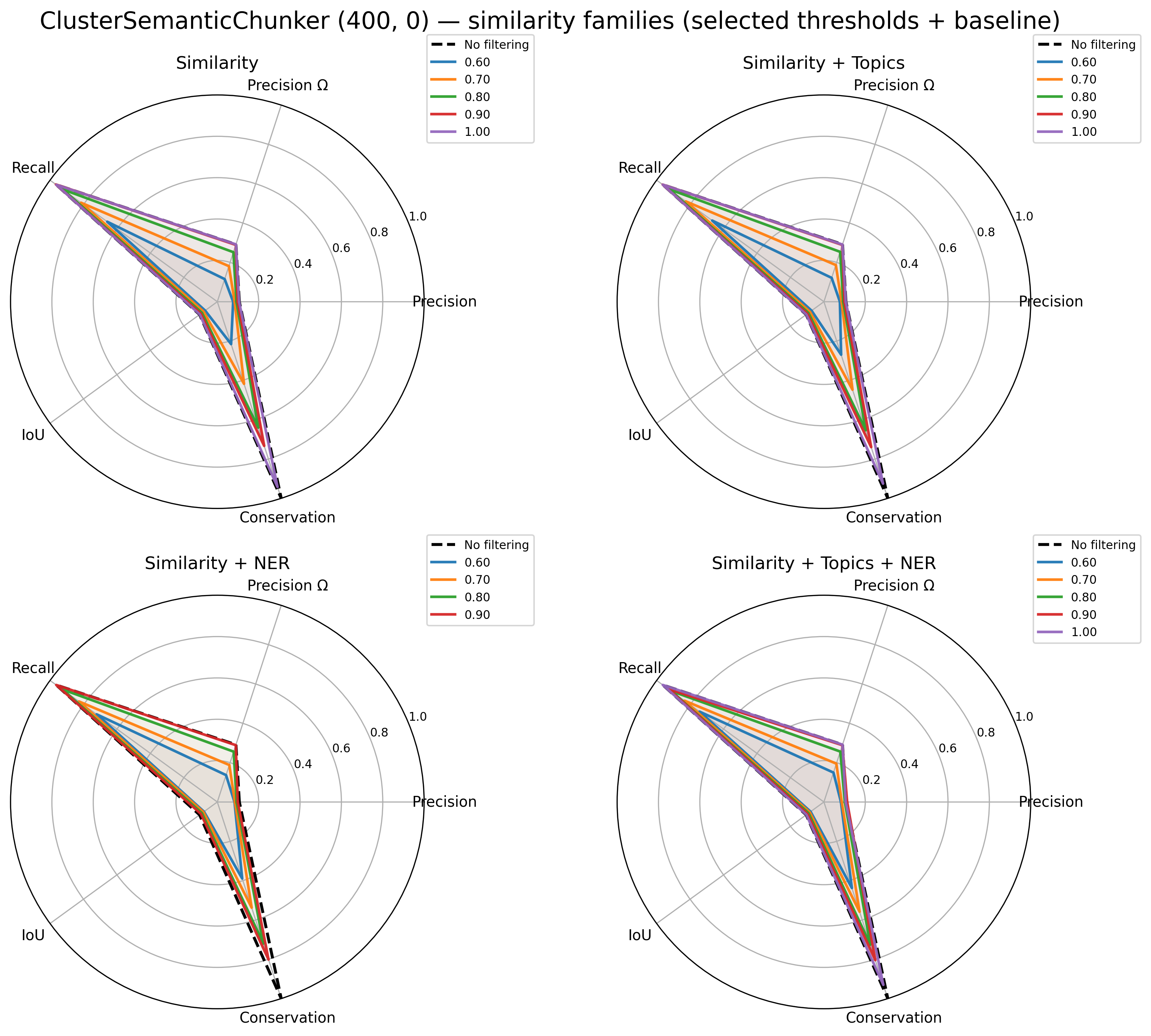}
        \caption{}
    \end{subfigure}

    \caption{ClusterSemanticChunker and similarity-based filtres for Chroma corpus}
    \label{chroma:clust-sim}
\end{figure}

Figures~\ref{chroma:rec-sim}, \ref{chroma:fix-sim}, \ref{chroma:clust-sim} report the results obtained with the similarity-based and hybrid similarity-driven filters. A common pattern appears across most chunking configurations: when filtering becomes more aggressive, the conservation ratio decreases, but recall and IoU also tend to decline. At high thresholds, the filtered configurations remain close to the no-filtering baseline. However the reduction in index size becomes limited. At lower thresholds, the index becomes more compact. Although, part of the useful information is also removed.

The effect of these filters depends on the chunking strategy. Recursive and ClusterSemantic chunkers tend to produce smoother and more stable precision--recall trade-offs, whereas several FixedToken configurations are more sensitive to threshold changes. In these fixed-window settings, small variations in filtering strength can lead to more irregular changes in precision and IoU, which suggests that locally redundant chunks are more difficult to remove without affecting retrieval quality.

Adding topic information does not systematically improve the trade-off compared with semantic similarity alone. In many configurations, the curves obtained with \emph{Similarity + Topics} remain very close to those of \emph{Similarity}, which suggests that topic information mainly structures the filtering process without fundamentally changing retrieval behavior. The hybrid variants involving NER follow the same general trend. Their motivation is conceptually strong, since shared entities provide an additional structural signal of factual redundancy, but in practice they do not consistently dominate the simpler similarity-based filters across all chunkers.

\subsubsection{NER-based filters}
\begin{figure}[!htbp]
    \centering

    \begin{subfigure}{0.48\textwidth}
        \includegraphics[width=\linewidth]{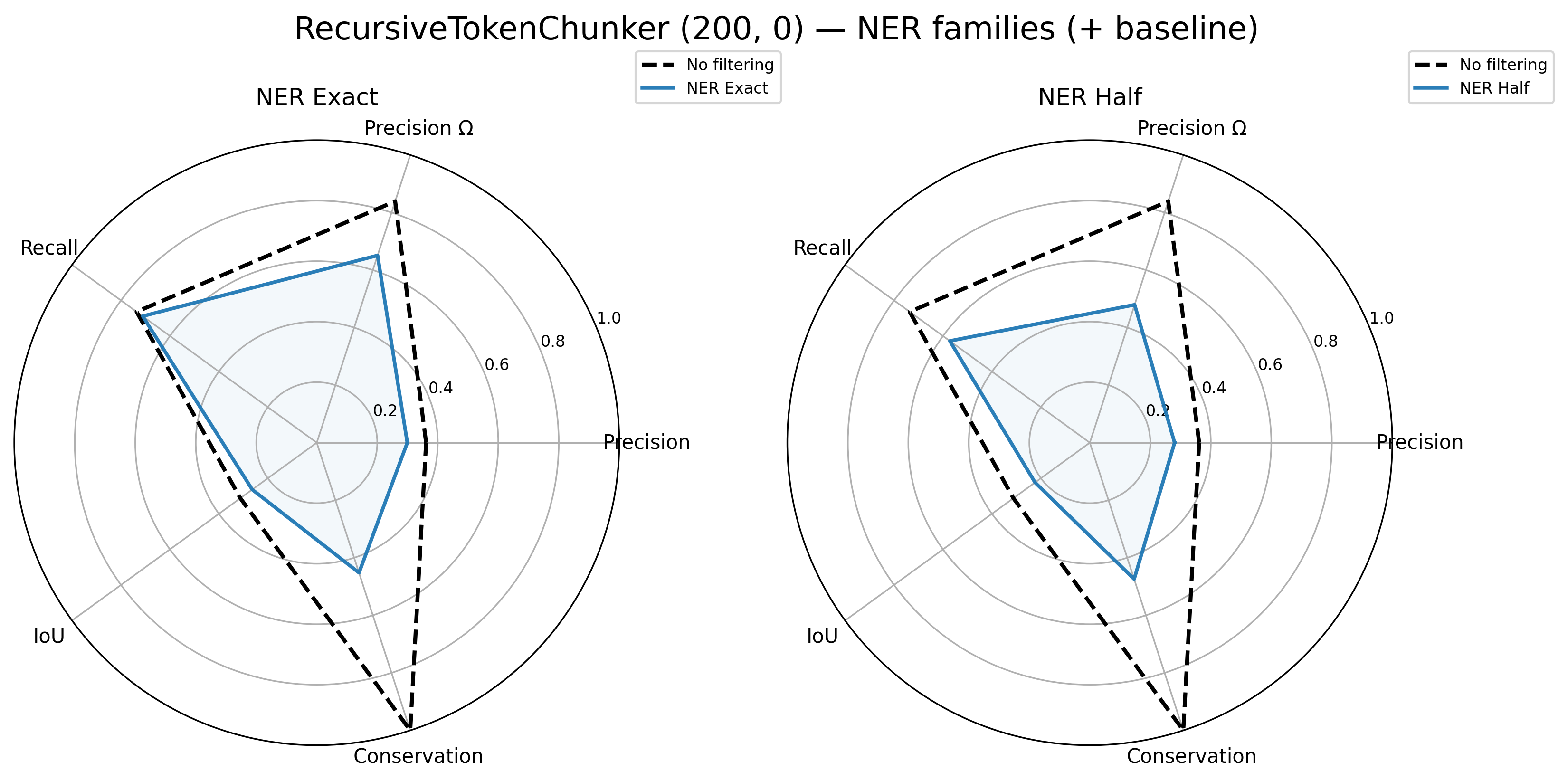}
        \caption{}
    \end{subfigure}
    \hfill
    \begin{subfigure}{0.48\textwidth}
        \includegraphics[width=\linewidth]{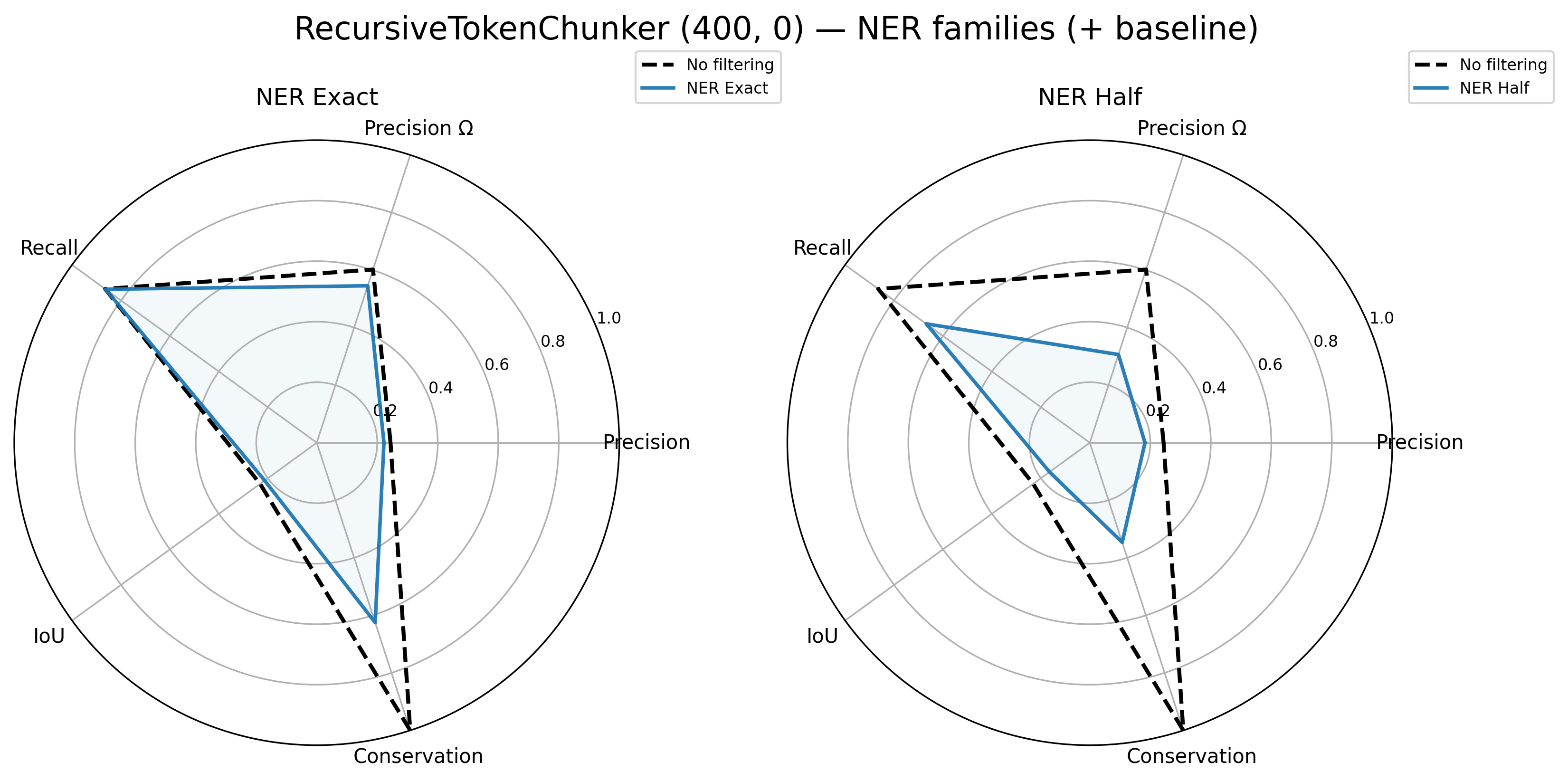}
        \caption{}
    \end{subfigure}

    \vspace{0.5em}

    \begin{subfigure}{0.48\textwidth}
        \includegraphics[width=\linewidth]{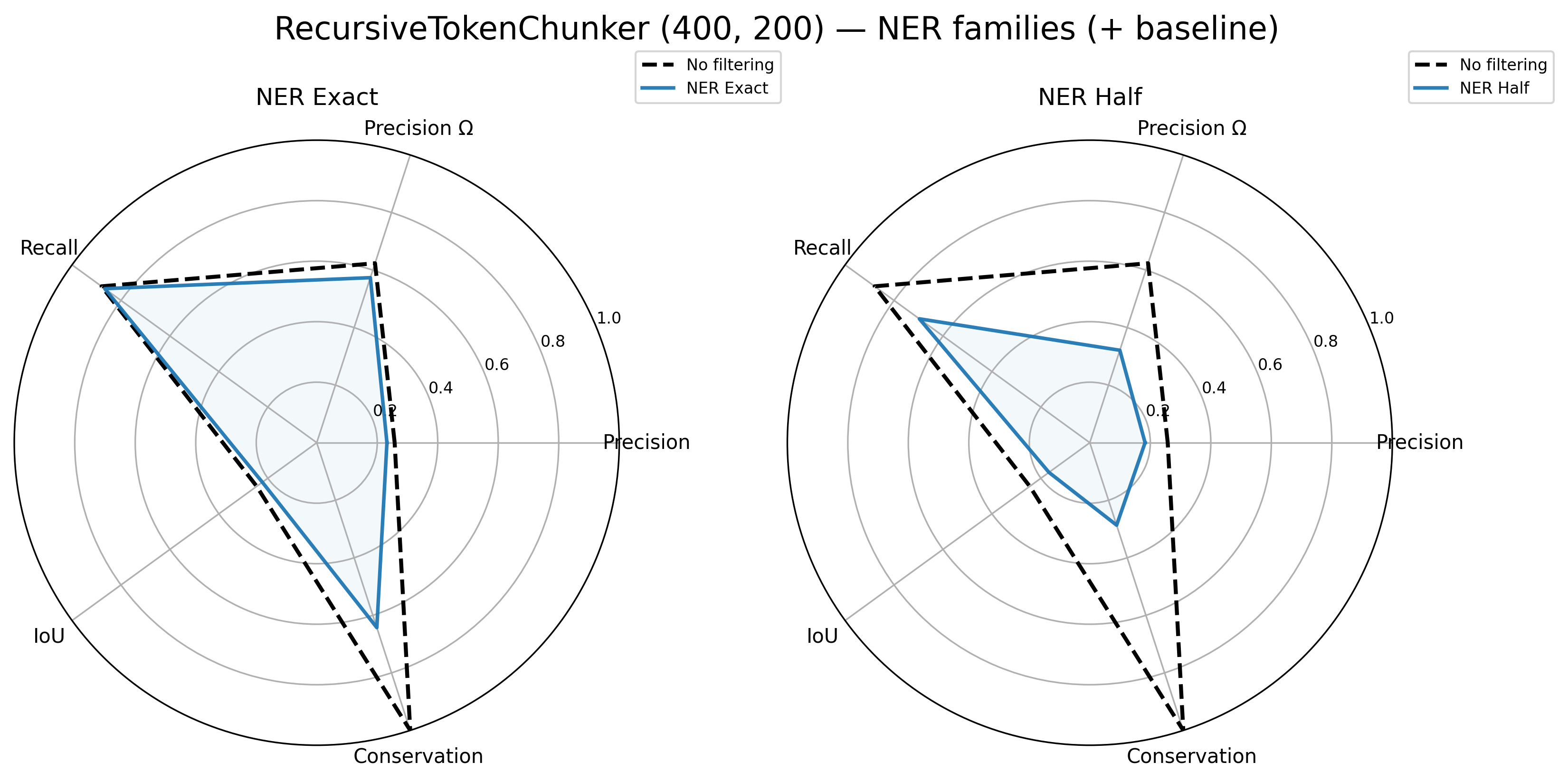}
        \caption{}
    \end{subfigure}
    \hfill
    \begin{subfigure}{0.48\textwidth}
        \includegraphics[width=\linewidth]{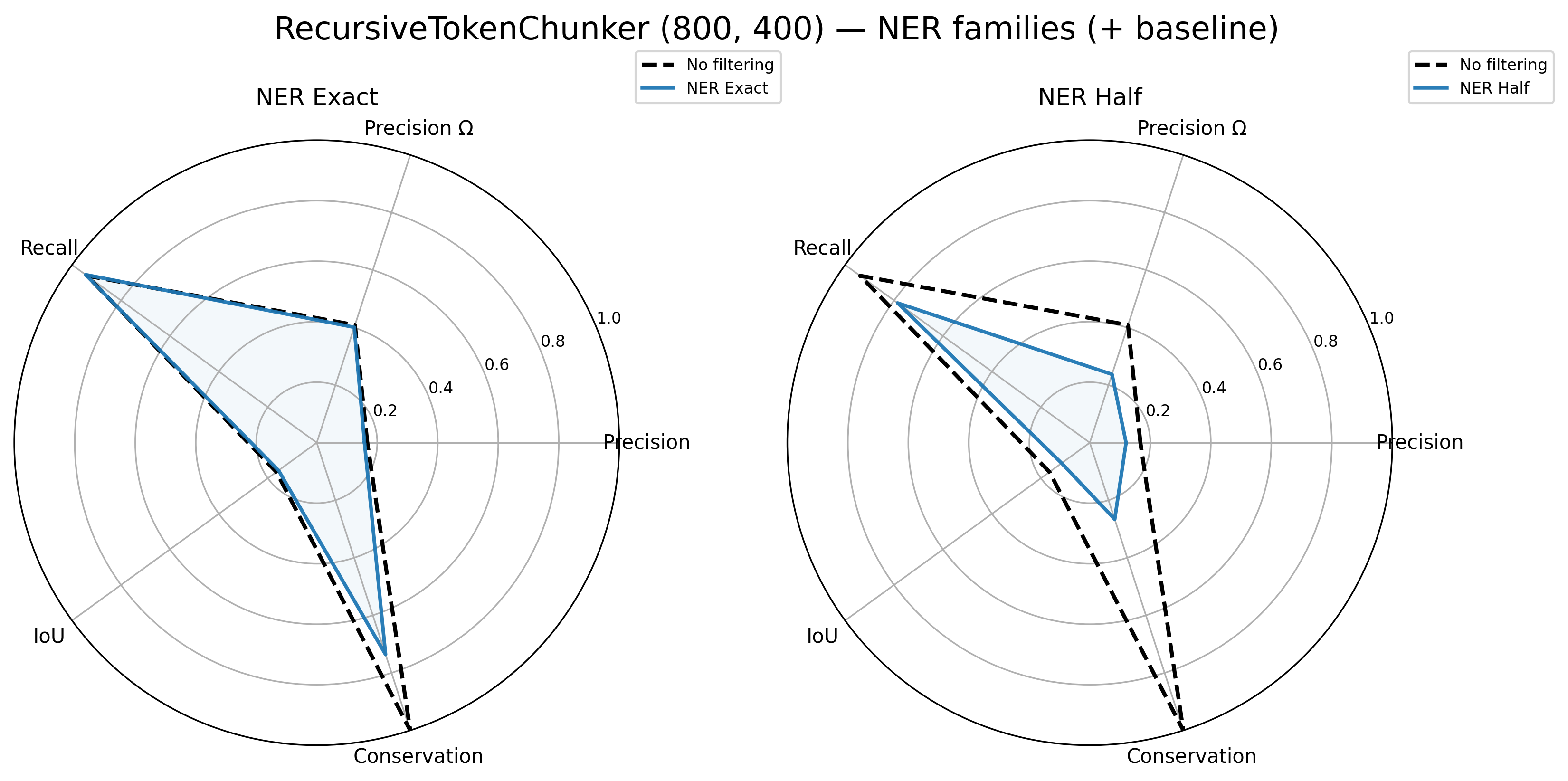}
        \caption{}
    \end{subfigure}

    \caption{RecursiveTokenChunker and NER-based filtres for Chroma corpus}
    \label{chroma:rec-ner}
\end{figure}

\begin{figure}[!htbp]
    \centering

    \begin{subfigure}{0.48\textwidth}
        \includegraphics[width=\linewidth]{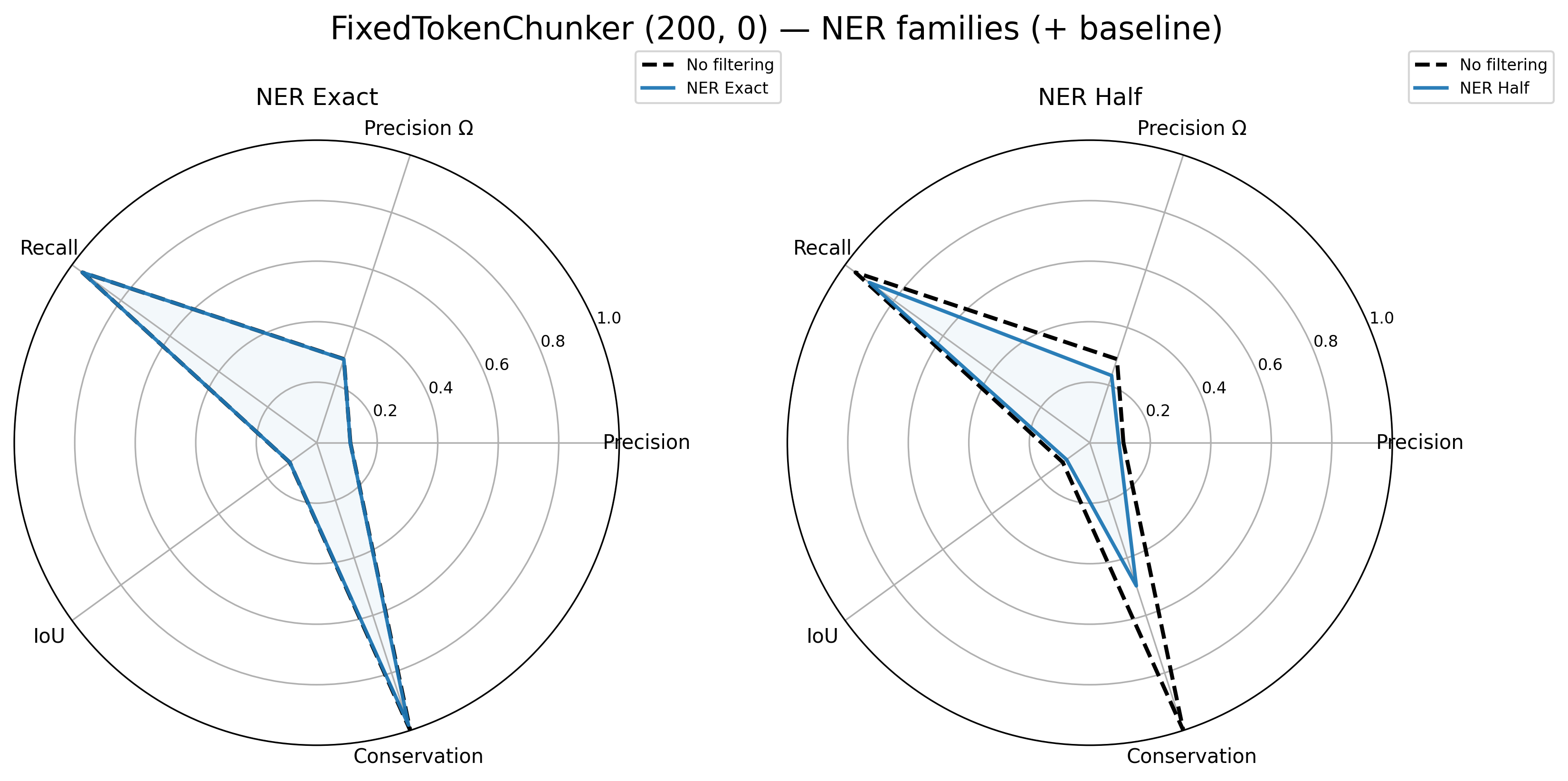}
        \caption{}
    \end{subfigure}
    \hfill
    \begin{subfigure}{0.48\textwidth}
        \includegraphics[width=\linewidth]{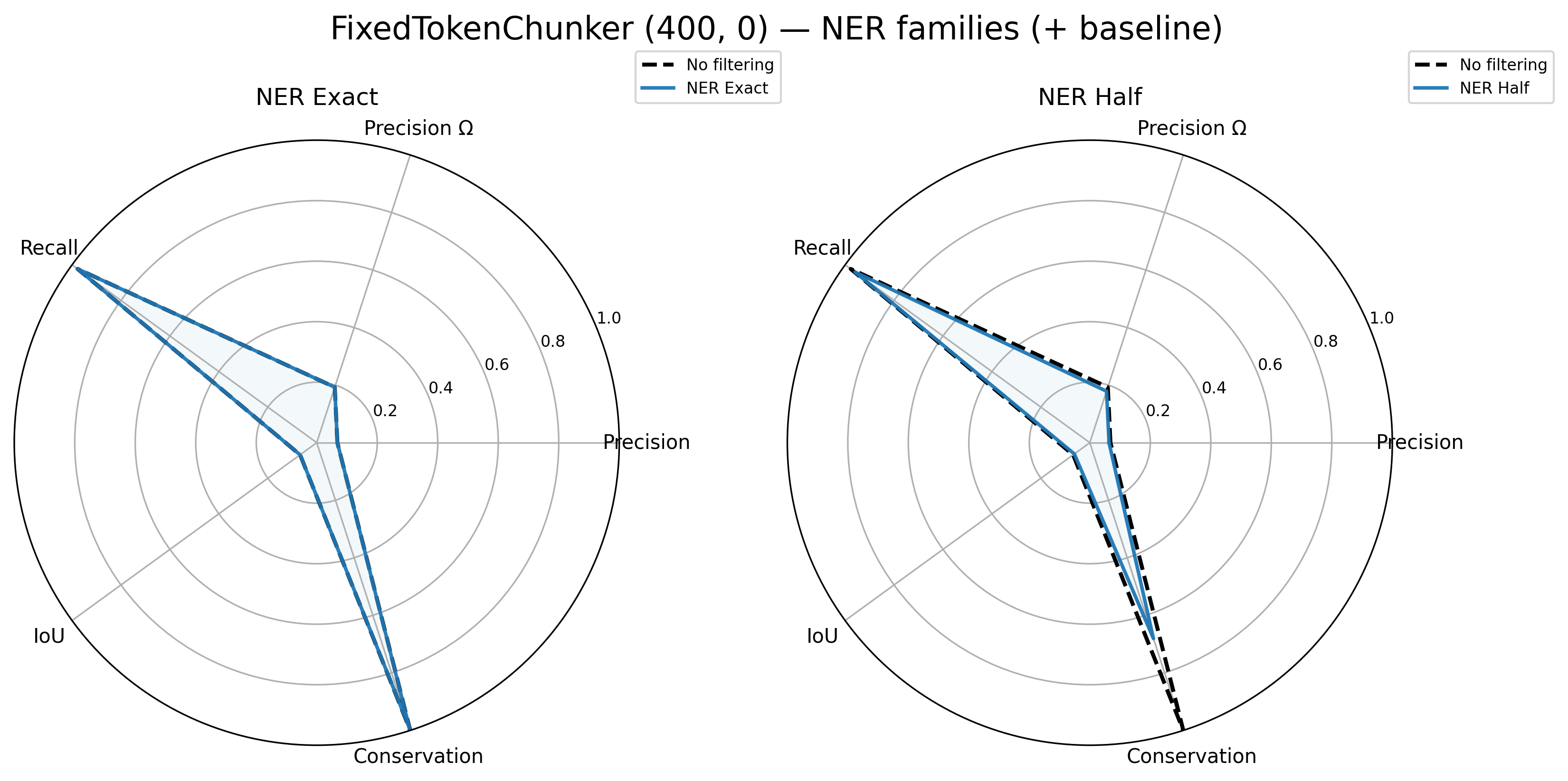}
        \caption{}
    \end{subfigure}

    \vspace{0.5em}

    \begin{subfigure}{0.48\textwidth}
        \includegraphics[width=\linewidth]{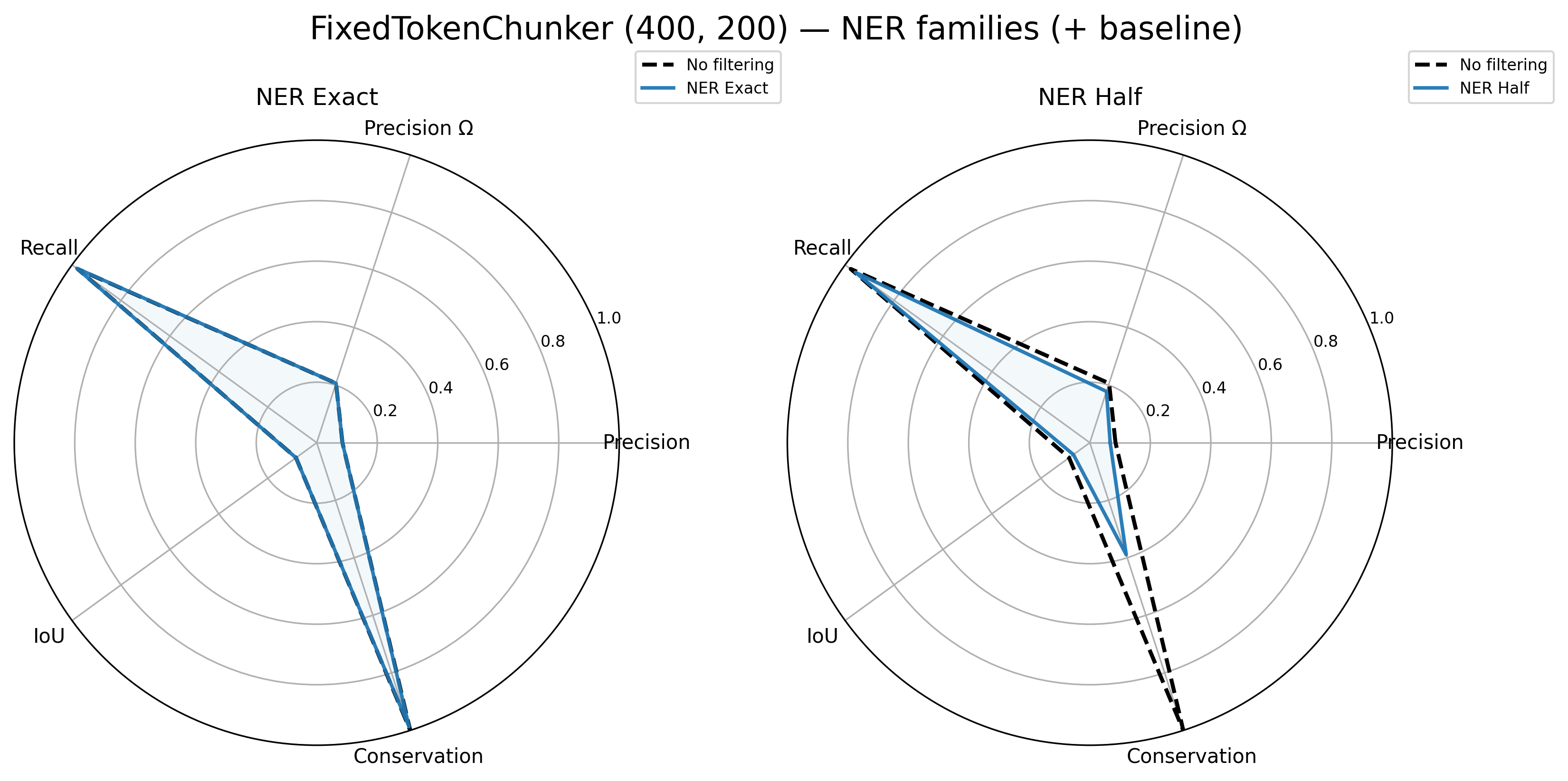}
        \caption{}
    \end{subfigure}
    \hfill
    \begin{subfigure}{0.48\textwidth}
        \includegraphics[width=\linewidth]{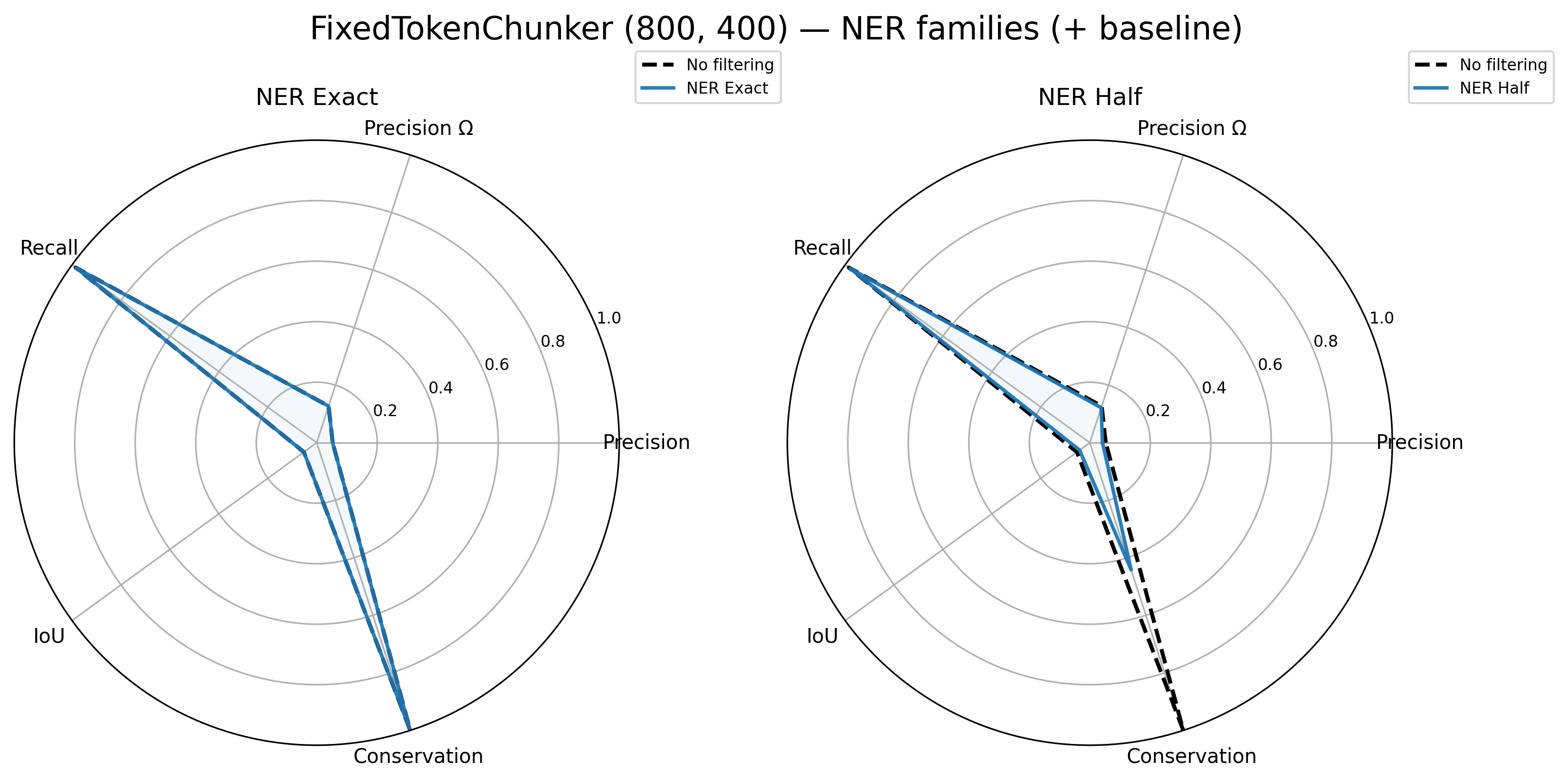}
        \caption{}
    \end{subfigure}

    \caption{FixedTokenChunker and NER-based filtres for Chroma corpus}
    \label{chroma:fix-ner}
\end{figure}

\begin{figure}[!htbp]
    \centering

    \begin{subfigure}{0.48\textwidth}
        \includegraphics[width=\linewidth]{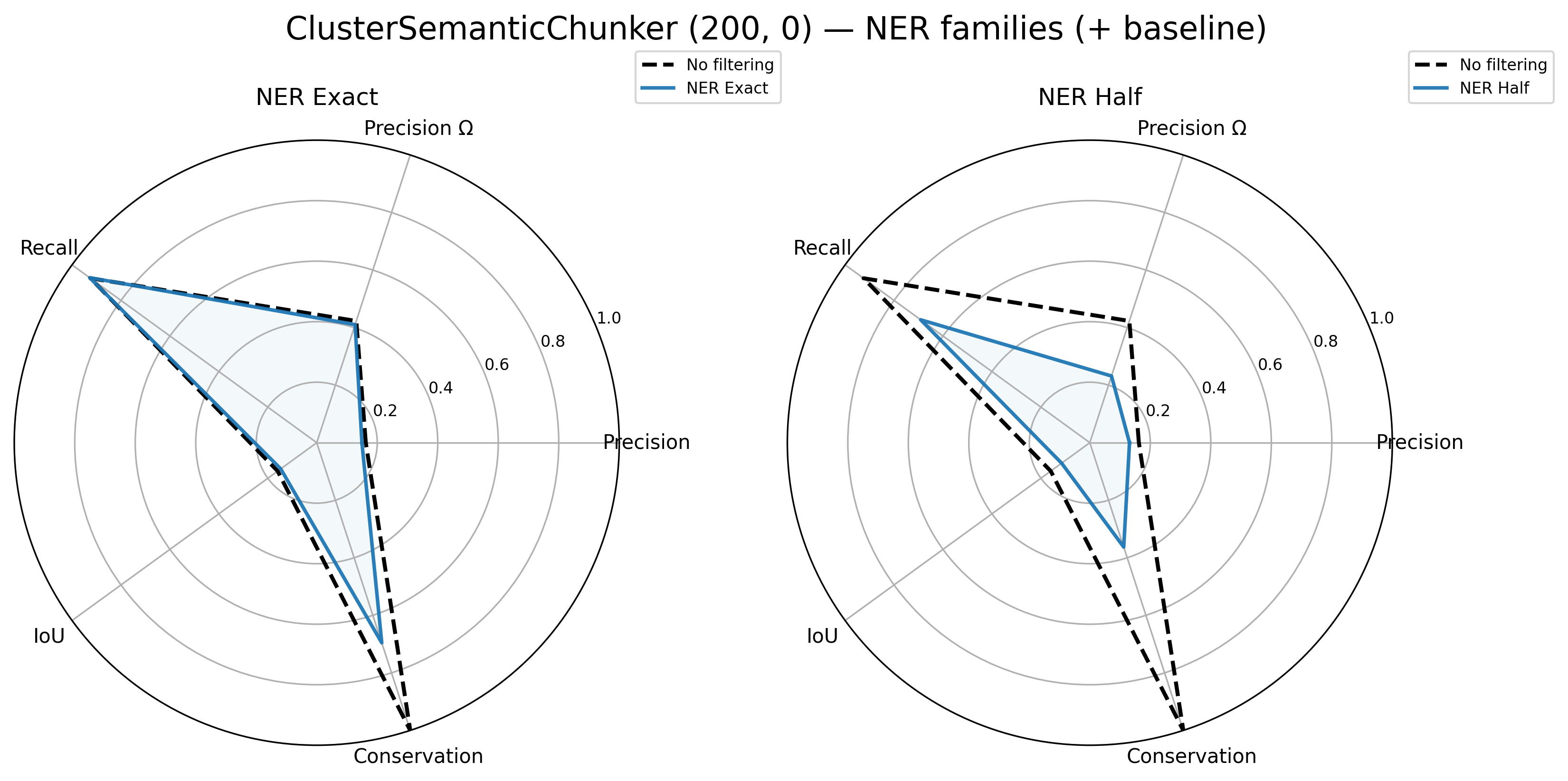}
        \caption{}
    \end{subfigure}
    \hfill
    \begin{subfigure}{0.48\textwidth}
        \includegraphics[width=\linewidth]{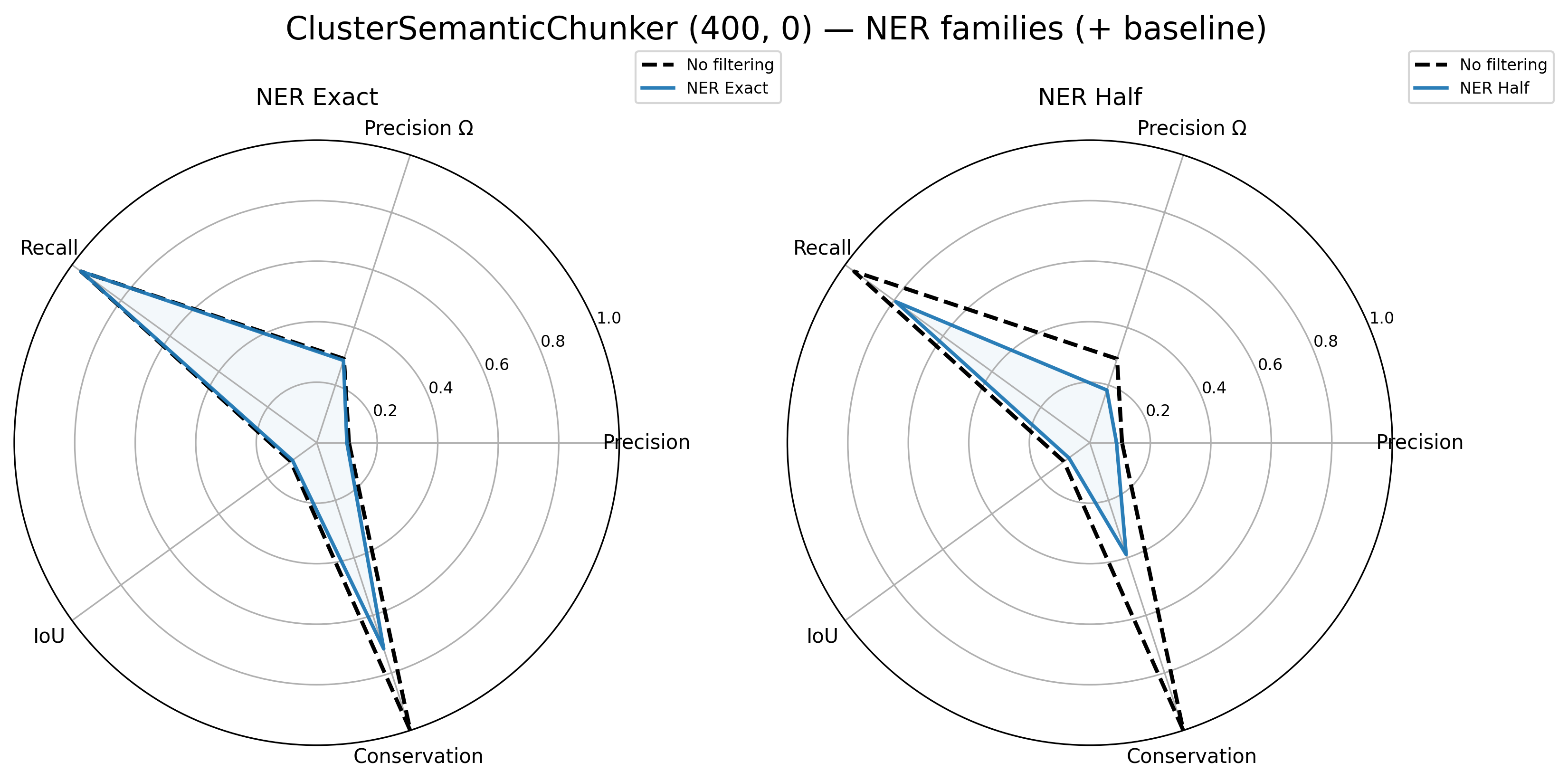}
        \caption{}
    \end{subfigure}

    \caption{ClusterSemanticChunker and NER-based filtres for Chroma corpus}
    \label{chroma:clust-ner}
\end{figure}

Figures~\ref{chroma:rec-ner}, \ref{chroma:fix-ner} and \ref{chroma:clust-ner} present the results obtained with the two entity-based strategies, \emph{NER Exact} and \emph{NER Half}. Among all methods, \emph{NER Exact} emerges as the most robust and practically useful strategy. Across several chunking configurations, it achieves a substantial reduction in index size. Moreover, it allows to maintain retrieval metrics close to the baseline.

This effect is especially visible for recursive and semantically coherent chunkers. In the \emph{Recursive 800/400} configuration, \emph{NER Exact} preserves a recall level that is almost identical to the baseline, while keeping only about 73\% of the original chunks, corresponding to a reduction of about 27\% of the index. In \emph{Recursive 400/200} and \emph{Recursive 400/0}, the conservation ratio falls to about 64\%, corresponding to an index reduction of about 36\%, while the drop in precision and IoU remains moderate. Similar behavior is observed for \emph{ClusterSemantic 400/0} and \emph{ClusterSemantic 200/0}, where the filtered configurations remain close to the baseline despite a clear reduction in the number of indexed chunks.

By contrast, \emph{NER Half} appears less stable. Although it can sometimes remove more chunks, this additional reduction is more often accompanied by a visible degradation in retrieval quality. This suggests that partial named-entity overlap is a weaker signal of true informational redundancy than exact entity matching. Two chunks may share some entities while still contributing complementary information, and this makes partial-overlap filtering more prone to removing useful passages.

\FloatBarrier
\subsubsection{Main findings}
Overall, this first benchmark shows that the effect of filtering depends strongly on the chunking strategy. Recursive and ClusterSemantic chunkers produce more stable and more predictable trade-offs, whereas FixedToken chunkers are generally more sensitive to threshold selection and local redundancy.

The main result of this benchmark is that \emph{NER Exact} provides the best global compromise between index reduction and retrieval quality. However, this advantage should beconsidered carefully. Entity-based filtering does not apply uniformly to all chunks: segments without recognized named entities are left unchanged. As a result, part of the apparent robustness of \emph{NER Exact} may come from the fact that it acts on a more restricted subset of the corpus, especially in domains with high entity density. 

Depending on the chunking configuration, it reduces the number of indexed chunks by about 27\% to 36\%. Moreover, it preserves metrics that remain close to the baseline. This behavior suggests that the exact set of named entities acts as a strong signature of factual content: chunks sharing the same entity set are often redundant from an informational point of view, even when they are not strictly identical at the surface level.

The comparison with lexical and random baselines further clarifies this result. ExactNorm and MinHash-LSH typically remove only a limited portion of the index, which suggests that redundancy introduced by chunking is not primarily driven by exact or near-exact lexical duplication. More importantly, random filtering causes a much sharper degradation in recall and IoU at comparable reduction levels, indicating that the observed trade-offs are not explained by index reduction alone. Instead, the filtering strategies studied here preserve retrieval quality because they remove chunks selectively rather than indiscriminately.

These results support the idea that redundancy in RAG indexes is not purely semantic. Structural signals such as named entities can help identify a form of redundancy that is safe to remove. In the Chroma corpora setting, this makes exact entity-based filtering the most convincing strategy among the tested methods.

\subsection{SQuAD 1.1}

The experiments in this section are conducted on the \textit{dev} partition of the SQuAD~1.1 dataset. Embeddings are generated using a Sentence Transformers model, and the retrieval protocol remains identical across all configurations: dense retrieval with top-$k = 5$ chunks and evaluation based on token coverage metrics (precision, recall, and IoU). The objective of this evaluation is to analyze how different filtering strategies affect both retrieval quality and the size of the indexed corpus.

In addition to the proposed filtering strategies, we compare the results with lexical deduplication baselines (\emph{ExactNorm} and \emph{MinHash-LSH}) and with a random filtering control. These baselines are useful for determining whether the observed retrieval behavior is due to the filtering signals themselves or merely to the reduction in index size.

\subsubsection{Similarity-based filters}

\begin{figure}[!htbp]
    \centering

    \begin{subfigure}{0.48\textwidth}
        \includegraphics[width=\linewidth]{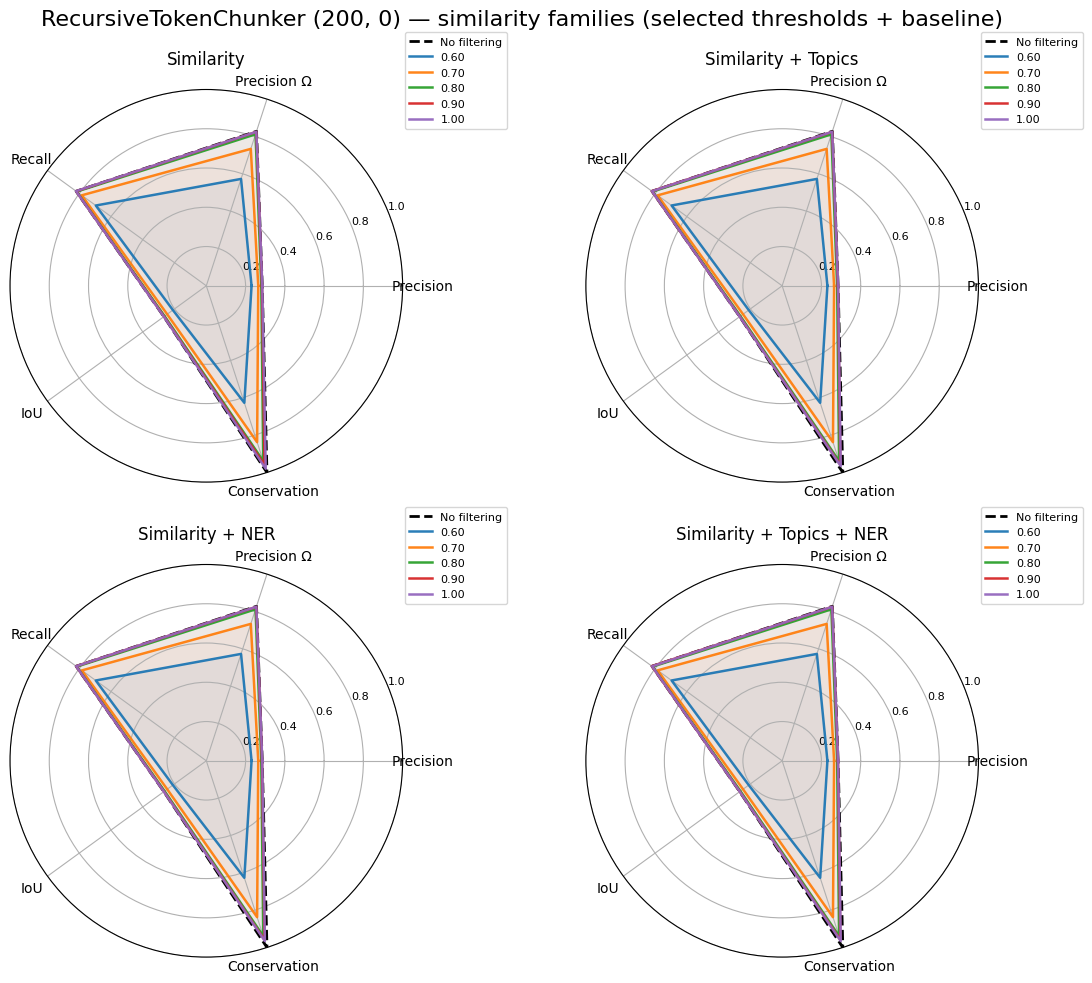}
        \caption{}
    \end{subfigure}
    \hfill
    \begin{subfigure}{0.48\textwidth}
        \includegraphics[width=\linewidth]{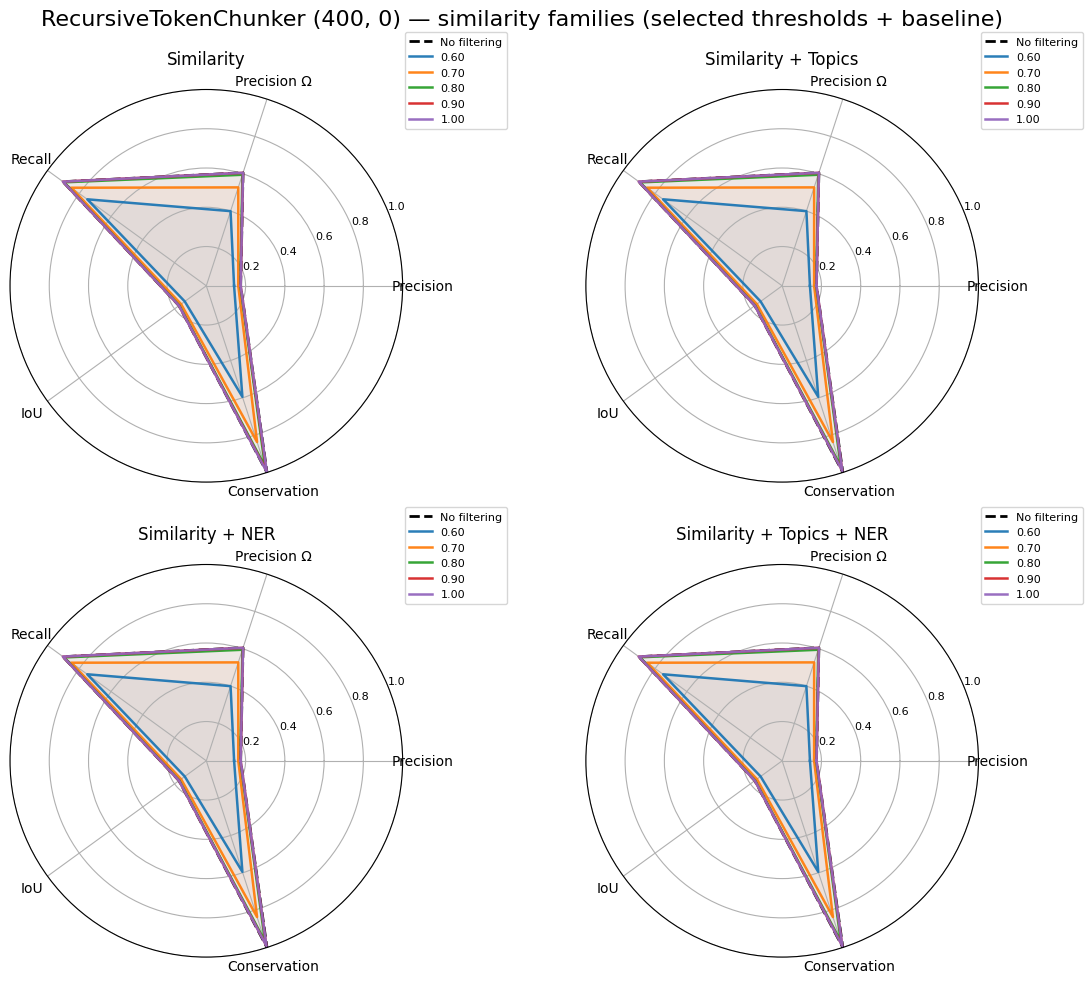}
        \caption{}
    \end{subfigure}

    \vspace{0.5em}

    \begin{subfigure}{0.48\textwidth}
        \includegraphics[width=\linewidth]{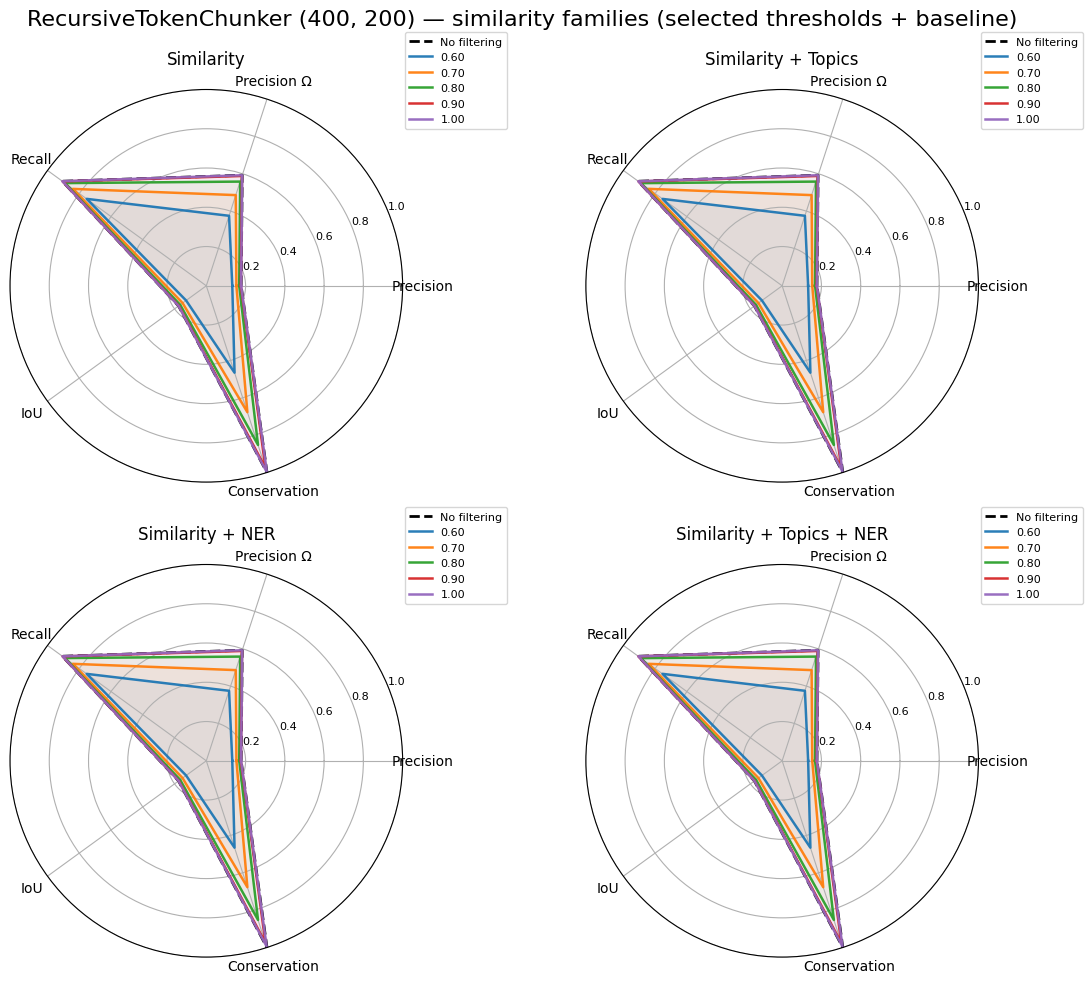}
        \caption{}
    \end{subfigure}
    \hfill
    \begin{subfigure}{0.48\textwidth}
        \includegraphics[width=\linewidth]{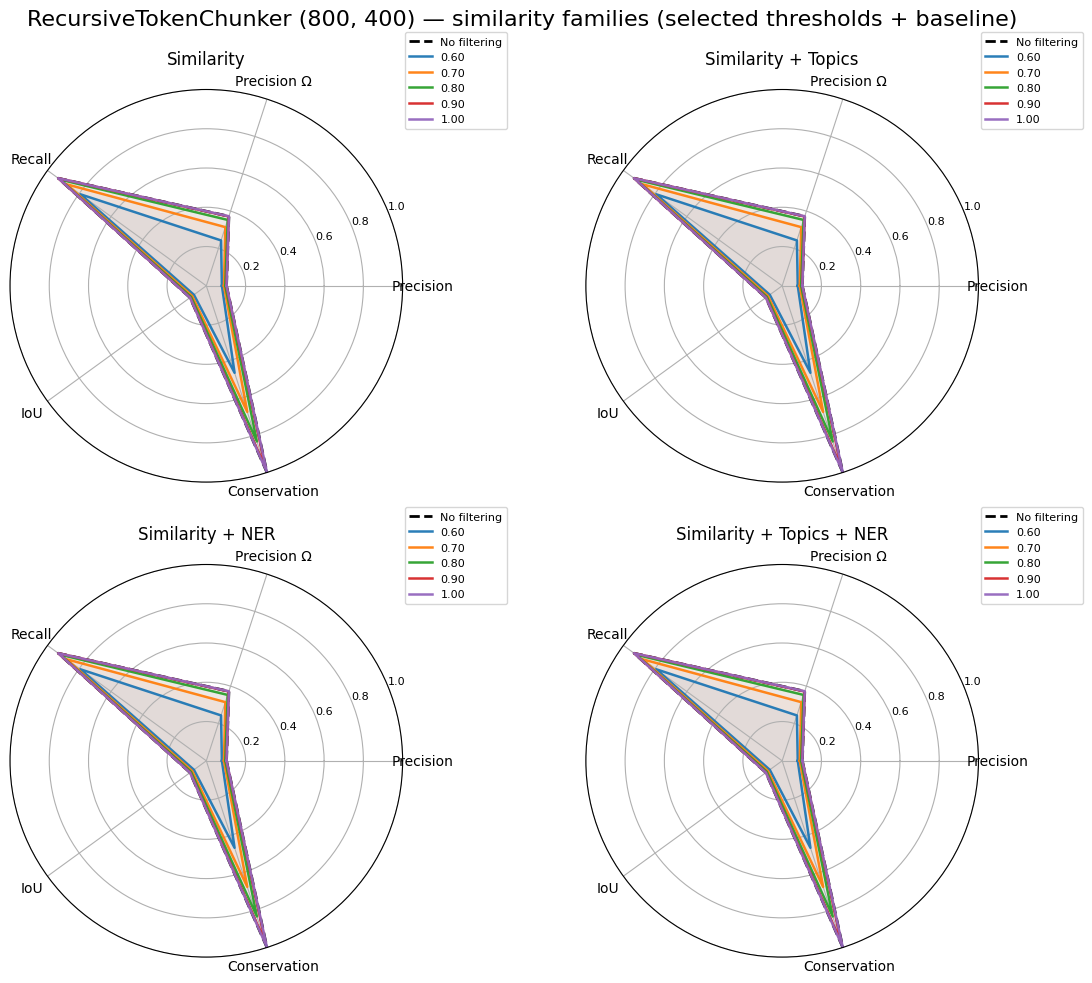}
        \caption{}
    \end{subfigure}

    \caption{RecursiveTokenChunker and similarity-based filtres for squad corpus}
    \label{squad:rec-sim}
\end{figure}

\begin{figure}[!htbp]
    \centering

    \begin{subfigure}{0.48\textwidth}
        \includegraphics[width=\linewidth]{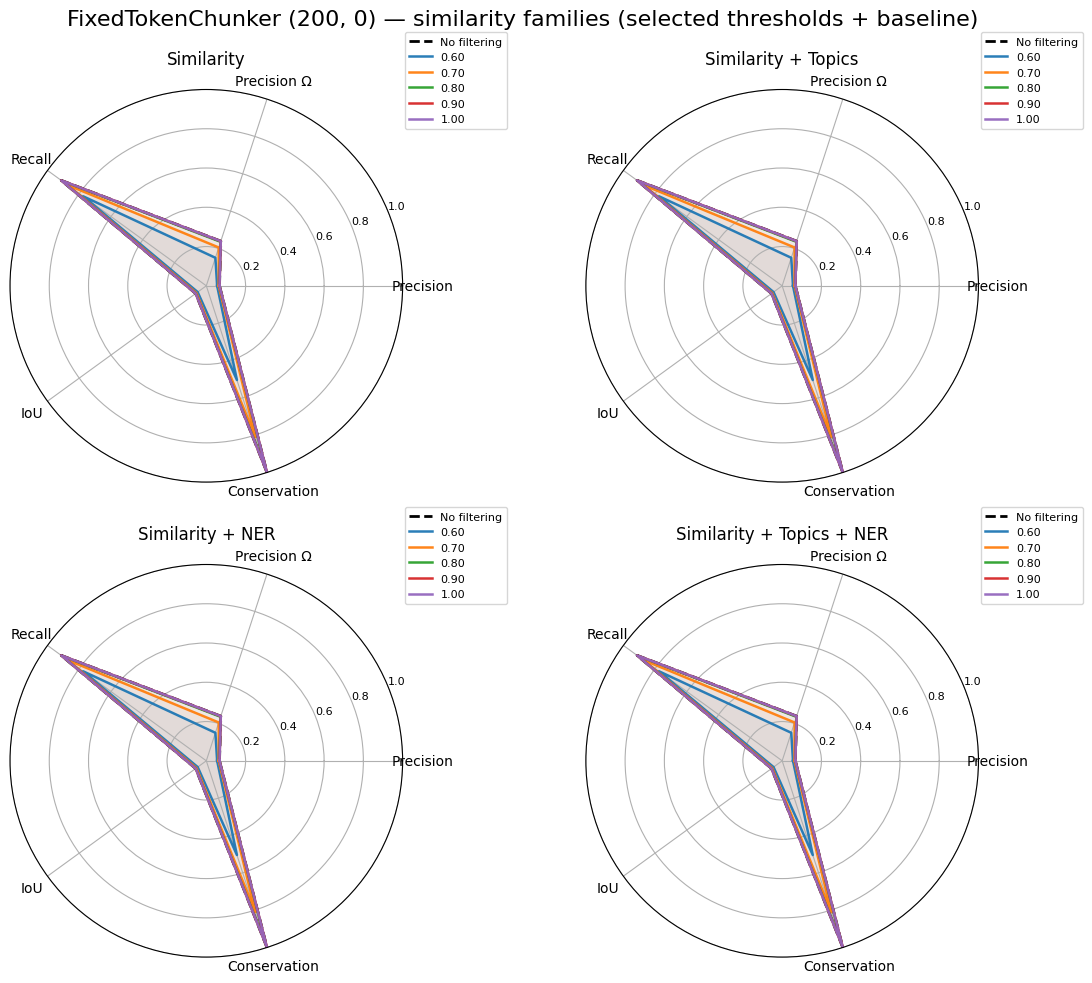}
        \caption{}
    \end{subfigure}
    \hfill
    \begin{subfigure}{0.48\textwidth}
        \includegraphics[width=\linewidth]{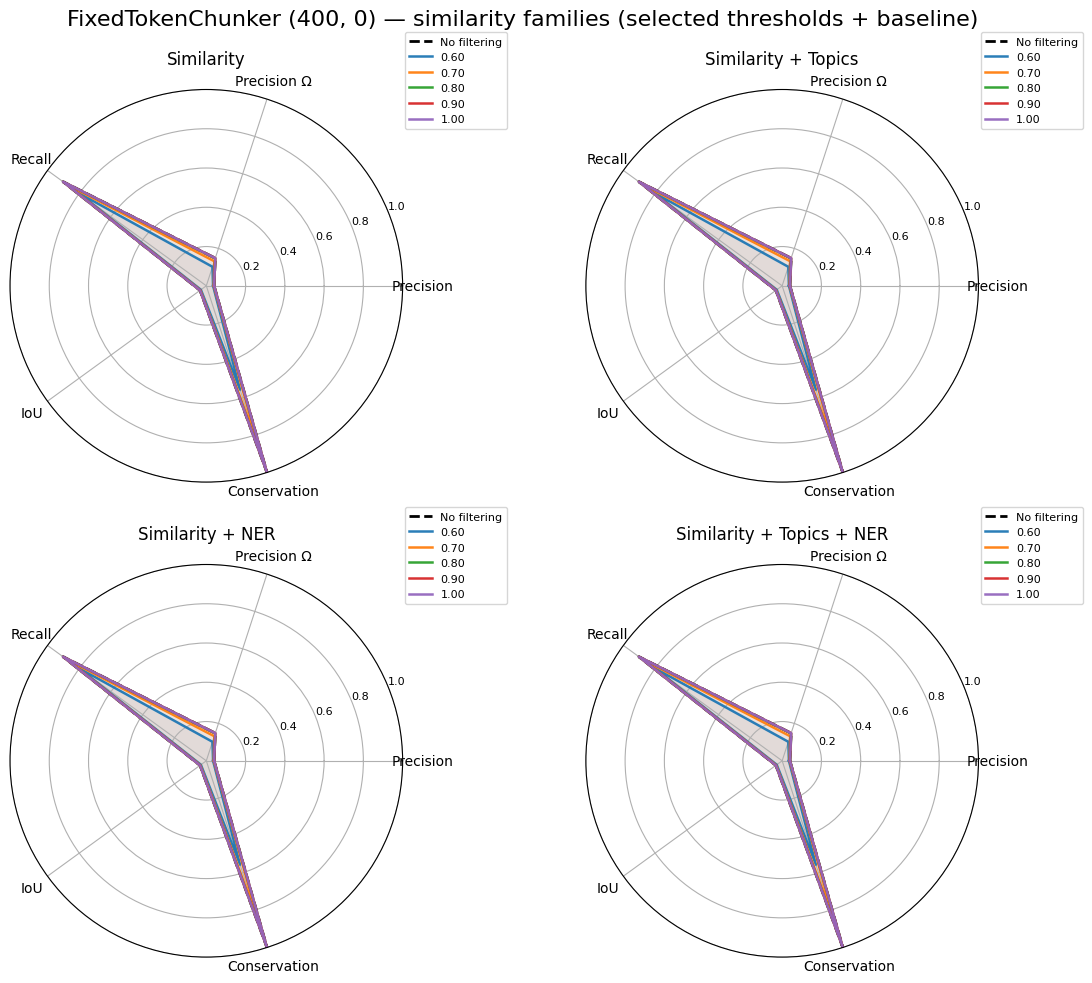}
        \caption{}
    \end{subfigure}

    \vspace{0.5em}

    \begin{subfigure}{0.48\textwidth}
        \includegraphics[width=\linewidth]{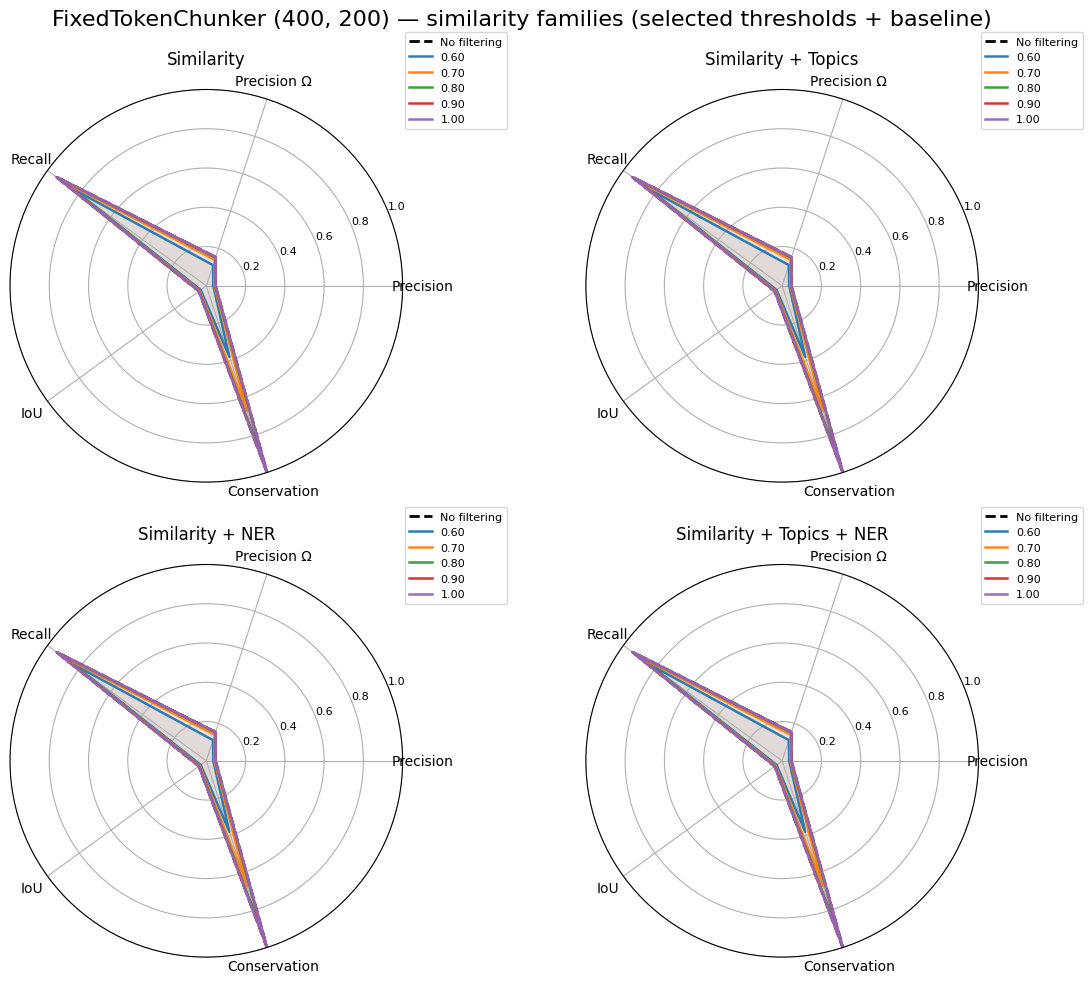}
        \caption{}
    \end{subfigure}
    \hfill
    \begin{subfigure}{0.48\textwidth}
        \includegraphics[width=\linewidth]{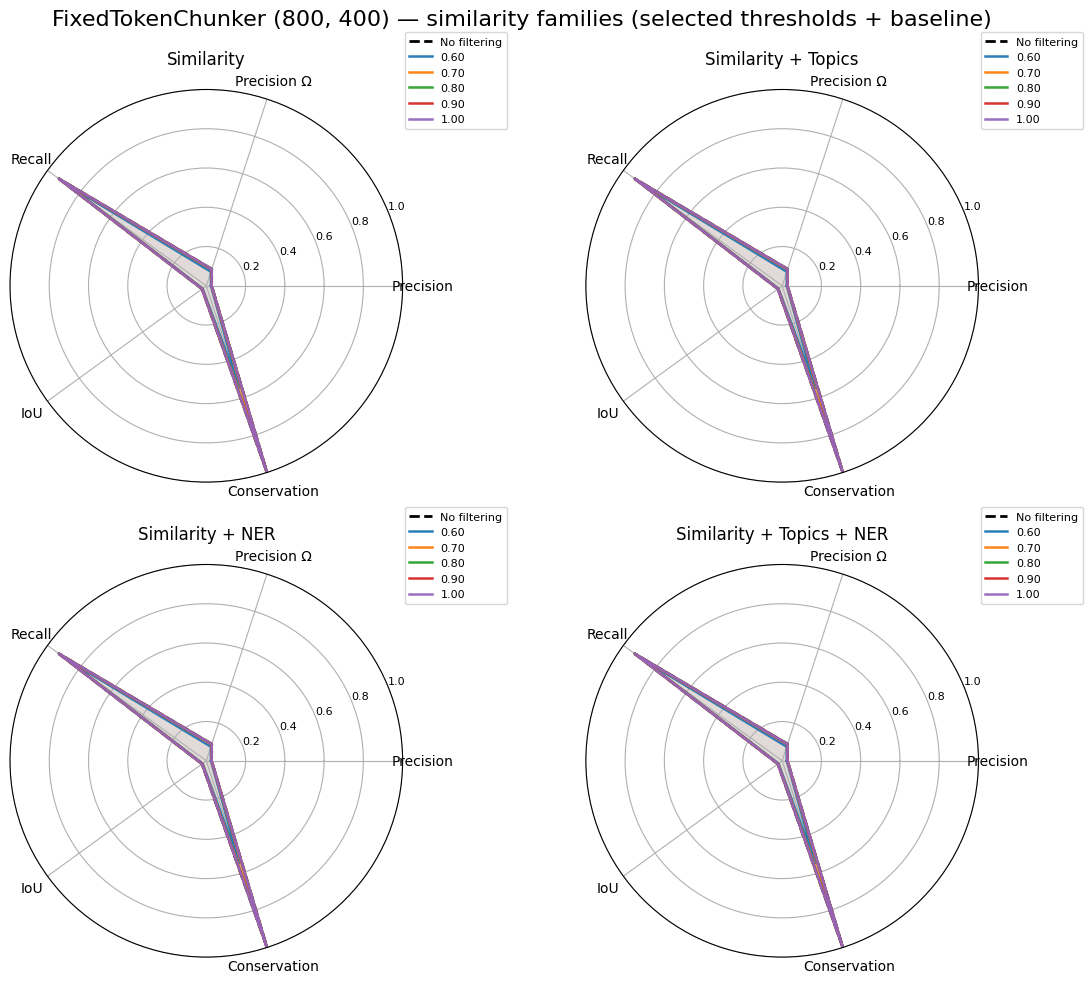}
        \caption{}
    \end{subfigure}

    \caption{FixedTokenChunker and similarity-based filtres for squad corpus}
    \label{squad:fix-sim}
\end{figure}

\begin{figure}[!htbp]
    \centering

    \begin{subfigure}{0.48\textwidth}
        \includegraphics[width=\linewidth]{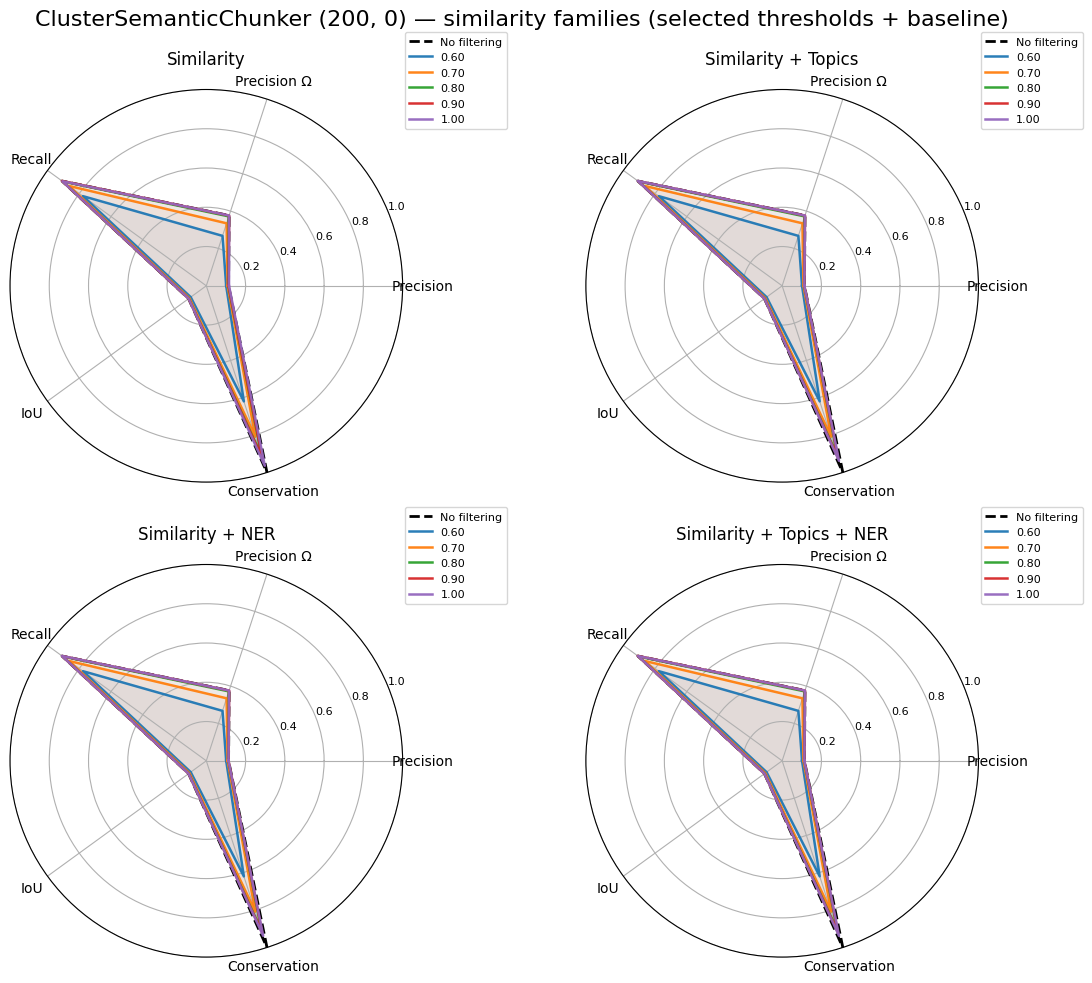}
        \caption{}
    \end{subfigure}
    \hfill
    \begin{subfigure}{0.48\textwidth}
        \includegraphics[width=\linewidth]{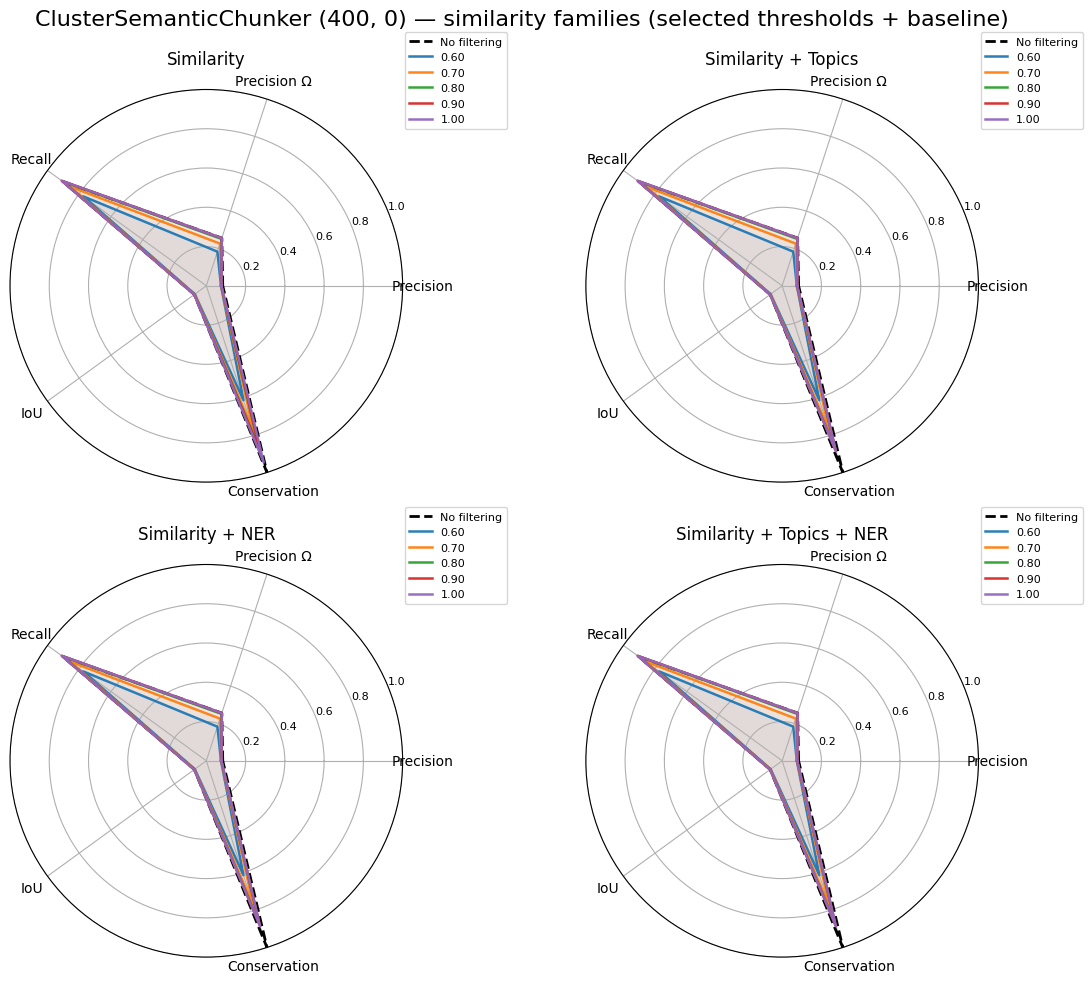}
        \caption{}
    \end{subfigure}

    \caption{ClusterSemanticChunker and similarity-based filtres for Chroma corpus}
    \label{squad:clust-sim}
\end{figure}

Figures~\ref{squad:rec-sim}, \ref{squad:fix-sim}, \ref{squad:clust-sim} present the results obtained with similarity-based and hybrid filtering strategies, including \emph{Similarity}, \emph{Similarity + Topics}, \emph{Similarity + NER}, and \emph{Similarity + Topics + NER}.

Across most configurations, filtering based solely on semantic similarity already provides a strong baseline. Increasing the similarity threshold progressively reduces the number of retained chunks while preserving relatively high recall levels. This behavior reflects the relatively homogeneous structure of the SQuAD corpus, where relevant information is often concentrated in a small number of passages.

The addition of topic-based filtering produces only limited improvements. Since most documents in SQuAD correspond to single Wikipedia articles centered on a specific subject, thematic redundancy between chunks remains relatively low. As a result, the curves obtained with \emph{Similarity + Topics} remain close to those observed with similarity filtering alone.

Hybrid strategies integrating named entities tend to produce slightly stronger filtering effects. However, the improvements remain moderate and depend on the chunking configuration. In many cases, these hybrid filters reduce the index size without significantly altering the overall precision–recall balance.

\subsubsection{NER-based filters}

\begin{figure}[!htbp]
    \centering

    \begin{subfigure}{0.48\textwidth}
        \includegraphics[width=\linewidth]{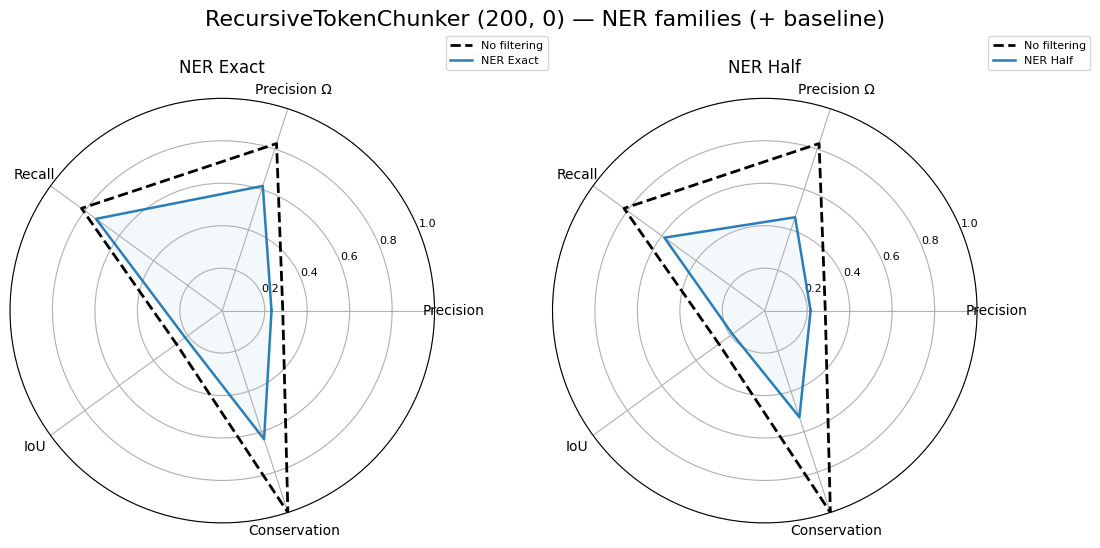}
        \caption{}
    \end{subfigure}
    \hfill
    \begin{subfigure}{0.48\textwidth}
        \includegraphics[width=\linewidth]{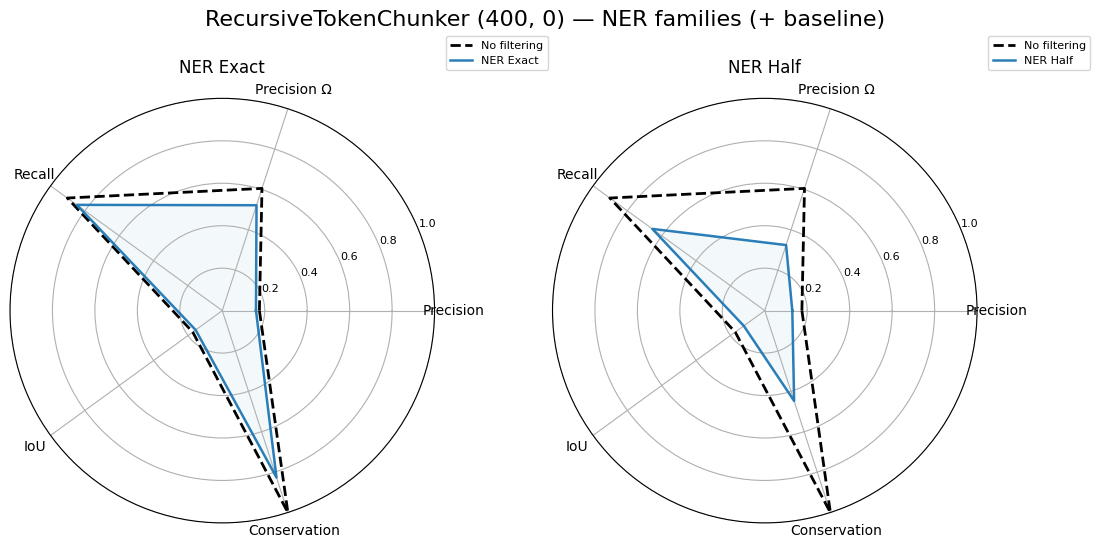}
        \caption{}
    \end{subfigure}

    \vspace{0.5em}

    \begin{subfigure}{0.48\textwidth}
        \includegraphics[width=\linewidth]{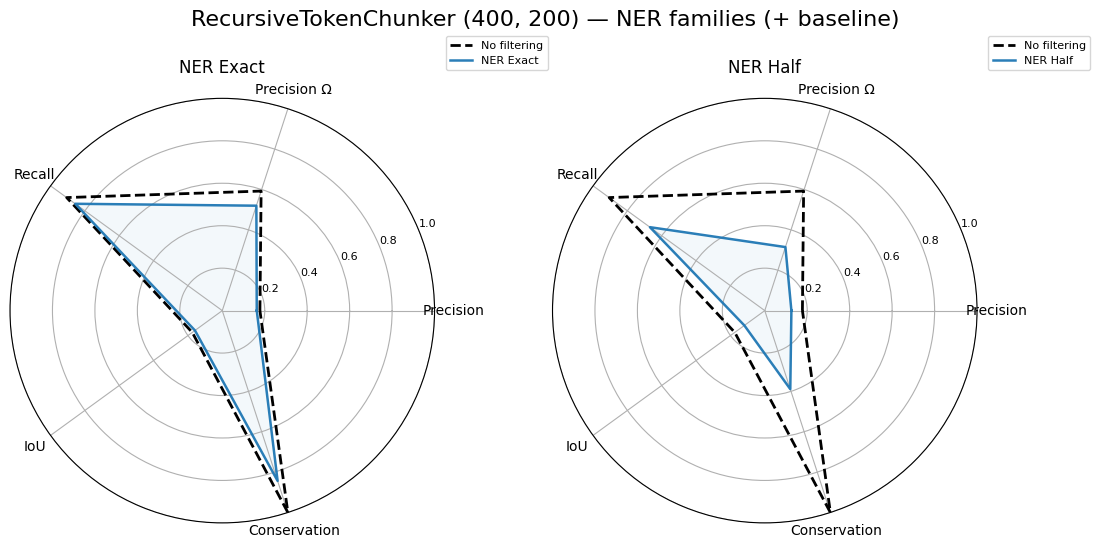}
        \caption{}
    \end{subfigure}
    \hfill
    \begin{subfigure}{0.48\textwidth}
        \includegraphics[width=\linewidth]{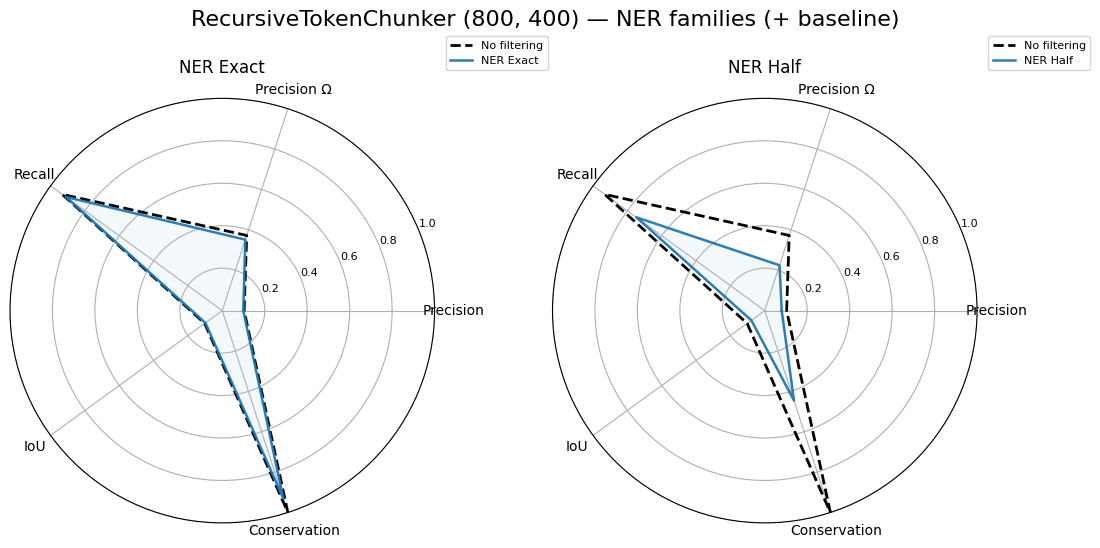}
        \caption{}
    \end{subfigure}

    \caption{RecursiveTokenChunker and NER-based filtres for squad corpus}
    \label{squad:rec-ner}
\end{figure}

\begin{figure}[!htbp]
    \centering

    \begin{subfigure}{0.48\textwidth}
        \includegraphics[width=\linewidth]{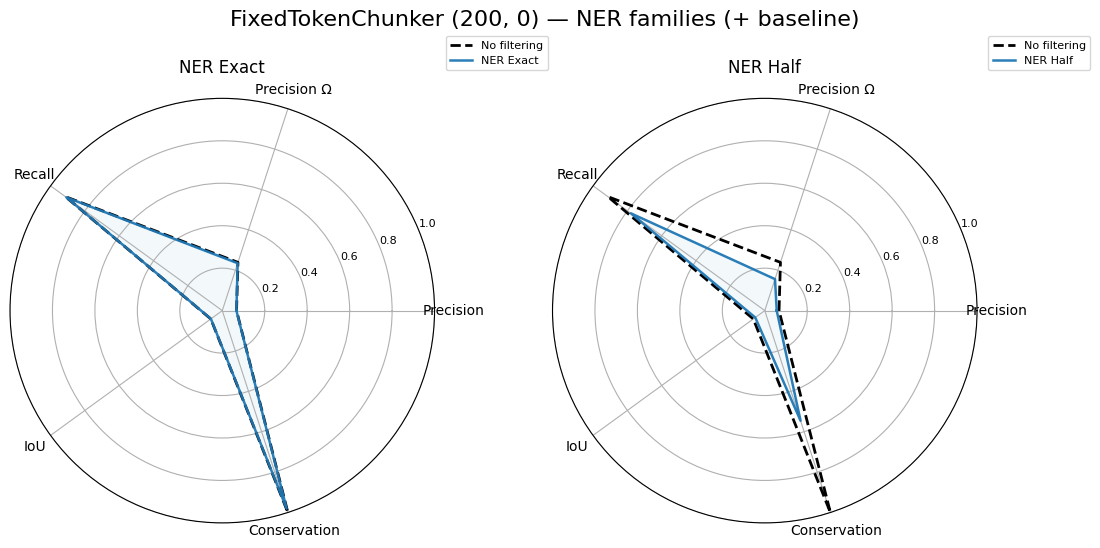}
        \caption{}
    \end{subfigure}
    \hfill
    \begin{subfigure}{0.48\textwidth}
        \includegraphics[width=\linewidth]{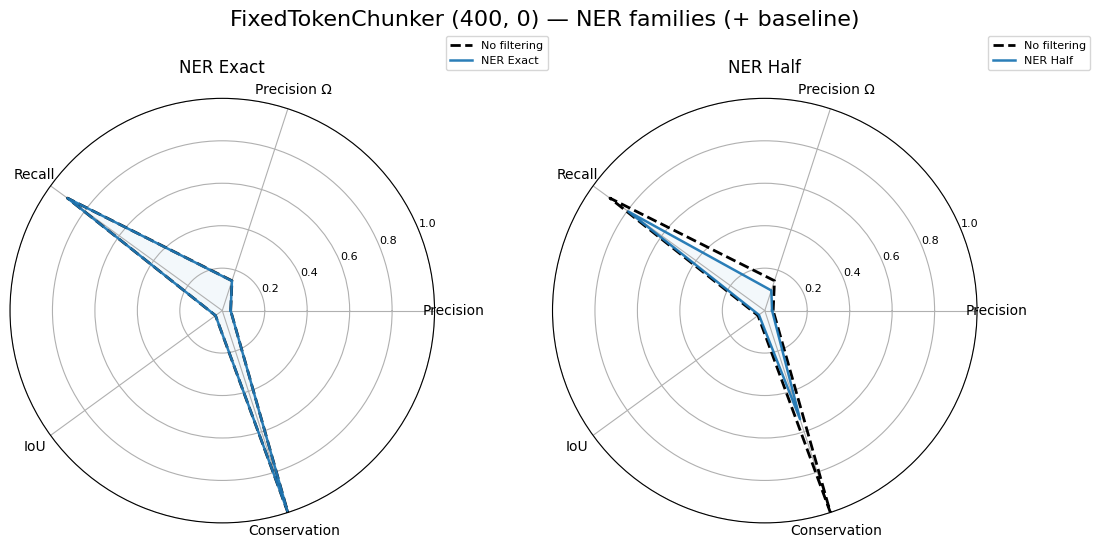}
        \caption{}
    \end{subfigure}

    \vspace{0.5em}

    \begin{subfigure}{0.48\textwidth}
        \includegraphics[width=\linewidth]{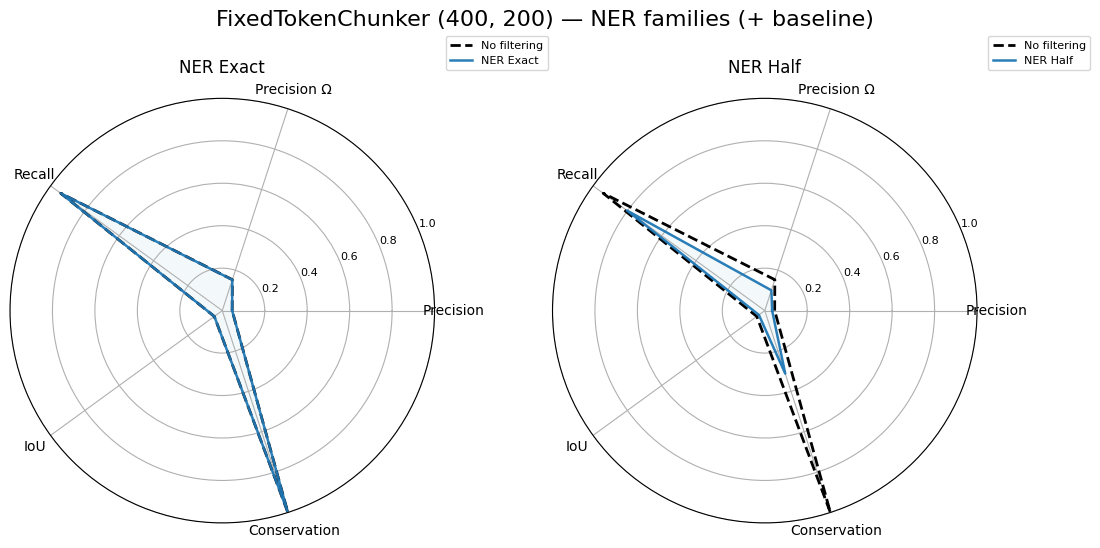}
        \caption{}
    \end{subfigure}
    \hfill
    \begin{subfigure}{0.48\textwidth}
        \includegraphics[width=\linewidth]{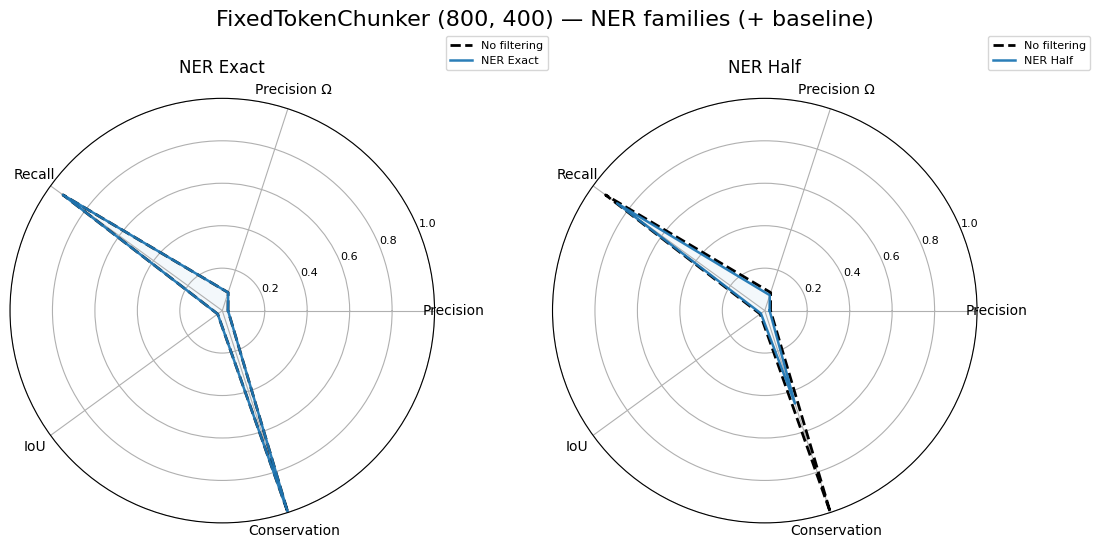}
        \caption{}
    \end{subfigure}

    \caption{FixedTokenChunker and NER-based filtres for squad corpus}
    \label{squad:fix-ner}
\end{figure}

\begin{figure}[!htbp]
    \centering

    \begin{subfigure}{0.48\textwidth}
        \includegraphics[width=\linewidth]{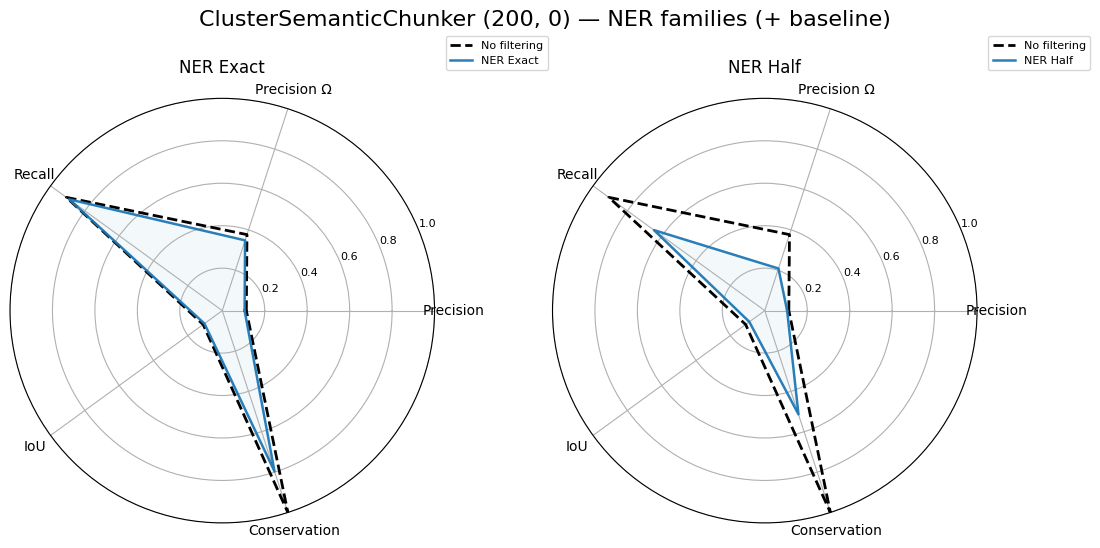}
        \caption{}
    \end{subfigure}
    \hfill
    \begin{subfigure}{0.48\textwidth}
        \includegraphics[width=\linewidth]{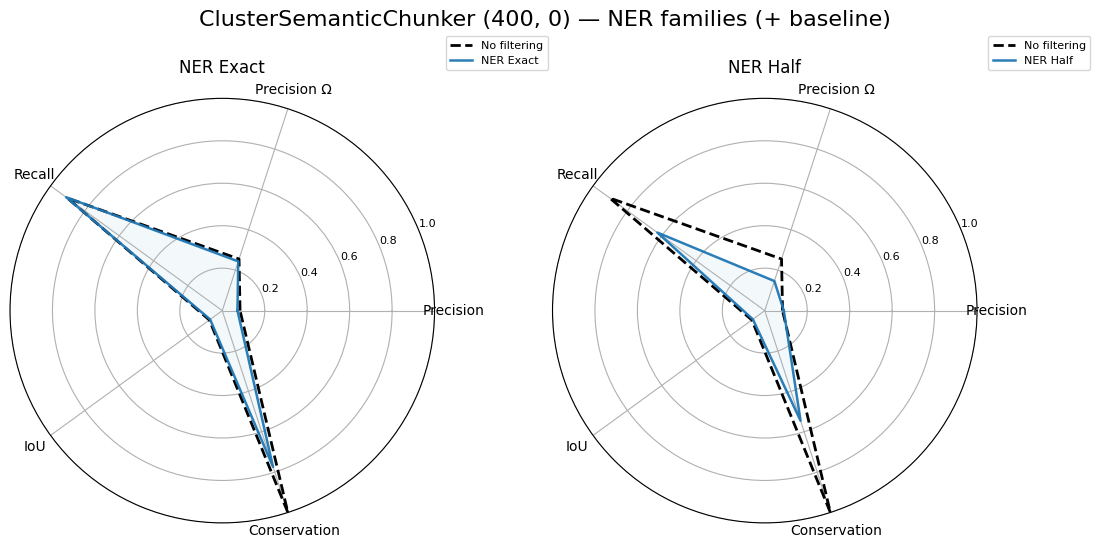}
        \caption{}
    \end{subfigure}

    \caption{ClusterSemanticChunker and NER-based filtres for squad corpus}
    \label{squad:clust-ner}
\end{figure}

Figures~\ref{squad:rec-ner}, \ref{squad:fix-ner} and \ref{squad:clust-ner} present the results obtained with the two entity-based filtering strategies: \emph{NER Exact} and \emph{NER Half}. These approaches rely exclusively on the overlap of named entities between chunks to detect potential redundancy.

In contrast to the previous benchmark on heterogeneous corpora, entity-based filtering appears more aggressive in the SQuAD setting. While both strategies substantially reduce the number of indexed chunks, they more frequently lead to a decrease in token coverage. This behavior reflects the nature of the SQuAD dataset: relevant information is not always directly associated with explicit named entities but may instead rely on descriptive phrases or contextual relations within the text.

As a consequence, strict entity-based deduplication can occasionally remove passages that still contain useful contextual information for answering the query. The \emph{NER Half} strategy, which relies on partial entity overlap, tends to amplify this effect by removing additional chunks that may still provide complementary information.

\subsubsection{Main findings}

Overall, the experiments on SQuAD reveal a more stable behavior of filtering strategies compared with the heterogeneous corpora used in the previous benchmark. Because the dataset consists of well-structured Wikipedia articles and extractive question–answer pairs, semantic similarity filtering alone provides a strong balance between index reduction and retrieval performance.

The comparison with lexical and random baselines is particularly informative in this setting. ExactNorm and MinHash-LSH produce only limited changes in both index size and retrieval metrics, which suggests that redundancy in SQuAD is only weakly lexical. By contrast, random filtering causes a much stronger decrease in recall and IoU for comparable reduction ratios. This confirms that the proposed filtering methods do not benefit merely from removing chunks, but from removing chunks in a structured way that better preserves access to relevant information.

Topic-based filtering contributes only marginally in this context, while entity-based filtering produces more pronounced reductions in index size but at the cost of a slightly larger loss in token coverage. Hybrid approaches combining similarity, topics, and NER mitigate this effect to some extent but do not consistently outperform similarity-based filtering alone.

Across all tested strategies, the size of the index can typically be reduced by approximately 20\% to 35\%. However, the comparison with random filtering shows that not all reductions are equivalent: structured filtering preserves retrieval quality substantially better than indiscriminate chunk removal. In the best cases, recall drops by only about 2\% to 6\%, precision remains relatively stable with variations below 3\%, and IoU decreases by roughly 3\% to 8\%. These results indicate that a substantial portion of redundant chunks can be removed while preserving most of the relevant information required for effective retrieval.

\subsection{WebFAQ}
The experiments in this section are conducted on a multilingual subset of the WebFAQ dataset restricted to long documents, in order to highlight the effects of chunking and filtering in realistic RAG scenarios involving large and heterogeneous corpora. Embeddings are generated using a Sentence Transformers model, and the retrieval protocol remains identical to the one used for SQuAD: dense retrieval with top-$k = 5$ chunks and evaluation based on token coverage metrics (precision, recall, and IoU). The figures presented below compare the impact of different filtering strategies on retrieval performance and index reduction across all languages combined.

As in the previous sections, we compare the proposed filtering strategies with lexical deduplication baselines and with random filtering. This comparison is especially important in a multilingual setting, where lexical overlap and semantic similarity may behave differently across languages.

\subsubsection{Similarity-based filters}

\begin{figure}[!htbp]
    \centering

    \begin{subfigure}{0.48\textwidth}
        \includegraphics[width=\linewidth]{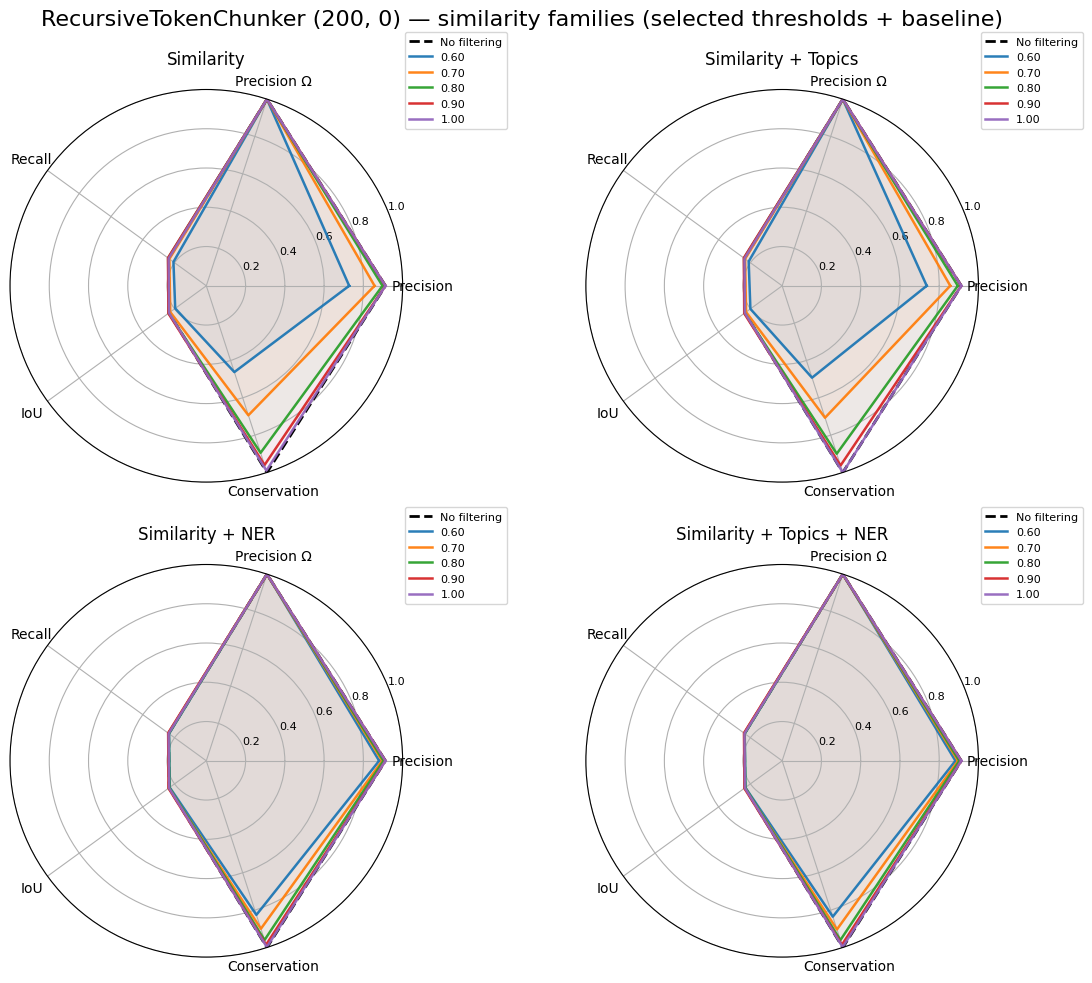}
        \caption{}
    \end{subfigure}
    \hfill
    \begin{subfigure}{0.48\textwidth}
        \includegraphics[width=\linewidth]{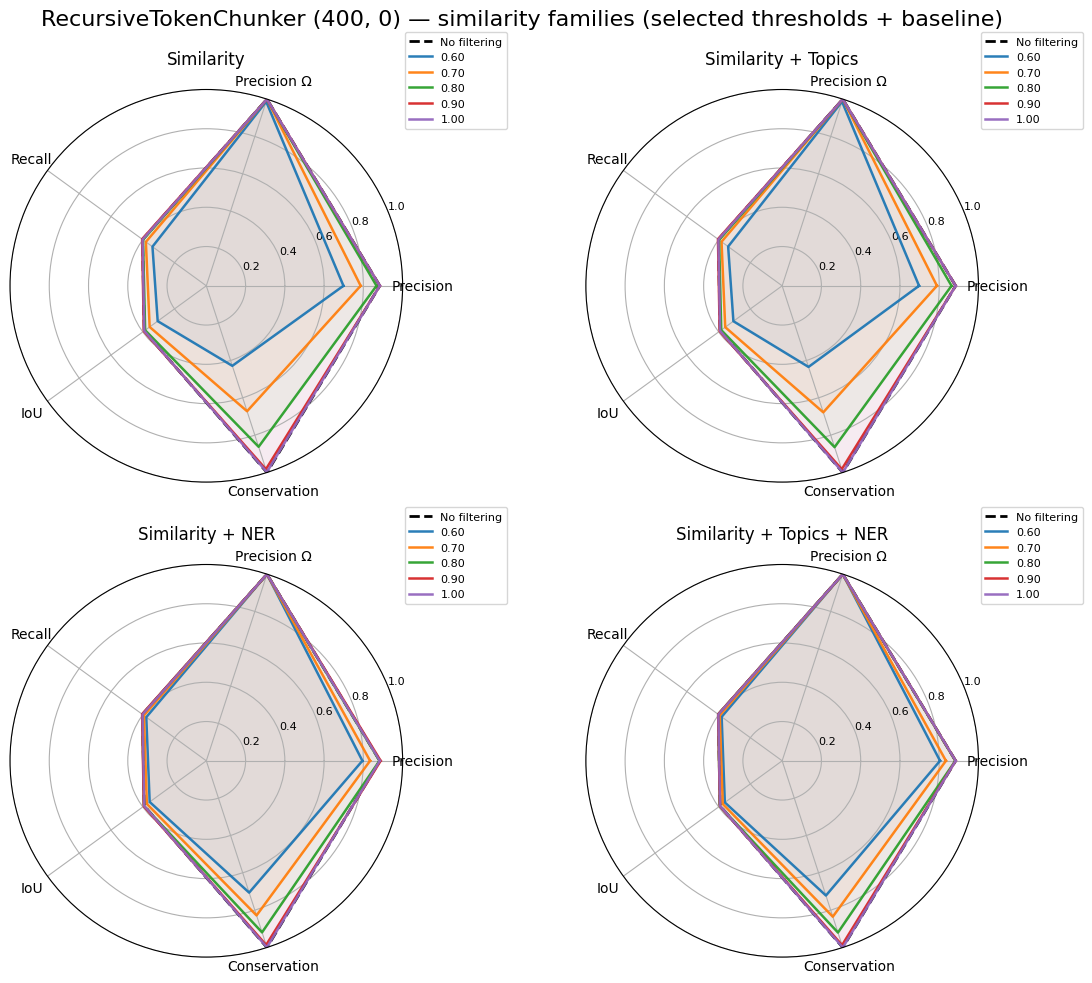}
        \caption{}
    \end{subfigure}

    \vspace{0.5em}

    \begin{subfigure}{0.48\textwidth}
        \includegraphics[width=\linewidth]{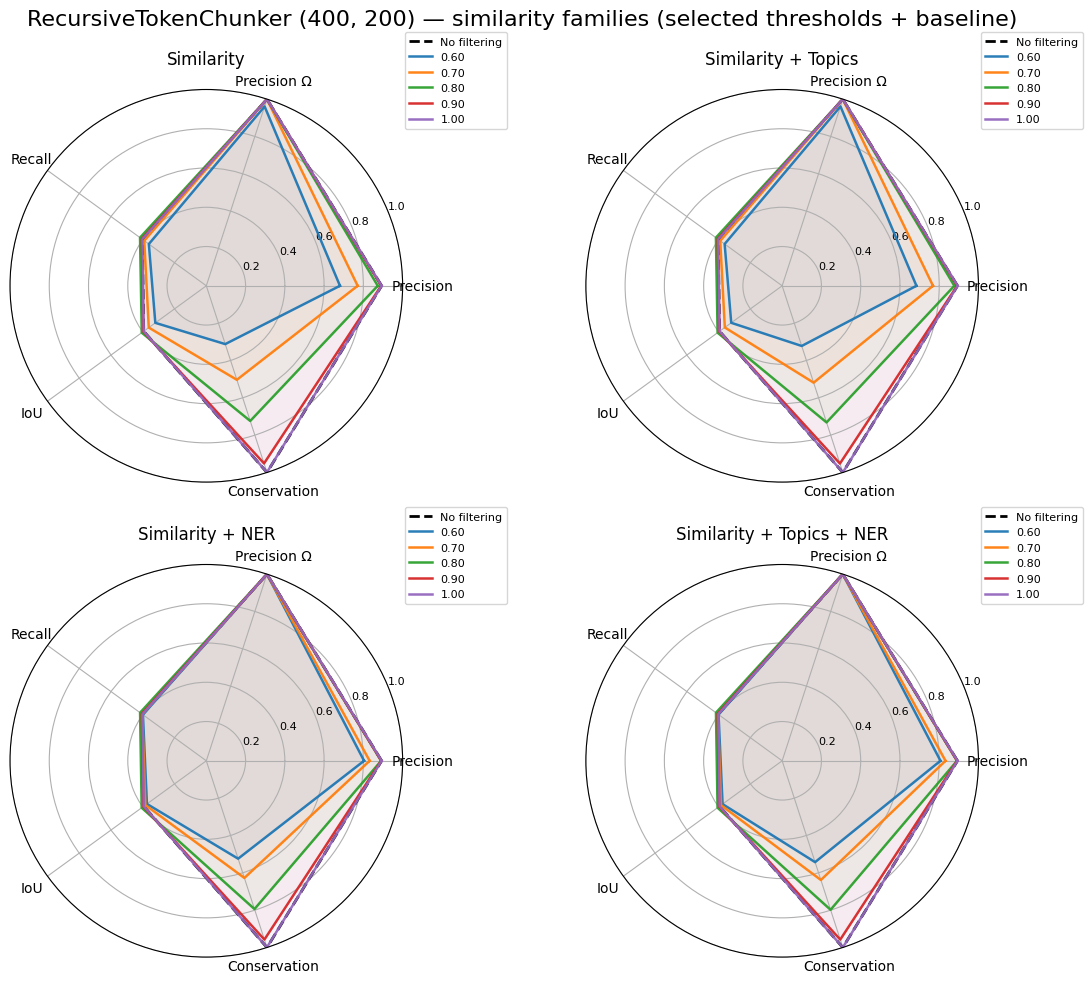}
        \caption{}
    \end{subfigure}
    \hfill
    \begin{subfigure}{0.48\textwidth}
        \includegraphics[width=\linewidth]{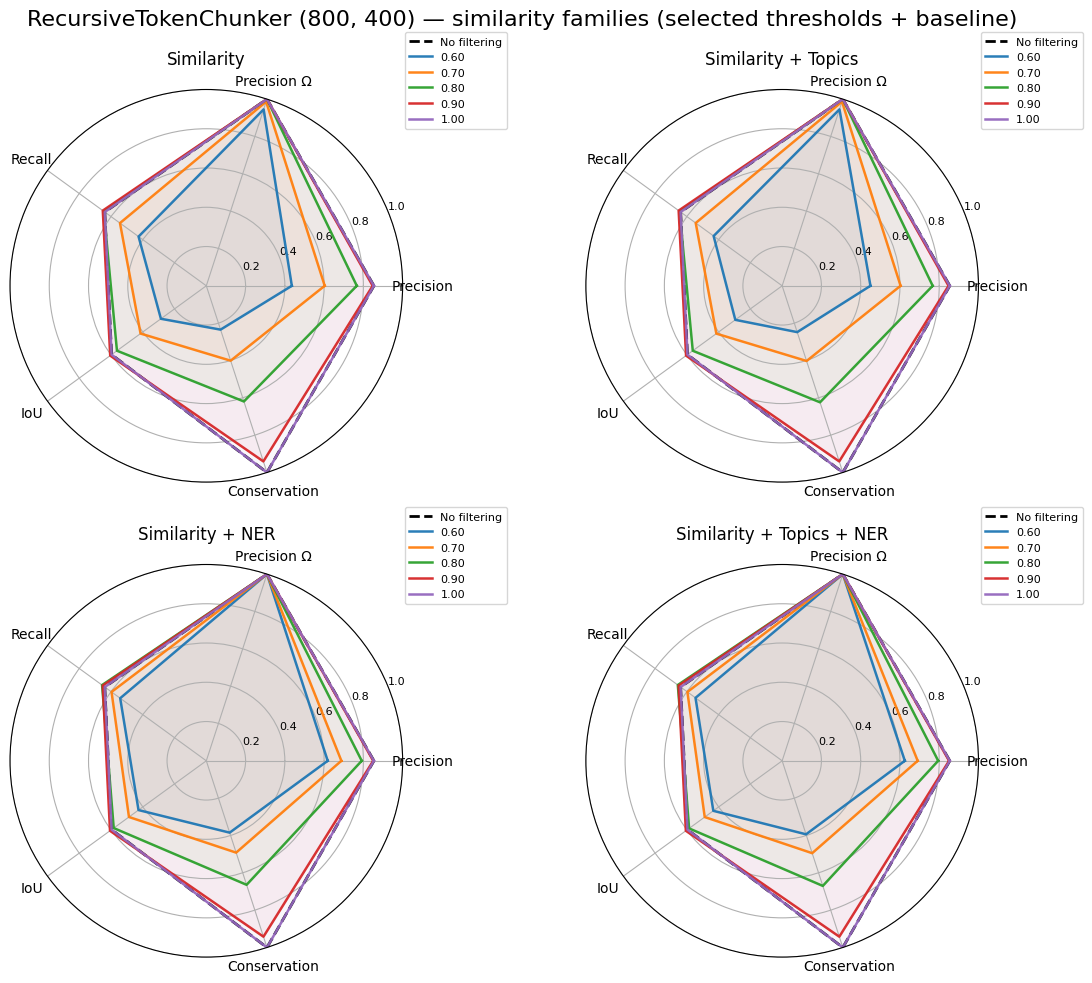}
        \caption{}
    \end{subfigure}

    \caption{RecursiveTokenChunker and similarity-based filtres for webfaq corpus}
    \label{webfaq:rec-sim}
\end{figure}

\begin{figure}[!htbp]
    \centering

    \begin{subfigure}{0.48\textwidth}
        \includegraphics[width=\linewidth]{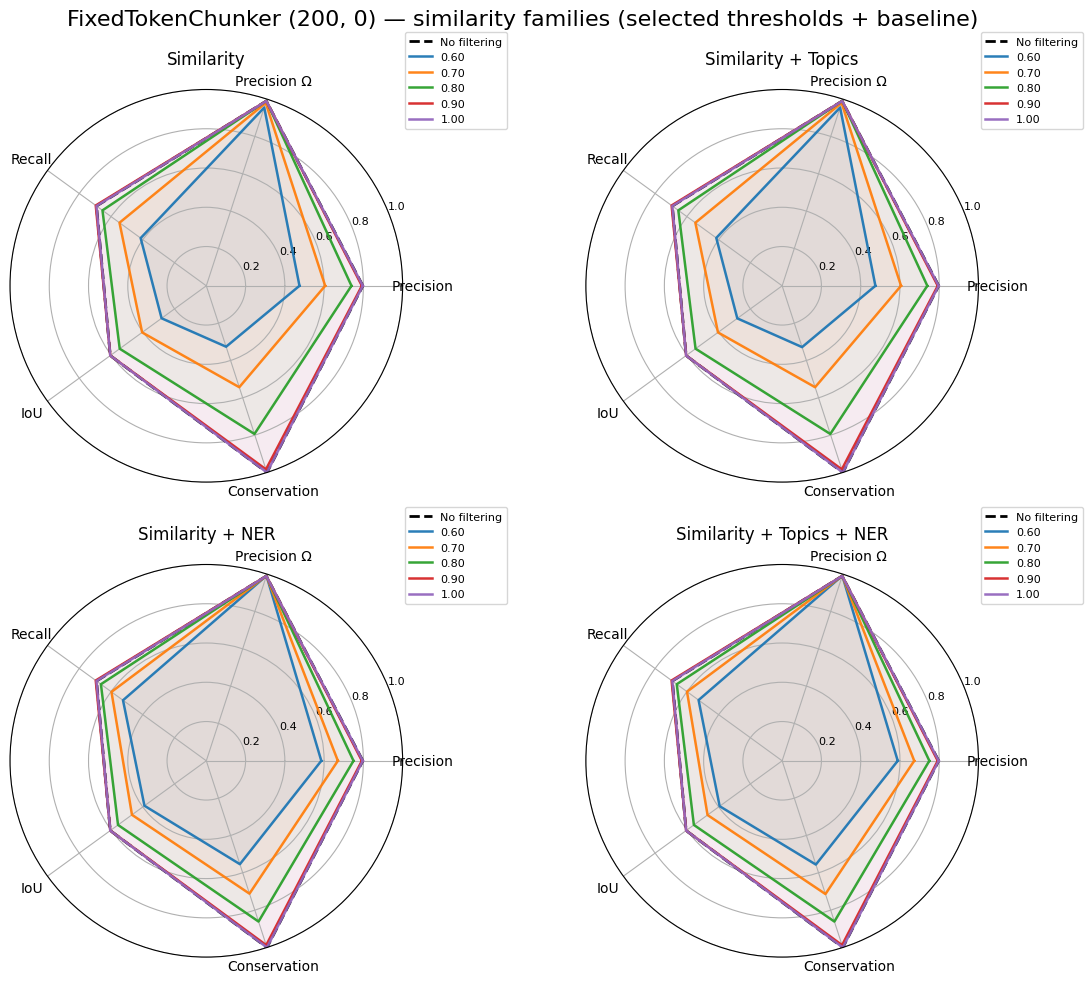}
        \caption{}
    \end{subfigure}
    \hfill
    \begin{subfigure}{0.48\textwidth}
        \includegraphics[width=\linewidth]{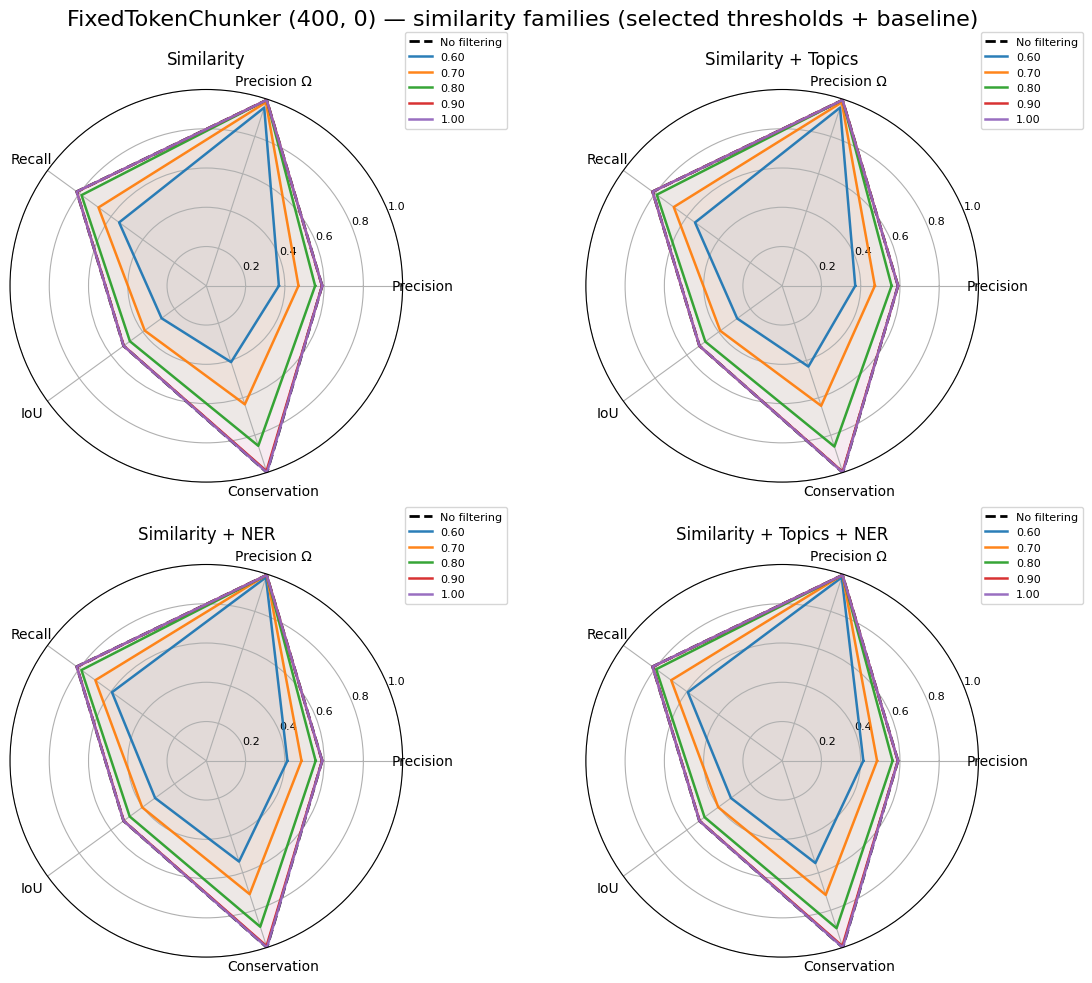}
        \caption{}
    \end{subfigure}

    \vspace{0.5em}

    \begin{subfigure}{0.48\textwidth}
        \includegraphics[width=\linewidth]{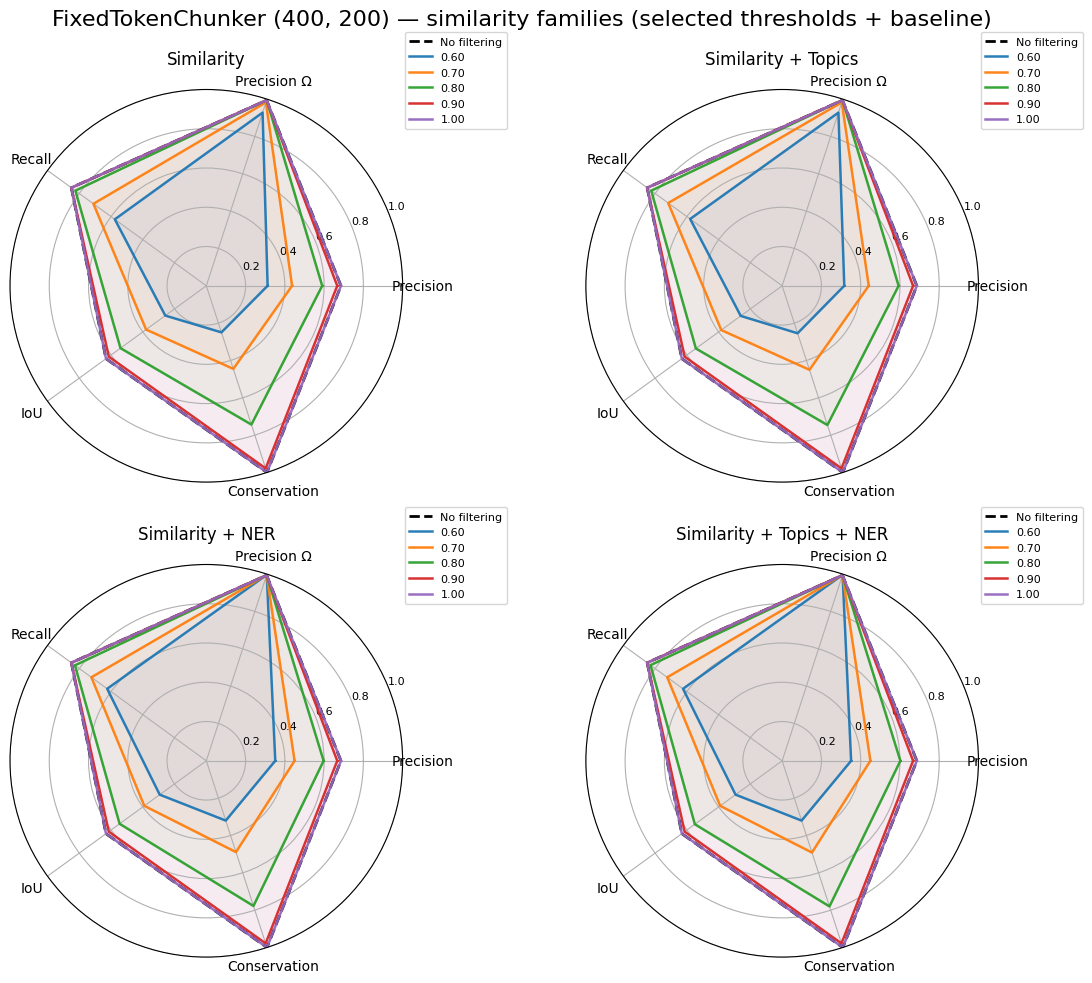}
        \caption{}
    \end{subfigure}
    \hfill
    \begin{subfigure}{0.48\textwidth}
        \includegraphics[width=\linewidth]{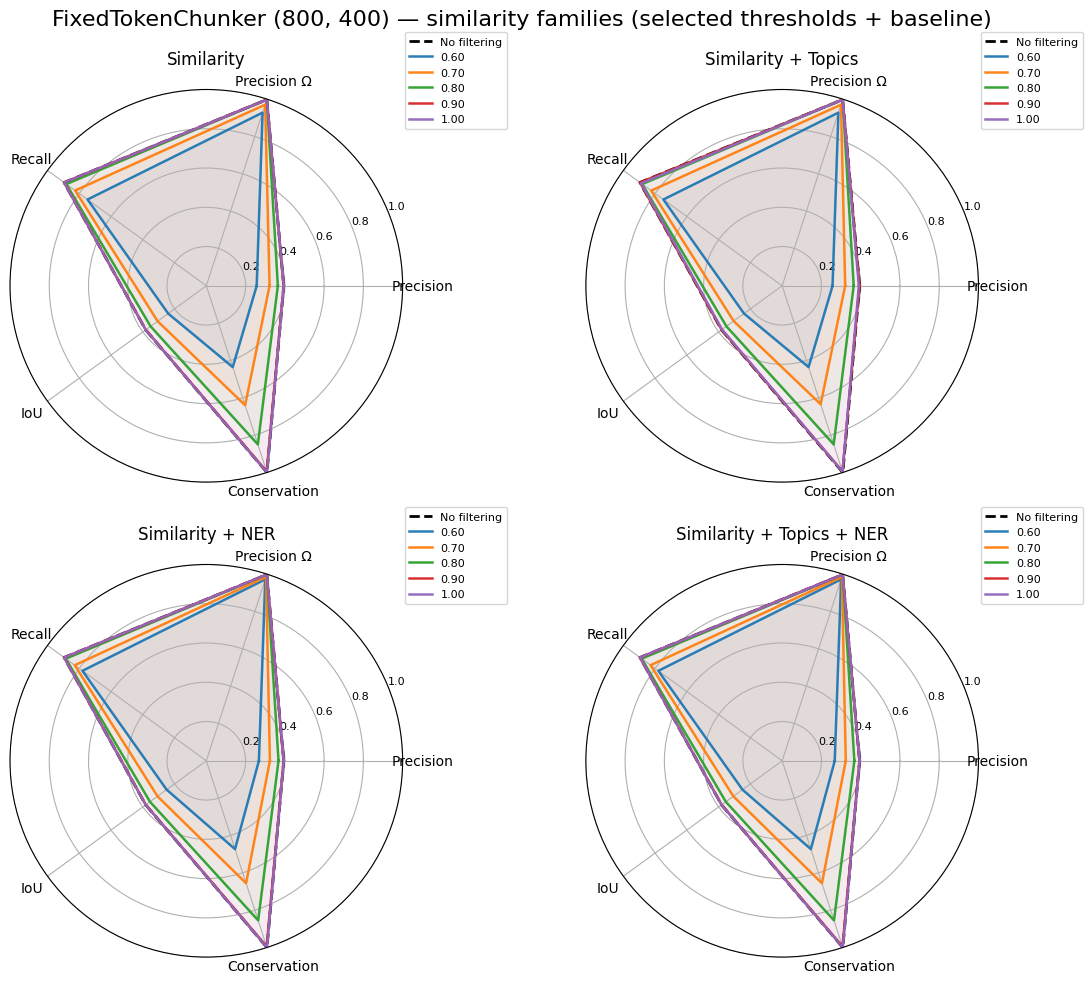}
        \caption{}
    \end{subfigure}

    \caption{FixedTokenChunker and similarity-based filtres for webfaq corpus}
    \label{webfaq:fix-sim}
\end{figure}

\begin{figure}[!htbp]
    \centering

    \begin{subfigure}{0.48\textwidth}
        \includegraphics[width=\linewidth]{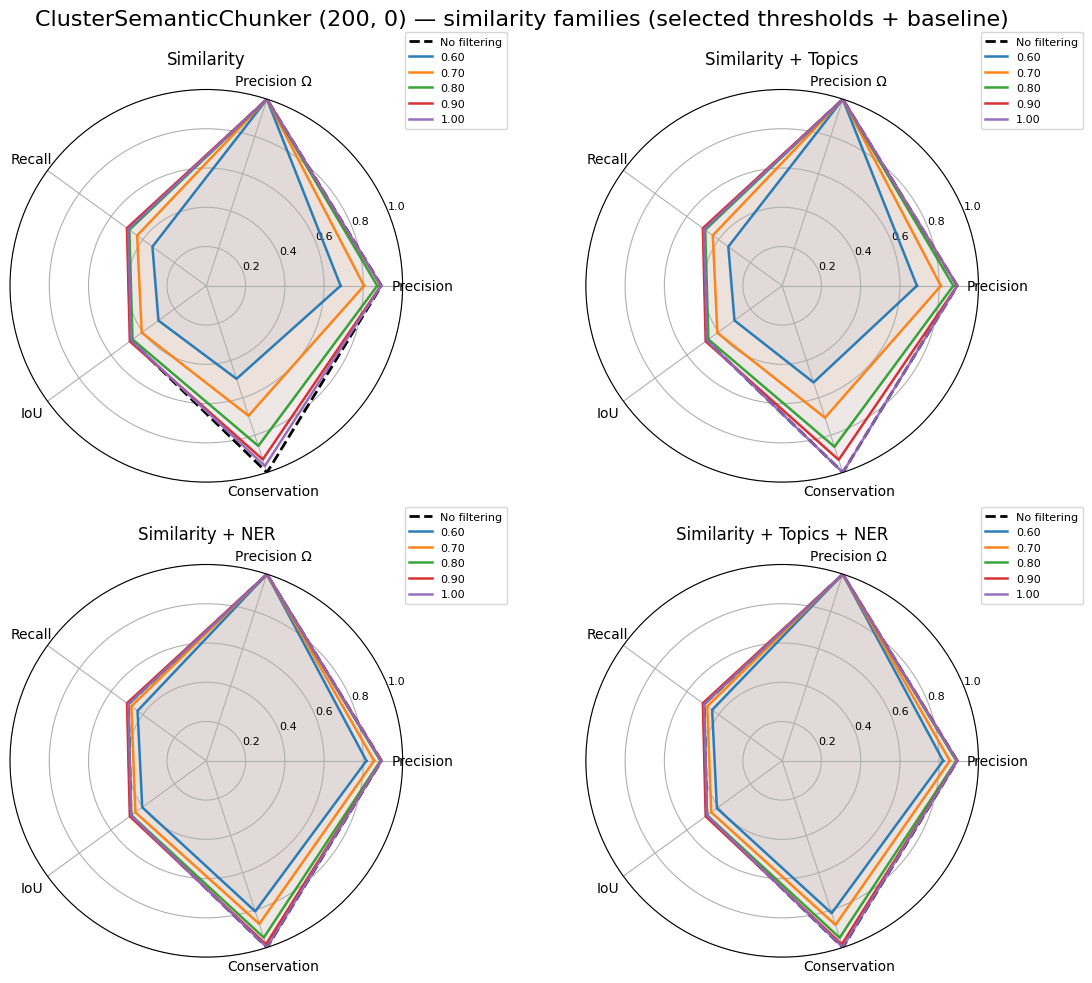}
        \caption{}
    \end{subfigure}
    \hfill
    \begin{subfigure}{0.48\textwidth}
        \includegraphics[width=\linewidth]{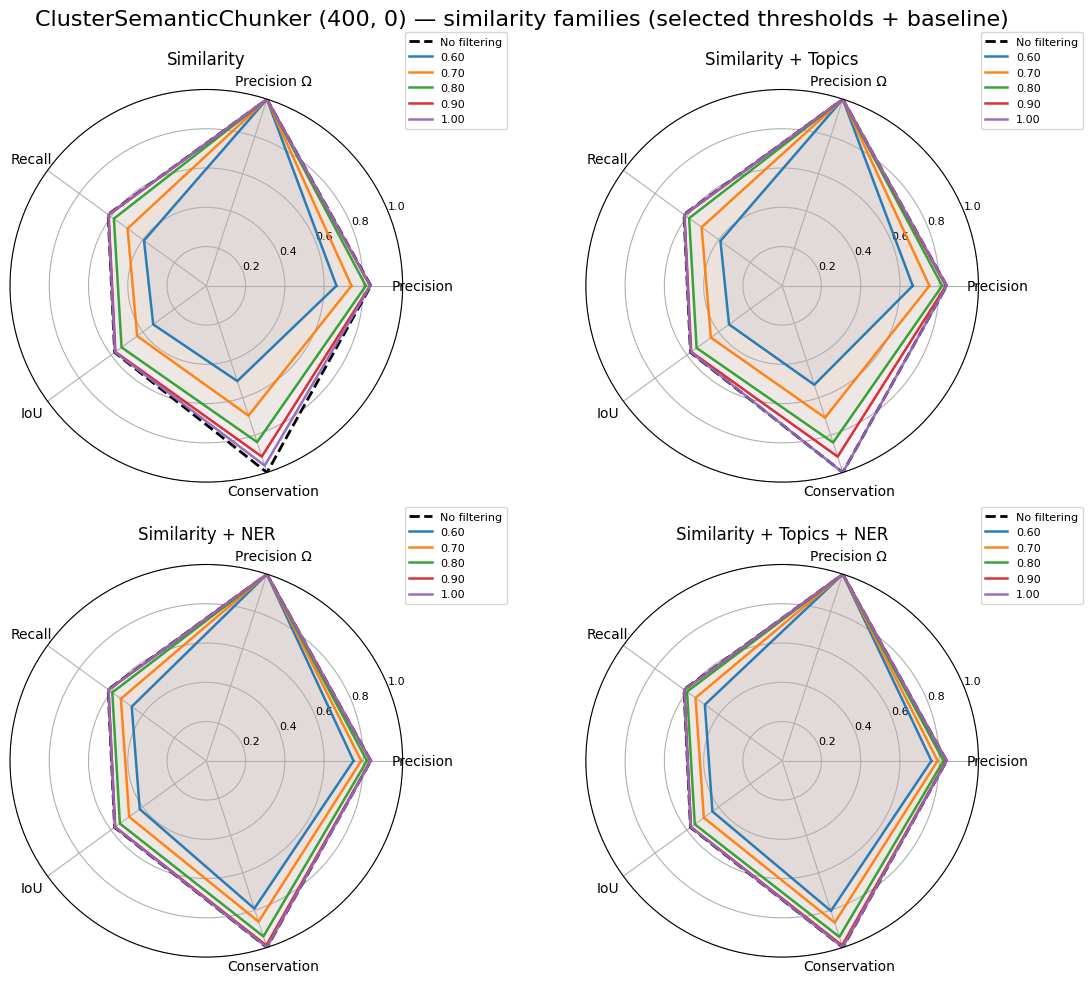}
        \caption{}
    \end{subfigure}

    \caption{ClusterSemanticChunker and similarity-based filtres for Chroma corpus}
    \label{webfaq:clust-sim}
\end{figure}

Figures~\ref{webfaq:rec-sim}, \ref{webfaq:fix-sim}, \ref{webfaq:clust-sim} show the results obtained with similarity-based and hybrid filtering strategies, including \emph{Similarity}, \emph{Similarity + Topics}, \emph{Similarity + NER}, and \emph{Similarity + Topics + NER}. 

In this multilingual setting, filtering based solely on semantic similarity provides a strong baseline in terms of recall. Because dense retrieval relies entirely on embedding proximity, the system tends to retrieve a broad set of passages that are thematically related to the query. While this behavior ensures high coverage of relevant tokens, it also introduces additional noise, resulting in lower precision and IoU values.

This effect is amplified by the heterogeneity of WebFAQ. Differences in vocabulary, syntax, and writing style across languages increase lexical variability, making semantic similarity less discriminative. As a result, chunks that share a general topic may still lack the precise information needed to answer the query.

Adding topic-based filtering leads to only modest improvements. Although topic grouping slightly reduces redundancy and increases precision in some configurations, its overall effect remains limited. The multilingual and multi-domain nature of the corpus reduces the discriminative power of topic modeling compared with more homogeneous datasets.

\subsubsection{NER-based filters}

\begin{figure}[!htbp]
    \centering

    \begin{subfigure}{0.48\textwidth}
        \includegraphics[width=\linewidth]{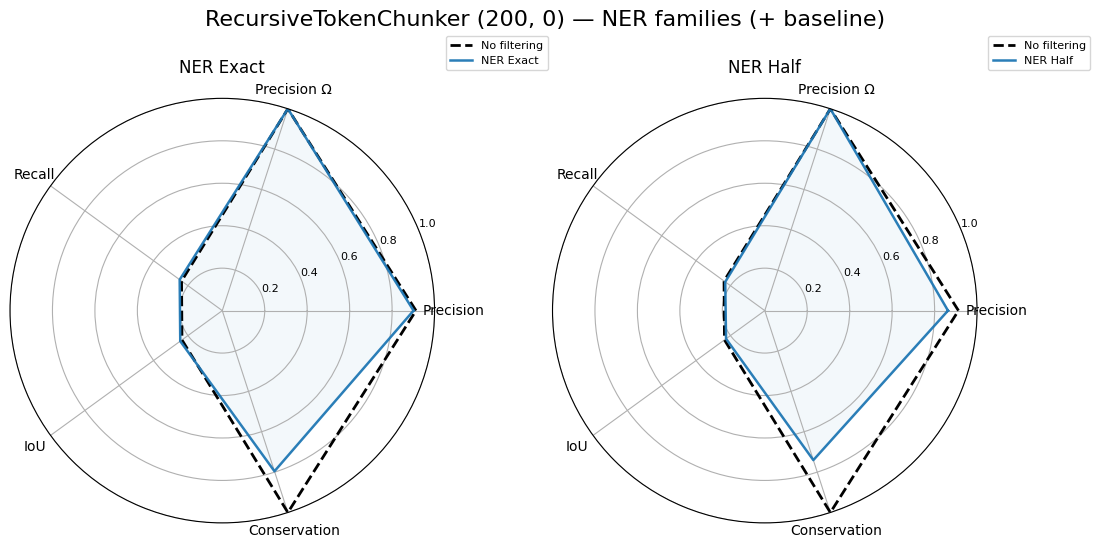}
        \caption{}
    \end{subfigure}
    \hfill
    \begin{subfigure}{0.48\textwidth}
        \includegraphics[width=\linewidth]{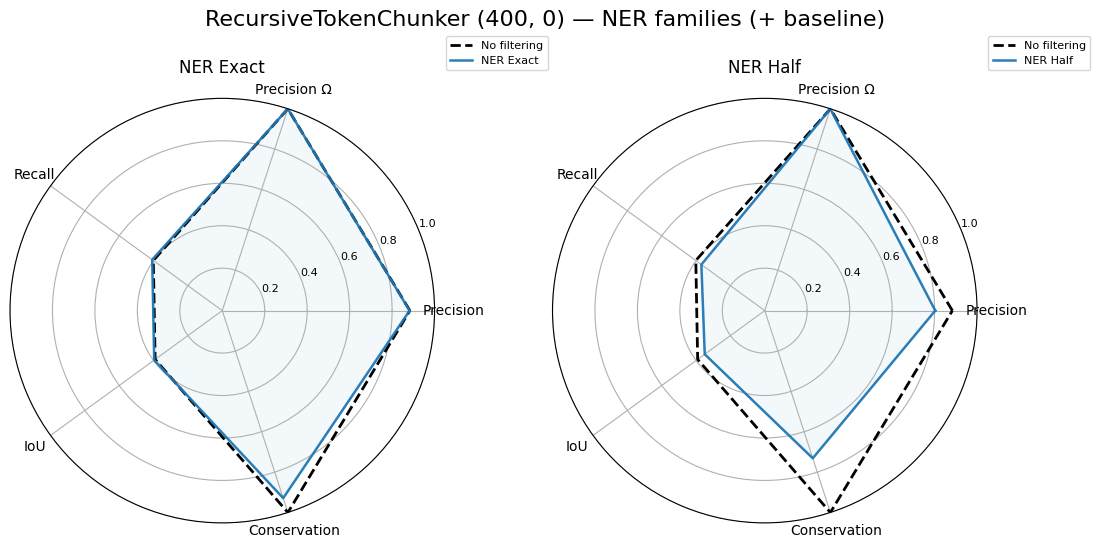}
        \caption{}
    \end{subfigure}

    \vspace{0.5em}

    \begin{subfigure}{0.48\textwidth}
        \includegraphics[width=\linewidth]{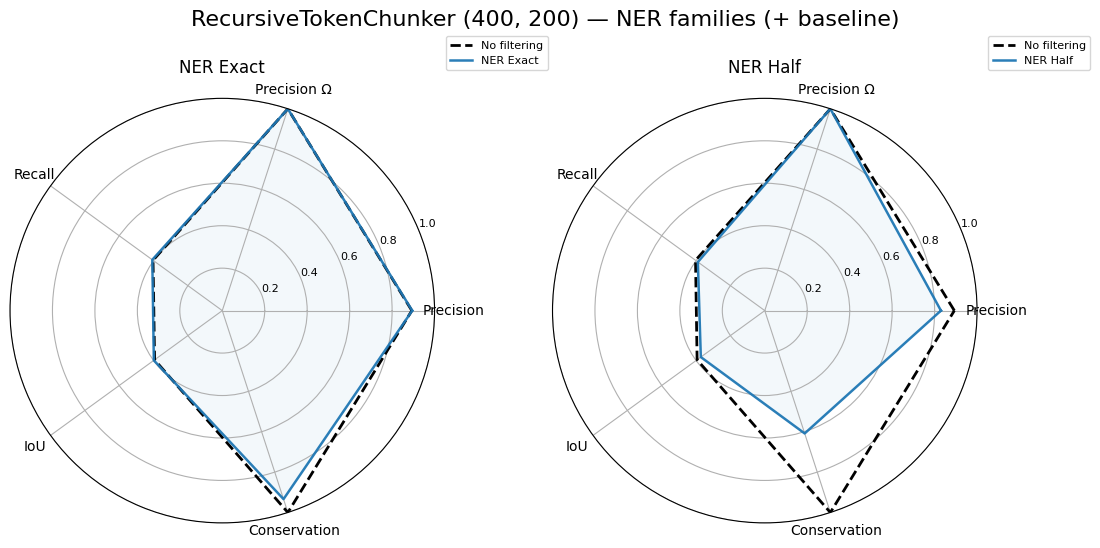}
        \caption{}
    \end{subfigure}
    \hfill
    \begin{subfigure}{0.48\textwidth}
        \includegraphics[width=\linewidth]{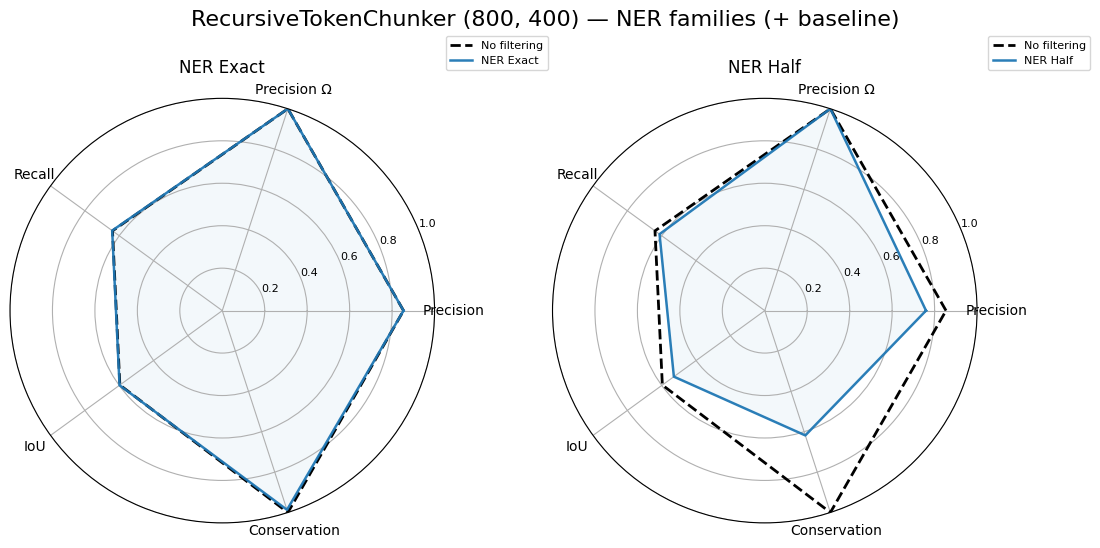}
        \caption{}
    \end{subfigure}

    \caption{RecursiveTokenChunker and NER-based filtres for webfaq corpus}
    \label{webfaq:rec-ner}
\end{figure}

\begin{figure}[!htbp]
    \centering

    \begin{subfigure}{0.48\textwidth}
        \includegraphics[width=\linewidth]{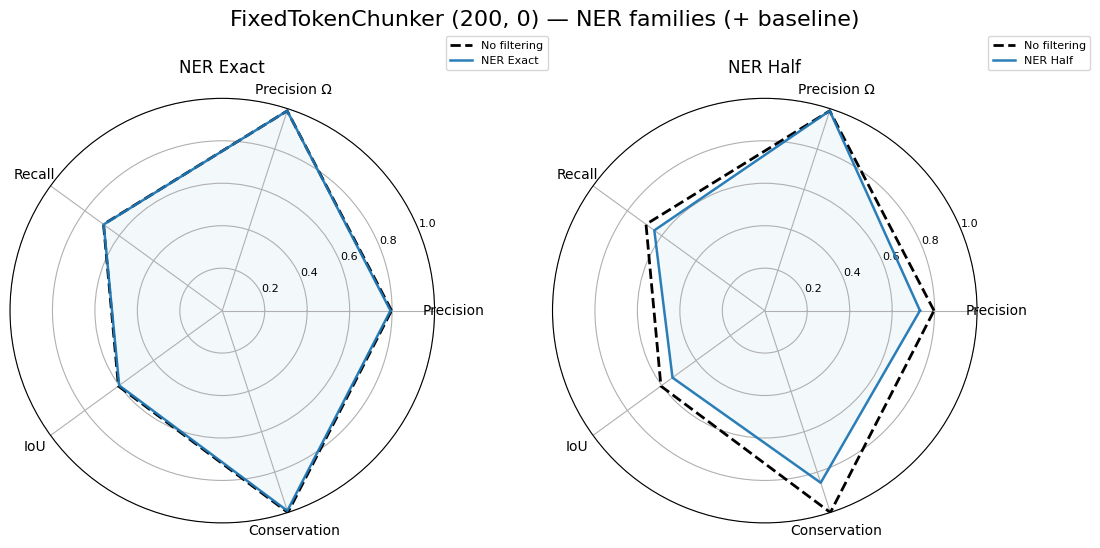}
        \caption{}
    \end{subfigure}
    \hfill
    \begin{subfigure}{0.48\textwidth}
        \includegraphics[width=\linewidth]{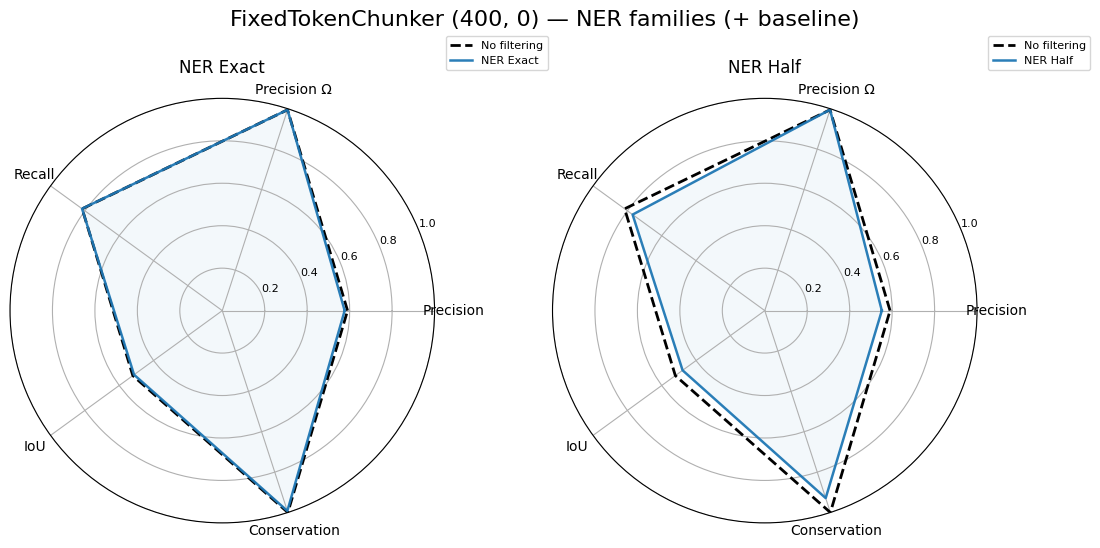}
        \caption{}
    \end{subfigure}

    \vspace{0.5em}

    \begin{subfigure}{0.48\textwidth}
        \includegraphics[width=\linewidth]{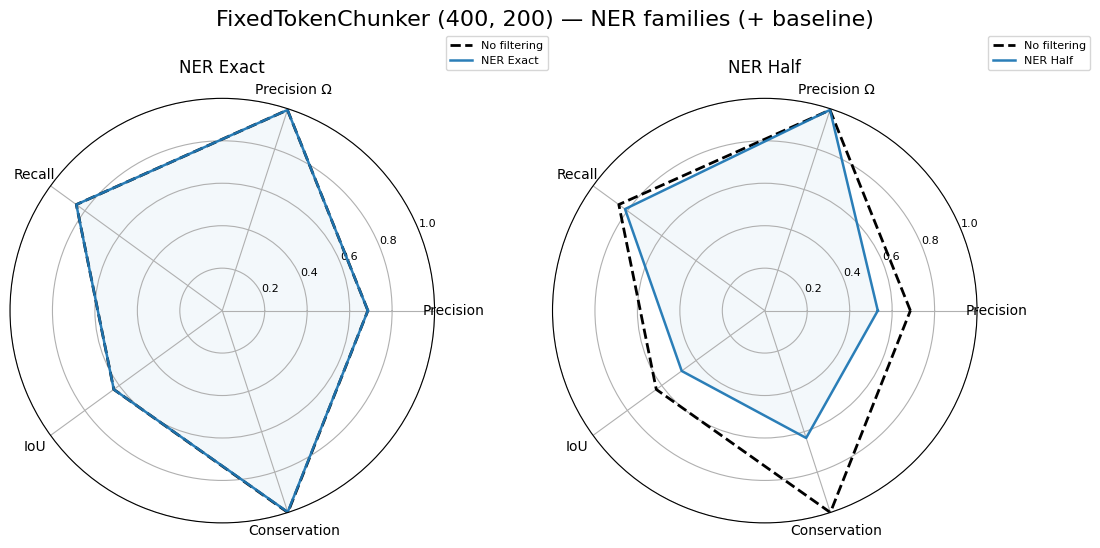}
        \caption{}
    \end{subfigure}
    \hfill
    \begin{subfigure}{0.48\textwidth}
        \includegraphics[width=\linewidth]{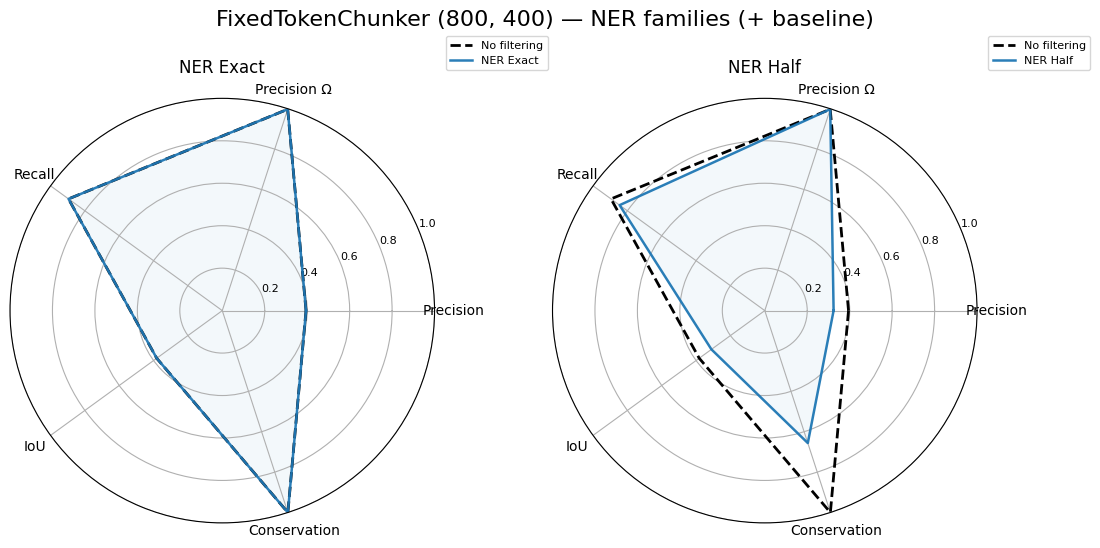}
        \caption{}
    \end{subfigure}

    \caption{FixedTokenChunker and NER-based filtres for webfaq corpus}
    \label{webfaq:fix-ner}
\end{figure}

\begin{figure}[!htbp]
    \centering

    \begin{subfigure}{0.48\textwidth}
        \includegraphics[width=\linewidth]{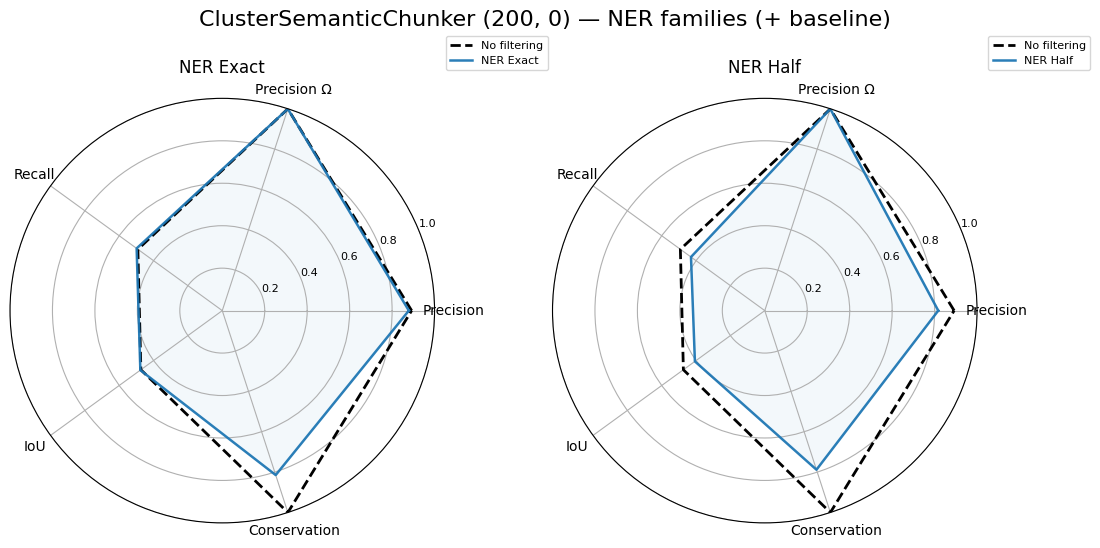}
        \caption{}
    \end{subfigure}
    \hfill
    \begin{subfigure}{0.48\textwidth}
        \includegraphics[width=\linewidth]{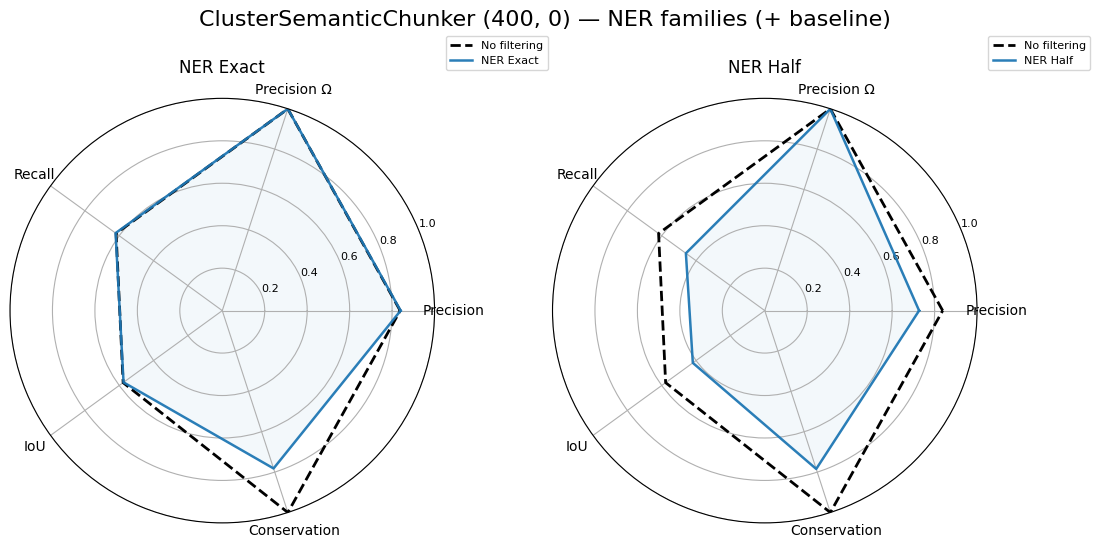}
        \caption{}
    \end{subfigure}

    \caption{ClusterSemanticChunker and NER-based filtres for webfaq corpus}
    \label{webfaq:clust-ner}
\end{figure}

Figures~\ref{webfaq:rec-ner}, \ref{webfaq:fix-ner} and \ref{webfaq:clust-ner} present the results obtained with the two entity-based filtering strategies: \emph{NER Exact} and \emph{NER Half}. In this multilingual context, named entities provide a stronger structural signal than thematic grouping alone.

Combining semantic similarity with entity constraints (\emph{Similarity + NER}) improves the balance between recall and precision. Named entities act as anchors that help identify passages containing concrete factual information, which increases the selectivity of the retrieved context.

Among all tested approaches, \emph{NER Exact} clearly provides the most favorable trade-off. This strategy removes chunks that share the exact same set of named entities, which often correspond to redundant factual descriptions. In practice, this filtering achieves a substantial reduction in index size while maintaining retrieval metrics comparable to, and sometimes better than, similarity-based filtering alone.

By contrast, \emph{NER Half} proves less reliable. Allowing partial entity overlap introduces additional noise, since chunks may share some entities while still describing different events or contexts. As a result, this strategy often produces a noticeable decrease in precision and IoU.

\subsubsection{Main findings}

Overall, the experiments on WebFAQ confirm the main trends observed on SQuAD while highlighting the specific challenges of multilingual and heterogeneous corpora. Similarity-based filtering alone favors recall and coverage but tends to retrieve broader and noisier contexts. Topic-based filtering brings only marginal improvements due to the diversity of themes present in the dataset.

The control baselines reinforce this interpretation. Lexical deduplication methods remove only a limited share of the index, which indicates that redundancy in WebFAQ is not mainly reducible through near-exact lexical matching. In addition, random filtering leads to much larger drops in recall and IoU at matched reduction levels, showing that the benefits of the proposed methods do not come from shrinking the index per se, but from removing chunks that are genuinely redundant from the perspective of retrieval.

Entity-based filtering plays a more decisive role in this setting. In particular, \emph{NER Exact} consistently provides the best compromise between index reduction and retrieval quality. By exploiting named entities as stable cross-lingual anchors, this strategy removes a large fraction of redundant passages while preserving the informational diversity required for effective retrieval.

Across configurations, filtering strategies reduce the size of the index by approximately 25\% to 35\% while maintaining high recall levels. Precision typically decreases by about 3\% to 8\%, whereas recall remains relatively stable with drops generally below 6\%. The IoU metric follows a similar trend, indicating that the filtering process primarily removes redundant segments rather than passages containing unique information.

These results suggest that entity-based redundancy detection is particularly effective in multilingual RAG pipelines, where semantic similarity alone may struggle to capture the structural relationships between documents written in different languages.

This result should nevertheless be interpreted in light of the structural bias of entity-based filtering. \emph{NER Exact} only affects chunks containing recognized named entities, and therefore does not operate uniformly across all document types. In more descriptive or narrative passages, where named entities are sparse, the filter may leave large portions of the corpus unchanged. Its favorable behavior may therefore partly reflect selective action on entity-rich chunks rather than a uniform capacity to remove redundancy across the full dataset.

\section{Conclusion}

This work investigated the impact of chunk filtering strategies in Retrieval-Augmented Generation (RAG) pipelines. We are particullary focusing on the trade-off between vector index compactness and retrieval quality. Rather than evaluating end-to-end answer generation, the study focused on the indexing and retrieval stages of RAG-oriented systems. The experiments conducted on several datasets—including the corpora of the Chroma study, SQuAD 1.1, and the multilingual WebFAQ corpus—show that a substantial portion of the redundancy introduced by certain chunking strategies can be removed without significantly degrading retrieval performance.

Across multiple configurations, the results demonstrate that the size of the vector index can be reduced by approximately 25\% to 36\% while maintaining retrieval metrics close to the baseline. In particular, the \emph{NER Exact} often provides one of the most favorable compromise between index reduction and retrieval effectiveness. In most cases, recall decreases by less than 6\% and IoU by less than 8\%, suggesting that many indexed chunks correspond to redundant information rather than genuinely new retrieval content.

The oracle metric reveals that some filtering strategies, including NER Exact, remove chunks that contain reference information, as reflected by a decrease in oracle coverage. This highlights an important problem. While filtering reduces redundancy, it may also remove informative variants of the same content. However, the limited impact on retrieval metrics suggests that the remaining chunks still provide sufficient coverage for most queries.

The comparison with ExactNorm and MinHash-LSH suggests that chunk-level redundancy in RAG pipelines is not mainly driven by exact or near-exact lexical duplication. The random filtering baseline further shows that these gains are not explained by index reduction alone: indiscriminate chunk removal produces substantially larger retrieval degradation than signal-based filtering at comparable reduction levels. Instead, a substantial part of the redundancy appears at the semantic or structural level, which explains why lexical baselines remain much less effective than embedding-based or entity-based filters.

The experiments reveal a strong link between chunking and filtering strategies. Chunkers producing semantically coherent segments (such as recursive or semantic chunkers) tend to provide more stable precision--recall balance. In contrast, fixed-window segmentation strategies often introduce higher local redundancy. This makes the filtering process more sensitive to the choice of thresholds.

Alongside the empirical results, this study also provides methodological contributions. First, it compares chunk filtering strategies based on different signals (e.g., semantic similarity, topic modeling, and named entities) in a controlled setting. Second, it introduces a token-coverage-based evaluation protocol that is less sensitive to variations in chunk boundaries than approaches based on character offsets. Third, the use of an oracle retrieval metric (Precision $\Omega@k$) allows a clearer interpretation of results by separating retrieval limitations from effects induced by segmentation or filtering within the available index.

A central limitation of the present study is that it evaluates retrieval-level behavior and do not analyze end-to-end answer generation. As a result, the reported gains should not be interpreted as direct evidence of improved final response quality. A chunk considered redundant from the perspective of token coverage may still provide useful contextual information for a downstream language model. Evaluating this interaction requires an end-to-end generative setup and is left for future work.

A further limitation concerns the use of named-entity-based filtering. Since such methods only apply to chunks containing recognized entities, their behavior is partly conditioned by corpus-specific entity density. The strong performance of \emph{NER Exact} should therefore not be interpreted as a domain-independent result, but rather as evidence that entity overlap can be an effective redundancy signal in corpora where factual content is strongly structured around named entities.

Several directions for future research rise from this work. A first extension would be to evaluate whether the retrieval-level gains observed here translate into improvements in downstream answer generation with an LLM. Other promising directions include studying the interaction between upstream filtering and downstream reranking mechanisms. Moreover, it could be useful to explore structural relationships between chunks through graph-based representations.

More broadly, these results suggest that improving retrieval-oriented preprocessing is an important component of scalable RAG systems. Even when final answer quality is not directly measured, the organization and selection of information at the document level remain critical for building efficient and compact retrieval pipelines over large document collections.

\section*{Acknowledgements}
The authors acknowledge the use of ChatGPT to assist in editing the text to improve its form, and they take full responsibility for the content of this editorial.
\FloatBarrier
\printbibliography

@misc{lewis_retrieval-augmented_2021,
	title = {Retrieval-{Augmented} {Generation} for {Knowledge}-{Intensive} {NLP} {Tasks}},
	url = {http://arxiv.org/abs/2005.11401},
	doi = {10.48550/arXiv.2005.11401},
	urldate = {2025-12-15},
	publisher = {arXiv},
	author = {Lewis, Patrick and Perez, Ethan and Piktus, Aleksandra and Petroni, Fabio and Karpukhin, Vladimir and Goyal, Naman and Küttler, Heinrich and Lewis, Mike and Yih, Wen-tau and Rocktäschel, Tim and Riedel, Sebastian and Kiela, Douwe},
	month = apr,
	year = {2021},
	note = {arXiv:2005.11401 [cs]},
}

@article{blei_latent_2003,
	title = {Latent dirichlet allocation},
	volume = {3},
	issn = {1532-4435},
	number = {null},
	journal = {J. Mach. Learn. Res.},
	author = {Blei, David M. and Ng, Andrew Y. and Jordan, Michael I.},
	month = mar,
	year = {2003},
	pages = {993--1022},
}

@misc{grootendorst_bertopic_2022,
	title = {{BERTopic}: {Neural} topic modeling with a class-based {TF}-{IDF} procedure},
	shorttitle = {{BERTopic}},
	url = {http://arxiv.org/abs/2203.05794},
	doi = {10.48550/arXiv.2203.05794},
	urldate = {2025-12-16},
	publisher = {arXiv},
	author = {Grootendorst, Maarten},
	month = mar,
	year = {2022},
	note = {arXiv:2203.05794 [cs]},
}

@misc{reimers_sentence-bert_2019,
	title = {Sentence-{BERT}: {Sentence} {Embeddings} using {Siamese} {BERT}-{Networks}},
	shorttitle = {Sentence-{BERT}},
	url = {http://arxiv.org/abs/1908.10084},
	doi = {10.48550/arXiv.1908.10084},
	urldate = {2025-12-16},
	publisher = {arXiv},
	author = {Reimers, Nils and Gurevych, Iryna},
	month = aug,
	year = {2019},
	note = {arXiv:1908.10084 [cs]},
}

@misc{salvatore_lost_2025,
	title = {Lost in the {Middle}: {An} {Emergent} {Property} from {Information} {Retrieval} {Demands} in {LLMs}},
	shorttitle = {Lost in the {Middle}},
	url = {http://arxiv.org/abs/2510.10276},
	doi = {10.48550/arXiv.2510.10276},
	urldate = {2025-12-16},
	publisher = {arXiv},
	author = {Salvatore, Nikolaus and Wang, Hao and Zhang, Qiong},
	month = oct,
	year = {2025},
	note = {arXiv:2510.10276 [cs]},
}

@article{honnibal_spacy_2020,
	title = {{spaCy}: {Industrial}-strength {Natural} {Language} {Processing} in {Python}},
	shorttitle = {{spaCy}},
	doi = {10.5281/zenodo.1212303},
	author = {Honnibal, Matthew and Montani, Ines and Van Landeghem, Sofie and Boyd, Adriane},
	year = {2020},
}

@misc{karpukhin_dense_2020,
	title = {Dense {Passage} {Retrieval} for {Open}-{Domain} {Question} {Answering}},
	url = {http://arxiv.org/abs/2004.04906},
	doi = {10.48550/arXiv.2004.04906},
	urldate = {2025-12-16},
	publisher = {arXiv},
	author = {Karpukhin, Vladimir and Oğuz, Barlas and Min, Sewon and Lewis, Patrick and Wu, Ledell and Edunov, Sergey and Chen, Danqi and Yih, Wen-tau},
	month = sep,
	year = {2020},
	note = {arXiv:2004.04906 [cs]},
}

@misc{gao_retrieval-augmented_2024,
	title = {Retrieval-{Augmented} {Generation} for {Large} {Language} {Models}: {A} {Survey}},
	shorttitle = {Retrieval-{Augmented} {Generation} for {Large} {Language} {Models}},
	url = {http://arxiv.org/abs/2312.10997},
	doi = {10.48550/arXiv.2312.10997},
	urldate = {2025-12-16},
	publisher = {arXiv},
	author = {Gao, Yunfan and Xiong, Yun and Gao, Xinyu and Jia, Kangxiang and Pan, Jinliu and Bi, Yuxi and Dai, Yi and Sun, Jiawei and Wang, Meng and Wang, Haofen},
	month = mar,
	year = {2024},
	note = {arXiv:2312.10997 [cs]},
}

@misc{merity_pointer_2016,
	title = {Pointer {Sentinel} {Mixture} {Models}},
	url = {http://arxiv.org/abs/1609.07843},
	doi = {10.48550/arXiv.1609.07843},
	urldate = {2025-12-17},
	publisher = {arXiv},
	author = {Merity, Stephen and Xiong, Caiming and Bradbury, James and Socher, Richard},
	month = sep,
	year = {2016},
	note = {arXiv:1609.07843 [cs]},
}

@misc{ding_enhancing_2023,
	title = {Enhancing {Chat} {Language} {Models} by {Scaling} {High}-quality {Instructional} {Conversations}},
	url = {http://arxiv.org/abs/2305.14233},
	doi = {10.48550/arXiv.2305.14233},
	urldate = {2025-12-17},
	publisher = {arXiv},
	author = {Ding, Ning and Chen, Yulin and Xu, Bokai and Qin, Yujia and Zheng, Zhi and Hu, Shengding and Liu, Zhiyuan and Sun, Maosong and Zhou, Bowen},
	month = may,
	year = {2023},
	note = {arXiv:2305.14233 [cs]},
}

@misc{chen_convfinqa_2022,
	title = {{ConvFinQA}: {Exploring} the {Chain} of {Numerical} {Reasoning} in {Conversational} {Finance} {Question} {Answering}},
	shorttitle = {{ConvFinQA}},
	url = {http://arxiv.org/abs/2210.03849},
	doi = {10.48550/arXiv.2210.03849},
	urldate = {2025-12-17},
	publisher = {arXiv},
	author = {Chen, Zhiyu and Li, Shiyang and Smiley, Charese and Ma, Zhiqiang and Shah, Sameena and Wang, William Yang},
	month = oct,
	year = {2022},
	note = {arXiv:2210.03849 [cs]},
}

@misc{noauthor_pmcopen_access_2023,
	title = {pmc/open\_access · {Datasets} at {Hugging} {Face}},
	url = {https://huggingface.co/datasets/pmc/open_access},
	urldate = {2025-12-17},
	month = aug,
	year = {2023},
}

@misc{noauthor_state_nodate,
	title = {State of the {Union} 2024},
	url = {https://bidenwhitehouse.archives.gov/state-of-the-union-2024/},
	language = {en-US},
	urldate = {2025-12-17},
	journal = {The White House},
}

@misc{noauthor_rajpurkarsquad_2021,
	title = {rajpurkar/squad · {Datasets} at {Hugging} {Face}},
	url = {https://huggingface.co/datasets/rajpurkar/squad},
	urldate = {2025-12-17},
	month = jun,
	year = {2021},
}

@misc{dinzinger_webfaq_2025,
	title = {{WebFAQ}: {A} {Multilingual} {Collection} of {Natural} {Q}\&{A} {Datasets} for {Dense} {Retrieval}},
	shorttitle = {{WebFAQ}},
	url = {http://arxiv.org/abs/2502.20936},
	doi = {10.48550/arXiv.2502.20936},
	urldate = {2025-12-17},
	publisher = {arXiv},
	author = {Dinzinger, Michael and Caspari, Laura and Dastidar, Kanishka Ghosh and Mitrović, Jelena and Granitzer, Michael},
	month = feb,
	year = {2025},
	note = {arXiv:2502.20936 [cs]},
}

@misc{chen_m3-embedding_2025,
	title = {M3-{Embedding}: {Multi}-{Linguality}, {Multi}-{Functionality}, {Multi}-{Granularity} {Text} {Embeddings} {Through} {Self}-{Knowledge} {Distillation}},
	shorttitle = {M3-{Embedding}},
	url = {http://arxiv.org/abs/2402.03216},
	doi = {10.48550/arXiv.2402.03216},
	urldate = {2025-12-18},
	publisher = {arXiv},
	author = {Chen, Jianlv and Xiao, Shitao and Zhang, Peitian and Luo, Kun and Lian, Defu and Liu, Zheng},
	month = dec,
	year = {2025},
	note = {arXiv:2402.03216 [cs]},
}

@misc{noauthor_chroma-corechroma_2025,
	title = {chroma-core/chroma},
	copyright = {Apache-2.0},
	url = {https://github.com/chroma-core/chroma},
	urldate = {2025-12-18},
	publisher = {Chroma},
	month = dec,
	year = {2025},
	note = {original-date: 2022-10-05T17:58:44Z},
}

@article{guu2020realm,
  author       = {Kelvin Guu and
                  Kenton Lee and
                  Zora Tung and
                  Panupong Pasupat and
                  Ming{-}Wei Chang},
  title        = {{REALM:} Retrieval-Augmented Language Model Pre-Training},
  journal      = {CoRR},
  volume       = {abs/2002.08909},
  year         = {2020},
  url          = {https://arxiv.org/abs/2002.08909},
  eprinttype    = {arXiv},
  eprint       = {2002.08909},
  bibsource    = {dblp computer science bibliography, https://dblp.org}
}

@inproceedings{izacard-grave-2021-leveraging,
    title = "Leveraging Passage Retrieval with Generative Models for Open Domain Question Answering",
    author = "Izacard, Gautier  and
      Grave, Edouard",
    editor = "Merlo, Paola  and
      Tiedemann, Jorg  and
      Tsarfaty, Reut",
    booktitle = "Proceedings of the 16th Conference of the European Chapter of the Association for Computational Linguistics: Main Volume",
    month = apr,
    year = "2021",
    address = "Online",
    publisher = "Association for Computational Linguistics",
    url = "https://aclanthology.org/2021.eacl-main.74/",
    doi = "10.18653/v1/2021.eacl-main.74",
    pages = "874--880"
}

@article{borgeaud2022retro,
  author       = {Sebastian Borgeaud and
                  Arthur Mensch and
                  Jordan Hoffmann and
                  Trevor Cai and
                  Eliza Rutherford and
                  Katie Millican and
                  George van den Driessche and
                  Jean{-}Baptiste Lespiau and
                  Bogdan Damoc and
                  Aidan Clark and
                  Diego de Las Casas and
                  Aurelia Guy and
                  Jacob Menick and
                  Roman Ring and
                  Tom Hennigan and
                  Saffron Huang and
                  Loren Maggiore and
                  Chris Jones and
                  Albin Cassirer and
                  Andy Brock and
                  Michela Paganini and
                  Geoffrey Irving and
                  Oriol Vinyals and
                  Simon Osindero and
                  Karen Simonyan and
                  Jack W. Rae and
                  Erich Elsen and
                  Laurent Sifre},
  title        = {Improving language models by retrieving from trillions of tokens},
  journal      = {CoRR},
  volume       = {abs/2112.04426},
  year         = {2021},
  url          = {https://arxiv.org/abs/2112.04426},
  eprinttype    = {arXiv},
  eprint       = {2112.04426},
  bibsource    = {dblp computer science bibliography, https://dblp.org}
}

@inproceedings{
asai2024selfrag,
author={Asai, Akari and Wu, Zeqiu and Wang, Yizhong and Sil, Avirup and Hajishirzi, Hannaneh},
title={Self-{RAG}: Learning to Retrieve, Generate, and Critique through Self-Reflection},
booktitle={The Twelfth International Conference on Learning Representations},
year={2024},
url={https://openreview.net/forum?id=hSyW5go0v8}
}

@inproceedings{
xiong2021approximate,
title={Approximate Nearest Neighbor Negative Contrastive Learning for Dense Text Retrieval},
author={Lee Xiong and Chenyan Xiong and Ye Li and Kwok-Fung Tang and Jialin Liu and Paul N. Bennett and Junaid Ahmed and Arnold Overwijk},
booktitle={International Conference on Learning Representations},
year={2021},
url={https://openreview.net/forum?id=zeFrfgyZln}
}

@inproceedings{lee-etal-2022-deduplicating,
    title = "Deduplicating Training Data Makes Language Models Better",
    author = "Lee, Katherine  and
      Ippolito, Daphne  and
      Nystrom, Andrew  and
      Zhang, Chiyuan  and
      Eck, Douglas  and
      Callison-Burch, Chris  and
      Carlini, Nicholas",
    editor = "Muresan, Smaranda  and
      Nakov, Preslav  and
      Villavicencio, Aline",
    booktitle = "Proceedings of the 60th Annual Meeting of the Association for Computational Linguistics (Volume 1: Long Papers)",
    month = may,
    year = "2022",
    address = "Dublin, Ireland",
    publisher = "Association for Computational Linguistics",
    url = "https://aclanthology.org/2022.acl-long.577/",
    doi = "10.18653/v1/2022.acl-long.577",
    pages = "8424--8445"
}

@article{Min2022RethinkingTR,
  title={Rethinking the Role of Demonstrations: What Makes In-Context Learning Work?},
  author={Sewon Min and Xinxi Lyu and Ari Holtzman and Mikel Artetxe and Mike Lewis and Hannaneh Hajishirzi and Luke Zettlemoyer},
  journal={ArXiv},
  year={2022},
  volume={abs/2202.12837},
  url={https://api.semanticscholar.org/CorpusID:247155069}
}

@inproceedings{broder1997resemblance,
  title={On the resemblance and containment of documents},
  author={Broder, Andrei Z.},
  booktitle={Compression and Complexity of Sequences},
  year={1997}
}

@inproceedings{broder2000identifying,
  title={Identifying and filtering near-duplicate documents},
  author={Broder, Andrei Z. and Charikar, Moses and Frieze, Alan M. and Mitzenmacher, Michael},
  booktitle={Combinatorial Pattern Matching},
  year={2000}
}

@book{leskovec2020mining,
  title={Mining of Massive Datasets},
  author={Leskovec, Jure and Rajaraman, Anand and Ullman, Jeffrey D.},
  year={2020},
  edition={3}
}

@misc{guenther2025latechunkingcontextualchunk,
      title={Late Chunking: Contextual Chunk Embeddings Using Long-Context Embedding Models}, 
      author={Michael Günther and Isabelle Mohr and Daniel James Williams and Bo Wang and Han Xiao},
      year={2025},
      eprint={2409.04701},
      archivePrefix={arXiv},
      primaryClass={cs.CL},
      url={https://arxiv.org/abs/2409.04701}, 
}

@inproceedings{gao-etal-2023-precise,
    title = "Precise Zero-Shot Dense Retrieval without Relevance Labels",
    author = "Gao, Luyu  and
      Ma, Xueguang  and
      Lin, Jimmy  and
      Callan, Jamie",
    editor = "Rogers, Anna  and
      Boyd-Graber, Jordan  and
      Okazaki, Naoaki",
    booktitle = "Proceedings of the 61st Annual Meeting of the Association for Computational Linguistics (Volume 1: Long Papers)",
    month = jul,
    year = "2023",
    address = "Toronto, Canada",
    publisher = "Association for Computational Linguistics",
    url = "https://aclanthology.org/2023.acl-long.99/",
    doi = "10.18653/v1/2023.acl-long.99",
    pages = "1762--1777"
}

@misc{sarthi2024raptorrecursiveabstractiveprocessing,
      title={RAPTOR: Recursive Abstractive Processing for Tree-Organized Retrieval}, 
      author={Parth Sarthi and Salman Abdullah and Aditi Tuli and Shubh Khanna and Anna Goldie and Christopher D. Manning},
      year={2024},
      eprint={2401.18059},
      archivePrefix={arXiv},
      primaryClass={cs.CL},
      url={https://arxiv.org/abs/2401.18059}, 
}

@misc{yu2024rankragunifyingcontextranking,
      title={RankRAG: Unifying Context Ranking with Retrieval-Augmented Generation in LLMs}, 
      author={Yue Yu and Wei Ping and Zihan Liu and Boxin Wang and Jiaxuan You and Chao Zhang and Mohammad Shoeybi and Bryan Catanzaro},
      year={2024},
      eprint={2407.02485},
      archivePrefix={arXiv},
      primaryClass={cs.CL},
      url={https://arxiv.org/abs/2407.02485}, 
}

@misc{anthropic2024contextualretrieval,
  author       = {{Anthropic}},
  title        = {Introducing Contextual Retrieval},
  year         = {2024},
  howpublished = {\url{https://www.anthropic.com/news/contextual-retrieval}},
  note         = {Accessed: 2025}
}
\clearpage

\section*{Complete Experimental Figures}


\clearpage

\section*{Experimental Tables}

\FloatBarrier
\subsection*{Chroma}

\begin{table}[!htbp]
	\centering
	\caption{Comparison of deduplication and filtering strategies on chroma for ClusterSemanticChunker (200, 0).}
	\label{tab:chroma_ClusterSemanticChunker_200_0_main}
	\begin{tabular}{lrrrrrr}
		\toprule
		Method & Threshold & Reduction (\%) & Precision & Recall & IoU & Oracle \\
		\midrule
		No filtering & - & 0.000 & 0.152 & 0.871 & 0.149 & 1.000 \\
		ExactNorm & - & 22.383 & 0.144 & 0.883 & 0.141 & 1.000 \\
		MinHash-LSH & 0.600 & 24.639 & 0.142 & 0.885 & 0.139 & 1.000 \\
		MinHash-LSH & 0.700 & 23.827 & 0.143 & 0.883 & 0.140 & 1.000 \\
		MinHash-LSH & 0.800 & 23.315 & 0.143 & 0.885 & 0.140 & 1.000 \\
		NER Exact & - & 30.445 & 0.149 & 0.928 & 0.147 & 0.410 \\
		NER Half & - & 63.718 & 0.131 & 0.691 & 0.115 & 0.232 \\
		Similarity & 0.800 & 34.386 & 0.145 & 0.910 & 0.142 & 0.380 \\
		Similarity + Topics & 0.800 & 33.424 & 0.146 & 0.911 & 0.143 & 0.383 \\
		Similarity + NER & 0.800 & 26.504 & 0.146 & 0.911 & 0.143 & 0.386 \\
		Similarity + Topics + NER & 0.800 & 25.782 & 0.156 & 0.898 & 0.143 & 0.389 \\
		random & 0.200 & 20.000 & 0.130 & 0.720 & 0.080 & 0.800 \\
		random & 0.400 & 40.000 & 0.110 & 0.540 & 0.060 & 0.600 \\
		random & 0.600 & 60.000 & 0.090 & 0.360 & 0.040 & 0.400 \\
		random & 0.800 & 80.000 & 0.070 & 0.180 & 0.020 & 0.200 \\
		\bottomrule
	\end{tabular}
\end{table}

\begin{table}[!htbp]
	\centering
	\caption{Comparison of deduplication and filtering strategies on chroma for ClusterSemanticChunker (400, 0).}
	\label{tab:chroma_ClusterSemanticChunker_400_0_main}
	\begin{tabular}{lrrrrrr}
		\toprule
		Method & Threshold & Reduction (\%) & Precision & Recall & IoU & Oracle \\
		\midrule
		No filtering & - & 0.000 & 0.100 & 0.903 & 0.099 & 1.000 \\
		ExactNorm & - & 21.298 & 0.096 & 0.909 & 0.095 & 1.000 \\
		MinHash-LSH & 0.600 & 25.903 & 0.093 & 0.913 & 0.093 & 1.000 \\
		MinHash-LSH & 0.700 & 24.908 & 0.094 & 0.912 & 0.093 & 1.000 \\
		MinHash-LSH & 0.800 & 23.967 & 0.094 & 0.910 & 0.093 & 1.000 \\
		NER Exact & - & 28.362 & 0.100 & 0.964 & 0.099 & 0.286 \\
		NER Half & - & 61.120 & 0.088 & 0.793 & 0.086 & 0.183 \\
		Similarity & 0.800 & 35.950 & 0.093 & 0.925 & 0.092 & 0.252 \\
		Similarity + Topics & 0.800 & 34.589 & 0.093 & 0.926 & 0.092 & 0.254 \\
		Similarity + NER & 0.800 & 27.996 & 0.093 & 0.927 & 0.092 & 0.256 \\
		Similarity + Topics + NER & 0.800 & 27.420 & 0.101 & 0.913 & 0.092 & 0.256 \\
		random & 0.200 & 20.000 & 0.130 & 0.720 & 0.080 & 0.800 \\
		random & 0.400 & 40.000 & 0.110 & 0.540 & 0.060 & 0.600 \\
		random & 0.600 & 60.000 & 0.090 & 0.360 & 0.040 & 0.400 \\
		random & 0.800 & 80.000 & 0.070 & 0.180 & 0.020 & 0.200 \\
		\bottomrule
	\end{tabular}
\end{table}

\begin{table}[!htbp]
	\centering
	\caption{Comparison of deduplication and filtering strategies on chroma for FixedTokenChunker (200, 0).}
	\label{tab:chroma_FixedTokenChunker_200_0_main}
	\begin{tabular}{lrrrrrr}
		\toprule
		Method & Threshold & Reduction (\%) & Precision & Recall & IoU & Oracle \\
		\midrule
		No filtering & - & 0.000 & 0.101 & 0.903 & 0.100 & 1.000 \\
		ExactNorm & - & 0.000 & 0.101 & 0.903 & 0.100 & 1.000 \\
		MinHash-LSH & 0.600 & 10.219 & 0.098 & 0.905 & 0.097 & 1.000 \\
		MinHash-LSH & 0.700 & 5.961 & 0.099 & 0.902 & 0.098 & 1.000 \\
		MinHash-LSH & 0.800 & 3.163 & 0.100 & 0.906 & 0.099 & 1.000 \\
		NER Exact & - & 1.825 & 0.111 & 0.958 & 0.110 & 0.290 \\
		NER Half & - & 50.243 & 0.096 & 0.901 & 0.095 & 0.233 \\
		Similarity & 0.800 & 31.934 & 0.098 & 0.931 & 0.097 & 0.259 \\
		Similarity + Topics & 0.800 & 29.684 & 0.099 & 0.931 & 0.098 & 0.259 \\
		Similarity + NER & 0.800 & 30.535 & 0.100 & 0.933 & 0.099 & 0.261 \\
		Similarity + Topics + NER & 0.800 & 28.224 & 0.100 & 0.933 & 0.099 & 0.262 \\
		random & 0.200 & 20.000 & 0.130 & 0.720 & 0.080 & 0.800 \\
		random & 0.400 & 40.000 & 0.110 & 0.540 & 0.060 & 0.600 \\
		random & 0.600 & 60.000 & 0.090 & 0.360 & 0.040 & 0.400 \\
		random & 0.800 & 80.000 & 0.070 & 0.180 & 0.020 & 0.200 \\
		\bottomrule
	\end{tabular}
\end{table}

\begin{table}[!htbp]
	\centering
	\caption{Comparison of deduplication and filtering strategies on chroma for FixedTokenChunker (400, 0).}
	\label{tab:chroma_FixedTokenChunker_400_0_main}
	\begin{tabular}{lrrrrrr}
		\toprule
		Method & Threshold & Reduction (\%) & Precision & Recall & IoU & Oracle \\
		\midrule
		No filtering & - & 0.000 & 0.061 & 0.916 & 0.061 & 1.000 \\
		ExactNorm & - & 0.000 & 0.061 & 0.916 & 0.061 & 1.000 \\
		MinHash-LSH & 0.600 & 9.223 & 0.060 & 0.919 & 0.060 & 1.000 \\
		MinHash-LSH & 0.700 & 6.917 & 0.060 & 0.918 & 0.060 & 1.000 \\
		MinHash-LSH & 0.800 & 3.762 & 0.061 & 0.917 & 0.060 & 1.000 \\
		NER Exact & - & 0.607 & 0.068 & 0.977 & 0.068 & 0.193 \\
		NER Half & - & 31.917 & 0.064 & 0.959 & 0.063 & 0.179 \\
		Similarity & 0.800 & 32.888 & 0.061 & 0.951 & 0.061 & 0.173 \\
		Similarity + Topics & 0.800 & 31.189 & 0.061 & 0.950 & 0.061 & 0.173 \\
		Similarity + NER & 0.800 & 32.160 & 0.061 & 0.954 & 0.061 & 0.175 \\
		Similarity + Topics + NER & 0.800 & 29.976 & 0.062 & 0.954 & 0.062 & 0.175 \\
		random & 0.200 & 20.000 & 0.130 & 0.720 & 0.080 & 0.800 \\
		random & 0.400 & 40.000 & 0.110 & 0.540 & 0.060 & 0.600 \\
		random & 0.600 & 60.000 & 0.090 & 0.359 & 0.040 & 0.399 \\
		random & 0.800 & 80.000 & 0.070 & 0.179 & 0.020 & 0.199 \\
		\bottomrule
	\end{tabular}
\end{table}

\begin{table}[!htbp]
	\centering
	\caption{Comparison of deduplication and filtering strategies on chroma for FixedTokenChunker (400, 200).}
	\label{tab:chroma_FixedTokenChunker_400_200_main}
	\begin{tabular}{lrrrrrr}
		\toprule
		Method & Threshold & Reduction (\%) & Precision & Recall & IoU & Oracle \\
		\midrule
		No filtering & - & 0.000 & 0.074 & 0.946 & 0.074 & 1.000 \\
		ExactNorm & - & 0.000 & 0.075 & 0.949 & 0.074 & 1.000 \\
		MinHash-LSH & 0.600 & 17.145 & 0.072 & 0.947 & 0.072 & 1.000 \\
		MinHash-LSH & 0.700 & 11.897 & 0.073 & 0.947 & 0.073 & 1.000 \\
		MinHash-LSH & 0.800 & 6.162 & 0.074 & 0.949 & 0.074 & 1.000 \\
		NER Exact & - & 1.281 & 0.085 & 0.980 & 0.085 & 0.206 \\
		NER Half & - & 61.074 & 0.067 & 0.958 & 0.067 & 0.178 \\
		Similarity & 0.800 & 58.328 & 0.066 & 0.961 & 0.065 & 0.183 \\
		Similarity + Topics & 0.800 & 54.057 & 0.067 & 0.965 & 0.067 & 0.185 \\
		Similarity + NER & 0.800 & 55.949 & 0.066 & 0.965 & 0.066 & 0.186 \\
		Similarity + Topics + NER & 0.800 & 53.935 & 0.067 & 0.967 & 0.067 & 0.186 \\
		random & 0.200 & 20.000 & 0.130 & 0.720 & 0.080 & 0.800 \\
		random & 0.400 & 40.000 & 0.110 & 0.540 & 0.060 & 0.600 \\
		random & 0.600 & 60.000 & 0.090 & 0.360 & 0.040 & 0.400 \\
		random & 0.800 & 80.000 & 0.070 & 0.180 & 0.020 & 0.200 \\
		\bottomrule
	\end{tabular}
\end{table}

\begin{table}[!htbp]
	\centering
	\caption{Comparison of deduplication and filtering strategies on chroma for FixedTokenChunker (800, 400).}
	\label{tab:chroma_FixedTokenChunker_800_400_main}
	\begin{tabular}{lrrrrrr}
		\toprule
		Method & Threshold & Reduction (\%) & Precision & Recall & IoU & Oracle \\
		\midrule
		No filtering & - & 0.000 & 0.043 & 0.946 & 0.043 & 1.000 \\
		ExactNorm & - & 0.000 & 0.043 & 0.946 & 0.043 & 1.000 \\
		MinHash-LSH & 0.600 & 14.408 & 0.042 & 0.945 & 0.042 & 1.000 \\
		MinHash-LSH & 0.700 & 10.012 & 0.042 & 0.946 & 0.042 & 1.000 \\
		MinHash-LSH & 0.800 & 6.227 & 0.043 & 0.947 & 0.043 & 1.000 \\
		NER Exact & - & 0.366 & 0.053 & 0.983 & 0.053 & 0.127 \\
		NER Half & - & 55.800 & 0.042 & 0.979 & 0.042 & 0.121 \\
		Similarity & 0.800 & 58.852 & 0.039 & 0.957 & 0.039 & 0.111 \\
		Similarity + Topics & 0.800 & 57.509 & 0.040 & 0.958 & 0.040 & 0.111 \\
		Similarity + NER & 0.800 & 55.067 & 0.040 & 0.958 & 0.040 & 0.111 \\
		Similarity + Topics + NER & 0.800 & 53.602 & 0.040 & 0.959 & 0.040 & 0.112 \\
		random & 0.200 & 20.000 & 0.130 & 0.720 & 0.080 & 0.800 \\
		random & 0.400 & 40.000 & 0.110 & 0.540 & 0.060 & 0.600 \\
		random & 0.600 & 60.000 & 0.090 & 0.359 & 0.040 & 0.399 \\
		random & 0.800 & 80.000 & 0.070 & 0.179 & 0.020 & 0.199 \\
		\bottomrule
	\end{tabular}
\end{table}

\begin{table}[!htbp]
	\centering
	\caption{Comparison of deduplication and filtering strategies on chroma for RecursiveTokenChunker (200, 0).}
	\label{tab:chroma_RecursiveTokenChunker_200_0_main}
	\begin{tabular}{lrrrrrr}
		\toprule
		Method & Threshold & Reduction (\%) & Precision & Recall & IoU & Oracle \\
		\midrule
		No filtering & - & 0.000 & 0.322 & 0.686 & 0.276 & 0.998 \\
		ExactNorm & - & 25.180 & 0.302 & 0.701 & 0.262 & 0.998 \\
		MinHash-LSH & 0.600 & 26.308 & 0.300 & 0.701 & 0.261 & 0.998 \\
		MinHash-LSH & 0.700 & 25.762 & 0.301 & 0.700 & 0.262 & 0.998 \\
		MinHash-LSH & 0.800 & 25.535 & 0.302 & 0.700 & 0.262 & 0.998 \\
		NER Exact & - & 54.791 & 0.299 & 0.711 & 0.264 & 0.651 \\
		NER Half & - & 52.580 & 0.280 & 0.572 & 0.224 & 0.479 \\
		Similarity & 0.800 & 32.833 & 0.330 & 0.750 & 0.289 & 0.785 \\
		Similarity + Topics & 0.800 & 31.786 & 0.330 & 0.749 & 0.290 & 0.787 \\
		Similarity + NER & 0.800 & 20.593 & 0.338 & 0.729 & 0.292 & 0.810 \\
		Similarity + Topics + NER & 0.800 & 20.020 & 0.342 & 0.747 & 0.298 & 0.815 \\
		random & 0.200 & 20.000 & 0.130 & 0.720 & 0.080 & 0.800 \\
		random & 0.400 & 40.000 & 0.110 & 0.540 & 0.060 & 0.600 \\
		random & 0.600 & 60.000 & 0.090 & 0.360 & 0.040 & 0.400 \\
		random & 0.800 & 80.000 & 0.070 & 0.180 & 0.020 & 0.200 \\
		\bottomrule
	\end{tabular}
\end{table}

\begin{table}[!htbp]
	\centering
	\caption{Comparison of deduplication and filtering strategies on chroma for RecursiveTokenChunker (400, 0).}
	\label{tab:chroma_RecursiveTokenChunker_400_0_main}
	\begin{tabular}{lrrrrrr}
		\toprule
		Method & Threshold & Reduction (\%) & Precision & Recall & IoU & Oracle \\
		\midrule
		No filtering & - & 0.000 & 0.227 & 0.805 & 0.213 & 1.000 \\
		ExactNorm & - & 21.851 & 0.213 & 0.817 & 0.203 & 1.000 \\
		MinHash-LSH & 0.600 & 23.214 & 0.212 & 0.817 & 0.201 & 1.000 \\
		MinHash-LSH & 0.700 & 22.600 & 0.213 & 0.819 & 0.203 & 1.000 \\
		MinHash-LSH & 0.800 & 22.197 & 0.213 & 0.818 & 0.203 & 1.000 \\
		NER Exact & - & 37.481 & 0.222 & 0.863 & 0.213 & 0.545 \\
		NER Half & - & 65.361 & 0.182 & 0.669 & 0.165 & 0.306 \\
		Similarity & 0.800 & 31.893 & 0.219 & 0.864 & 0.208 & 0.548 \\
		Similarity + Topics & 0.800 & 30.568 & 0.221 & 0.868 & 0.211 & 0.552 \\
		Similarity + NER & 0.800 & 25.653 & 0.224 & 0.868 & 0.213 & 0.559 \\
		Similarity + Topics + NER & 0.800 & 24.578 & 0.226 & 0.869 & 0.215 & 0.566 \\
		random & 0.200 & 20.000 & 0.130 & 0.720 & 0.080 & 0.800 \\
		random & 0.400 & 40.000 & 0.110 & 0.540 & 0.060 & 0.600 \\
		random & 0.600 & 60.000 & 0.090 & 0.360 & 0.040 & 0.400 \\
		random & 0.800 & 80.000 & 0.070 & 0.180 & 0.020 & 0.200 \\
		\bottomrule
	\end{tabular}
\end{table}

\begin{table}[!htbp]
	\centering
	\caption{Comparison of deduplication and filtering strategies on chroma for RecursiveTokenChunker (400, 200).}
	\label{tab:chroma_RecursiveTokenChunker_400_200_main}
	\begin{tabular}{lrrrrrr}
		\toprule
		Method & Threshold & Reduction (\%) & Precision & Recall & IoU & Oracle \\
		\midrule
		No filtering & - & 0.000 & 0.236 & 0.819 & 0.222 & 1.000 \\
		ExactNorm & - & 21.538 & 0.223 & 0.836 & 0.212 & 1.000 \\
		MinHash-LSH & 0.600 & 23.082 & 0.221 & 0.836 & 0.211 & 1.000 \\
		MinHash-LSH & 0.700 & 22.262 & 0.222 & 0.835 & 0.212 & 1.000 \\
		MinHash-LSH & 0.800 & 21.924 & 0.222 & 0.834 & 0.212 & 1.000 \\
		NER Exact & - & 35.725 & 0.232 & 0.866 & 0.222 & 0.574 \\
		NER Half & - & 71.337 & 0.183 & 0.697 & 0.167 & 0.321 \\
		Similarity & 0.800 & 43.606 & 0.216 & 0.858 & 0.206 & 0.526 \\
		Similarity + Topics & 0.800 & 41.387 & 0.217 & 0.861 & 0.208 & 0.535 \\
		Similarity + NER & 0.800 & 37.108 & 0.220 & 0.858 & 0.210 & 0.543 \\
		Similarity + Topics + NER & 0.800 & 35.226 & 0.223 & 0.860 & 0.213 & 0.554 \\
		random & 0.200 & 20.000 & 0.130 & 0.720 & 0.080 & 0.800 \\
		random & 0.400 & 40.000 & 0.110 & 0.540 & 0.060 & 0.600 \\
		random & 0.600 & 60.000 & 0.090 & 0.360 & 0.040 & 0.400 \\
		random & 0.800 & 80.000 & 0.070 & 0.180 & 0.020 & 0.200 \\
		\bottomrule
	\end{tabular}
\end{table}

\begin{table}[!htbp]
	\centering
	\caption{Comparison of deduplication and filtering strategies on chroma for RecursiveTokenChunker (800, 400).}
	\label{tab:chroma_RecursiveTokenChunker_800_400_main}
	\begin{tabular}{lrrrrrr}
		\toprule
		Method & Threshold & Reduction (\%) & Precision & Recall & IoU & Oracle \\
		\midrule
		No filtering & - & NaN & 0.000 & 0.000 & 0.000 & 0.000 \\
		ExactNorm & - & 20.811 & 0.147 & 0.892 & 0.145 & 1.000 \\
		MinHash-LSH & 0.600 & 23.303 & 0.146 & 0.896 & 0.144 & 1.000 \\
		MinHash-LSH & 0.700 & 22.583 & 0.146 & 0.895 & 0.144 & 1.000 \\
		MinHash-LSH & 0.800 & 21.832 & 0.147 & 0.893 & 0.144 & 1.000 \\
		NER Exact & - & 26.366 & 0.158 & 0.945 & 0.156 & 0.401 \\
		NER Half & - & 73.363 & 0.120 & 0.786 & 0.115 & 0.238 \\
		Similarity & 0.800 & 50.360 & 0.136 & 0.918 & 0.134 & 0.338 \\
		Similarity + Topics & 0.800 & 48.258 & 0.137 & 0.917 & 0.134 & 0.338 \\
		Similarity + NER & 0.800 & 47.087 & 0.139 & 0.917 & 0.137 & 0.344 \\
		Similarity + Topics + NER & 0.800 & 45.345 & 0.140 & 0.918 & 0.137 & 0.346 \\
		random & 0.200 & 20.000 & 0.130 & 0.720 & 0.080 & 0.800 \\
		random & 0.400 & 40.000 & 0.110 & 0.540 & 0.060 & 0.600 \\
		random & 0.600 & 60.000 & 0.090 & 0.360 & 0.040 & 0.400 \\
		random & 0.800 & 80.000 & 0.070 & 0.180 & 0.020 & 0.200 \\
		\bottomrule
	\end{tabular}
\end{table}

\FloatBarrier
\subsection*{SQuAD}

\begin{table}[!htbp]
	\centering
	\caption{Comparison of deduplication and filtering strategies on squad for ClusterSemanticChunker (200, 0).}
	\label{tab:squad_ClusterSemanticChunker_200_0_main}
	\begin{tabular}{llrrrrr}
		\toprule
		Method & Threshold & Reduction (\%) & Precision & Recall & IoU & Oracle \\
		\midrule
		No filtering & - & 0.000 & 0.105 & 0.901 & 0.104 & 1.000 \\
		ExactNorm & - & 8.058 & 0.105 & 0.902 & 0.104 & 1.000 \\
		MinHash-LSH & 0.600 & 8.058 & 0.105 & 0.902 & 0.104 & 1.000 \\
		MinHash-LSH & 0.700 & 8.058 & 0.105 & 0.902 & 0.104 & 1.000 \\
		MinHash-LSH & 0.800 & 8.058 & 0.105 & 0.902 & 0.104 & 1.000 \\
		NER Exact & - & 20.352 & 0.105 & 0.892 & 0.104 & 0.348 \\
		NER Half & - & 48.625 & 0.105 & 0.645 & 0.090 & 0.209 \\
		Similarity & 0.800 & 9.406 & 0.112 & 0.905 & 0.111 & 0.372 \\
		Similarity + Topics & 0.800 & 9.406 & 0.112 & 0.904 & 0.111 & 0.372 \\
		Similarity + NER & 0.800 & 9.406 & 0.112 & 0.904 & 0.111 & 0.372 \\
		Similarity + Topics + NER & 0.800 & 9.406 & 0.112 & 0.904 & 0.111 & 0.372 \\
		Random & 0.200 & 20.000 & 0.130 & 0.720 & 0.080 & 0.800 \\
		Random & 0.400 & 40.000 & 0.110 & 0.540 & 0.060 & 0.600 \\
		Random & 0.600 & 60.000 & 0.090 & 0.360 & 0.040 & 0.400 \\
		Random & 0.800 & 80.000 & 0.070 & 0.180 & 0.020 & 0.200 \\
		\bottomrule
	\end{tabular}
\end{table}

\begin{table}[!htbp]
	\centering
	\caption{Comparison of deduplication and filtering strategies on squad for ClusterSemanticChunker (400, 0).}
	\label{tab:squad_ClusterSemanticChunker_400_0_main}
	\begin{tabular}{llrrrrr}
		\toprule
		Method & Threshold & Reduction (\%) & Precision & Recall & IoU & Oracle \\
		\midrule
		No filtering & - & 0.000 & 0.071 & 0.902 & 0.070 & 1.000 \\
		ExactNorm & - & 11.646 & 0.071 & 0.903 & 0.070 & 1.000 \\
		MinHash-LSH & 0.600 & 11.646 & 0.071 & 0.903 & 0.070 & 1.000 \\
		MinHash-LSH & 0.700 & 11.646 & 0.071 & 0.903 & 0.070 & 1.000 \\
		MinHash-LSH & 0.800 & 11.646 & 0.071 & 0.903 & 0.070 & 1.000 \\
		NER Exact & - & 22.616 & 0.072 & 0.909 & 0.071 & 0.244 \\
		NER Half & - & 45.443 & 0.090 & 0.626 & 0.070 & 0.146 \\
		Similarity & 0.800 & 13.165 & 0.076 & 0.903 & 0.075 & 0.251 \\
		Similarity + Topics & 0.800 & 13.165 & 0.076 & 0.903 & 0.075 & 0.251 \\
		Similarity + NER & 0.800 & 13.165 & 0.077 & 0.903 & 0.075 & 0.251 \\
		Similarity + Topics + NER & 0.800 & 13.165 & 0.076 & 0.904 & 0.075 & 0.251 \\
		Random & 0.200 & 20.000 & 0.130 & 0.720 & 0.080 & 0.800 \\
		Random & 0.400 & 40.000 & 0.110 & 0.540 & 0.060 & 0.600 \\
		Random & 0.600 & 60.000 & 0.090 & 0.360 & 0.040 & 0.400 \\
		Random & 0.800 & 80.000 & 0.070 & 0.180 & 0.020 & 0.200 \\
		\bottomrule
	\end{tabular}
\end{table}

\begin{table}[!htbp]
	\centering
	\caption{Comparison of deduplication and filtering strategies on squad for FixedTokenChunker (200, 0).}
	\label{tab:squad_FixedTokenChunker_200_0_main}
	\begin{tabular}{llrrrrr}
		\toprule
		Method & Threshold & Reduction (\%) & Precision & Recall & IoU & Oracle \\
		\midrule
		No filtering & - & 0.000 & 0.062 & 0.905 & 0.062 & 1.000 \\
		ExactNorm & - & 0.000 & 0.062 & 0.905 & 0.062 & 1.000 \\
		MinHash-LSH & 0.600 & 0.000 & 0.062 & 0.905 & 0.062 & 1.000 \\
		MinHash-LSH & 0.700 & 0.000 & 0.062 & 0.905 & 0.062 & 1.000 \\
		MinHash-LSH & 0.800 & 0.000 & 0.062 & 0.905 & 0.062 & 1.000 \\
		NER Exact & - & 1.950 & 0.067 & 0.908 & 0.066 & 0.235 \\
		NER Half & - & 45.449 & 0.056 & 0.780 & 0.055 & 0.156 \\
		Similarity & 0.800 & 2.009 & 0.067 & 0.910 & 0.066 & 0.236 \\
		Similarity + Topics & 0.800 & 2.009 & 0.067 & 0.910 & 0.066 & 0.236 \\
		Similarity + NER & 0.800 & 2.009 & 0.067 & 0.910 & 0.066 & 0.236 \\
		Similarity + Topics + NER & 0.800 & 2.009 & 0.067 & 0.910 & 0.066 & 0.236 \\
		Random & 0.200 & 20.000 & 0.130 & 0.720 & 0.080 & 0.800 \\
		Random & 0.400 & 40.000 & 0.110 & 0.540 & 0.060 & 0.600 \\
		Random & 0.600 & 60.000 & 0.090 & 0.360 & 0.040 & 0.400 \\
		Random & 0.800 & 80.000 & 0.070 & 0.180 & 0.020 & 0.200 \\
		\bottomrule
	\end{tabular}
\end{table}

\begin{table}[!htbp]
	\centering
	\caption{Comparison of deduplication and filtering strategies on squad for FixedTokenChunker (400, 0).}
	\label{tab:squad_FixedTokenChunker_400_0_main}
	\begin{tabular}{llrrrrr}
		\toprule
		Method & Threshold & Reduction (\%) & Precision & Recall & IoU & Oracle \\
		\midrule
		No filtering & - & 0.000 & 0.036 & 0.891 & 0.036 & 1.000 \\
		ExactNorm & - & 0.000 & 0.036 & 0.891 & 0.036 & 1.000 \\
		MinHash-LSH & 0.600 & 0.000 & 0.036 & 0.891 & 0.036 & 1.000 \\
		MinHash-LSH & 0.700 & 0.000 & 0.036 & 0.891 & 0.036 & 1.000 \\
		MinHash-LSH & 0.800 & 0.000 & 0.036 & 0.891 & 0.036 & 1.000 \\
		NER Exact & - & 0.815 & 0.039 & 0.901 & 0.039 & 0.147 \\
		NER Half & - & 46.217 & 0.034 & 0.794 & 0.034 & 0.100 \\
		Similarity & 0.800 & 0.931 & 0.039 & 0.900 & 0.039 & 0.147 \\
		Similarity + Topics & 0.800 & 0.931 & 0.039 & 0.900 & 0.039 & 0.147 \\
		Similarity + NER & 0.800 & 0.931 & 0.039 & 0.900 & 0.039 & 0.147 \\
		Similarity + Topics + NER & 0.800 & 0.931 & 0.039 & 0.900 & 0.039 & 0.147 \\
		Random & 0.200 & 20.000 & 0.130 & 0.720 & 0.080 & 0.800 \\
		Random & 0.400 & 40.000 & 0.110 & 0.540 & 0.060 & 0.600 \\
		Random & 0.600 & 60.000 & 0.090 & 0.359 & 0.040 & 0.399 \\
		Random & 0.800 & 80.000 & 0.070 & 0.179 & 0.020 & 0.199 \\
		\bottomrule
	\end{tabular}
\end{table}

\begin{table}[!htbp]
	\centering
	\caption{Comparison of deduplication and filtering strategies on squad for FixedTokenChunker (400, 200).}
	\label{tab:squad_FixedTokenChunker_400_200_main}
	\begin{tabular}{llrrrrr}
		\toprule
		Method & Threshold & Reduction (\%) & Precision & Recall & IoU & Oracle \\
		\midrule
		No filtering & - & 0.000 & 0.043 & 0.934 & 0.043 & 1.000 \\
		ExactNorm & - & 0.000 & 0.043 & 0.934 & 0.043 & 1.000 \\
		MinHash-LSH & 0.600 & 0.000 & 0.043 & 0.934 & 0.043 & 1.000 \\
		MinHash-LSH & 0.700 & 0.000 & 0.043 & 0.934 & 0.043 & 1.000 \\
		MinHash-LSH & 0.800 & 0.000 & 0.043 & 0.934 & 0.043 & 1.000 \\
		NER Exact & - & 0.669 & 0.046 & 0.940 & 0.046 & 0.153 \\
		NER Half & - & 68.796 & 0.035 & 0.802 & 0.034 & 0.101 \\
		Similarity & 0.800 & 6.934 & 0.045 & 0.937 & 0.045 & 0.152 \\
		Similarity + Topics & 0.800 & 6.934 & 0.045 & 0.937 & 0.045 & 0.152 \\
		Similarity + NER & 0.800 & 6.934 & 0.045 & 0.937 & 0.045 & 0.152 \\
		Similarity + Topics + NER & 0.800 & 6.934 & 0.045 & 0.937 & 0.045 & 0.152 \\
		Random & 0.200 & 20.000 & 0.130 & 0.720 & 0.080 & 0.800 \\
		Random & 0.400 & 40.000 & 0.110 & 0.540 & 0.060 & 0.600 \\
		Random & 0.600 & 60.000 & 0.090 & 0.360 & 0.040 & 0.400 \\
		Random & 0.800 & 80.000 & 0.070 & 0.180 & 0.020 & 0.200 \\
		\bottomrule
	\end{tabular}
\end{table}

\begin{table}[!htbp]
	\centering
	\caption{Comparison of deduplication and filtering strategies on squad for FixedTokenChunker (800, 400).}
	\label{tab:squad_FixedTokenChunker_800_400_main}
	\begin{tabular}{llrrrrr}
		\toprule
		Method & Threshold & Reduction (\%) & Precision & Recall & IoU & Oracle \\
		\midrule
		No filtering & - & 0.000 & 0.025 & 0.919 & 0.025 & 1.000 \\
		ExactNorm & - & 0.000 & 0.025 & 0.919 & 0.025 & 1.000 \\
		MinHash-LSH & 0.600 & 0.000 & 0.025 & 0.919 & 0.025 & 1.000 \\
		MinHash-LSH & 0.700 & 0.000 & 0.025 & 0.919 & 0.025 & 1.000 \\
		MinHash-LSH & 0.800 & 0.000 & 0.025 & 0.919 & 0.025 & 1.000 \\
		NER Exact & - & 0.000 & 0.027 & 0.928 & 0.027 & 0.089 \\
		NER Half & - & 54.007 & 0.023 & 0.860 & 0.023 & 0.076 \\
		Similarity & 0.800 & 0.863 & 0.027 & 0.927 & 0.027 & 0.089 \\
		Similarity + Topics & 0.800 & 0.863 & 0.027 & 0.927 & 0.027 & 0.089 \\
		Similarity + NER & 0.800 & 0.863 & 0.027 & 0.927 & 0.027 & 0.089 \\
		Similarity + Topics + NER & 0.800 & 0.863 & 0.027 & 0.927 & 0.027 & 0.089 \\
		Random & 0.200 & 20.000 & 0.130 & 0.719 & 0.080 & 0.799 \\
		Random & 0.400 & 40.000 & 0.110 & 0.539 & 0.060 & 0.599 \\
		Random & 0.600 & 60.000 & 0.090 & 0.360 & 0.040 & 0.400 \\
		Random & 0.800 & 80.000 & 0.070 & 0.180 & 0.020 & 0.200 \\
		\bottomrule
	\end{tabular}
\end{table}

\begin{table}[!htbp]
	\centering
	\caption{Comparison of deduplication and filtering strategies on squad for RecursiveTokenChunker (200, 0).}
	\label{tab:squad_RecursiveTokenChunker_200_0_main}
	\begin{tabular}{llrrrrr}
		\toprule
		Method & Threshold & Reduction (\%) & Precision & Recall & IoU & Oracle \\
		\midrule
		No filtering & - & 0.000 & 0.269 & 0.809 & 0.253 & 1.000 \\
		ExactNorm & - & 4.053 & 0.269 & 0.809 & 0.253 & 1.000 \\
		MinHash-LSH & 0.600 & 4.102 & 0.269 & 0.810 & 0.253 & 1.000 \\
		MinHash-LSH & 0.700 & 4.069 & 0.269 & 0.810 & 0.253 & 1.000 \\
		MinHash-LSH & 0.800 & 4.069 & 0.269 & 0.810 & 0.253 & 1.000 \\
		NER Exact & - & 36.289 & 0.232 & 0.735 & 0.216 & 0.618 \\
		NER Half & - & 47.189 & 0.216 & 0.585 & 0.188 & 0.463 \\
		Similarity & 0.800 & 6.137 & 0.280 & 0.816 & 0.263 & 0.811 \\
		Similarity + Topics & 0.800 & 6.137 & 0.280 & 0.816 & 0.263 & 0.811 \\
		Similarity + NER & 0.800 & 6.137 & 0.280 & 0.816 & 0.263 & 0.811 \\
		Similarity + Topics + NER & 0.800 & 6.137 & 0.280 & 0.816 & 0.263 & 0.811 \\
		Random & 0.200 & 20.000 & 0.130 & 0.720 & 0.080 & 0.800 \\
		Random & 0.400 & 40.000 & 0.110 & 0.540 & 0.060 & 0.600 \\
		Random & 0.600 & 60.000 & 0.090 & 0.360 & 0.040 & 0.400 \\
		Random & 0.800 & 80.000 & 0.070 & 0.180 & 0.020 & 0.200 \\
		\bottomrule
	\end{tabular}
\end{table}

\begin{table}[!htbp]
	\centering
	\caption{Comparison of deduplication and filtering strategies on squad for RecursiveTokenChunker (400, 0).}
	\label{tab:squad_RecursiveTokenChunker_400_0_main}
	\begin{tabular}{llrrrrr}
		\toprule
		Method & Threshold & Reduction (\%) & Precision & Recall & IoU & Oracle \\
		\midrule
		No filtering & - & 0.000 & 0.164 & 0.895 & 0.162 & 1.000 \\
		ExactNorm & - & 1.028 & 0.164 & 0.895 & 0.162 & 1.000 \\
		MinHash-LSH & 0.600 & 1.028 & 0.164 & 0.896 & 0.162 & 1.000 \\
		MinHash-LSH & 0.700 & 1.028 & 0.164 & 0.895 & 0.162 & 1.000 \\
		MinHash-LSH & 0.800 & 1.028 & 0.164 & 0.896 & 0.162 & 1.000 \\
		NER Exact & - & 17.309 & 0.159 & 0.848 & 0.155 & 0.522 \\
		NER Half & - & 55.255 & 0.130 & 0.655 & 0.122 & 0.325 \\
		Similarity & 0.800 & 3.347 & 0.173 & 0.897 & 0.170 & 0.595 \\
		Similarity + Topics & 0.800 & 3.347 & 0.173 & 0.896 & 0.170 & 0.595 \\
		Similarity + NER & 0.800 & 3.347 & 0.173 & 0.896 & 0.170 & 0.595 \\
		Similarity + Topics + NER & 0.800 & 3.347 & 0.173 & 0.897 & 0.170 & 0.595 \\
		Random & 0.200 & 20.000 & 0.130 & 0.720 & 0.080 & 0.800 \\
		Random & 0.400 & 40.000 & 0.110 & 0.540 & 0.060 & 0.600 \\
		Random & 0.600 & 60.000 & 0.090 & 0.360 & 0.040 & 0.400 \\
		Random & 0.800 & 80.000 & 0.070 & 0.180 & 0.020 & 0.200 \\
		\bottomrule
	\end{tabular}
\end{table}

\begin{table}[!htbp]
	\centering
	\caption{Comparison of deduplication and filtering strategies on squad for RecursiveTokenChunker (400, 200).}
	\label{tab:squad_RecursiveTokenChunker_400_200_main}
	\begin{tabular}{llrrrrr}
		\toprule
		Method & Threshold & Reduction (\%) & Precision & Recall & IoU & Oracle \\
		\midrule
		No filtering & - & 0.000 & 0.167 & 0.899 & 0.164 & 1.000 \\
		ExactNorm & - & 0.800 & 0.167 & 0.899 & 0.164 & 1.000 \\
		MinHash-LSH & 0.600 & 0.800 & 0.167 & 0.899 & 0.164 & 1.000 \\
		MinHash-LSH & 0.700 & 0.800 & 0.167 & 0.899 & 0.164 & 1.000 \\
		MinHash-LSH & 0.800 & 0.800 & 0.167 & 0.899 & 0.164 & 1.000 \\
		NER Exact & - & 15.642 & 0.162 & 0.856 & 0.159 & 0.520 \\
		NER Half & - & 61.105 & 0.125 & 0.669 & 0.119 & 0.315 \\
		Similarity & 0.800 & 14.752 & 0.168 & 0.890 & 0.165 & 0.558 \\
		Similarity + Topics & 0.800 & 14.752 & 0.167 & 0.890 & 0.165 & 0.558 \\
		Similarity + NER & 0.800 & 14.752 & 0.168 & 0.890 & 0.165 & 0.558 \\
		Similarity + Topics + NER & 0.800 & 14.752 & 0.167 & 0.890 & 0.164 & 0.558 \\
		Random & 0.200 & 20.000 & 0.130 & 0.720 & 0.080 & 0.800 \\
		Random & 0.400 & 40.000 & 0.110 & 0.540 & 0.060 & 0.600 \\
		Random & 0.600 & 60.000 & 0.090 & 0.360 & 0.040 & 0.400 \\
		Random & 0.800 & 80.000 & 0.070 & 0.180 & 0.020 & 0.200 \\
		\bottomrule
	\end{tabular}
\end{table}

\begin{table}[!htbp]
	\centering
	\caption{Comparison of deduplication and filtering strategies on squad for RecursiveTokenChunker (800, 400).}
	\label{tab:squad_RecursiveTokenChunker_800_400_main}
	\begin{tabular}{lllrrrr}
		\toprule
		Method & Threshold & Reduction (\%) & Precision & Recall & IoU & Oracle \\
		\midrule
		No filtering & - & - & 0.096 & 0.926 & 0.095 & 1.000 \\
		ExactNorm & - & 0.000 & 0.096 & 0.926 & 0.095 & 1.000 \\
		MinHash-LSH & 0.600 & 0.033 & 0.096 & 0.926 & 0.095 & 1.000 \\
		MinHash-LSH & 0.700 & 0.033 & 0.096 & 0.926 & 0.095 & 1.000 \\
		MinHash-LSH & 0.800 & 0.000 & 0.096 & 0.926 & 0.095 & 1.000 \\
		NER Exact & - & 6.817 & 0.099 & 0.911 & 0.098 & 0.352 \\
		NER Half & - & 55.490 & 0.080 & 0.748 & 0.078 & 0.225 \\
		Similarity & 0.800 & 16.618 & 0.097 & 0.921 & 0.096 & 0.353 \\
		Similarity + Topics & 0.800 & 16.618 & 0.097 & 0.920 & 0.096 & 0.353 \\
		Similarity + NER & 0.800 & 16.618 & 0.097 & 0.920 & 0.096 & 0.353 \\
		Similarity + Topics + NER & 0.800 & 16.618 & 0.096 & 0.920 & 0.096 & 0.353 \\
		Random & 0.200 & 20.000 & 0.130 & 0.720 & 0.080 & 0.800 \\
		Random & 0.400 & 40.000 & 0.110 & 0.540 & 0.060 & 0.600 \\
		Random & 0.600 & 60.000 & 0.090 & 0.360 & 0.040 & 0.400 \\
		Random & 0.800 & 80.000 & 0.070 & 0.180 & 0.020 & 0.200 \\
		\bottomrule
	\end{tabular}
\end{table}

\FloatBarrier
\subsection*{WebFaq}

\begin{table}[!htbp]
	\centering
	\caption{Comparison of deduplication and filtering strategies on webfaq for ClusterSemanticChunker (200, 0).}
	\label{tab:webfaq_ClusterSemanticChunker_200_0_main}
	\begin{tabular}{llrrrrr}
		\toprule
		Method & Threshold & Reduction (\%) & Precision & Recall & IoU & Oracle \\
		\midrule
		No filtering & - & 0.000 & 0.889 & 0.487 & 0.469 & 0.701 \\
		ExactNorm & - & 5.477 & 0.889 & 0.489 & 0.472 & 0.701 \\
		MinHash-LSH & 0.600 & 6.272 & 0.889 & 0.495 & 0.477 & 0.700 \\
		MinHash-LSH & 0.700 & 6.184 & 0.889 & 0.495 & 0.477 & 0.700 \\
		MinHash-LSH & 0.800 & 6.095 & 0.889 & 0.495 & 0.478 & 0.700 \\
		NER Exact & - & 18.551 & 0.879 & 0.499 & 0.477 & 0.999 \\
		NER Half & - & 21.201 & 0.817 & 0.430 & 0.407 & 0.999 \\
		Similarity & 0.800 & 14.223 & 0.871 & 0.485 & 0.465 & 0.999 \\
		Similarity + Topics & 0.800 & 13.781 & 0.872 & 0.484 & 0.464 & 0.999 \\
		Similarity + NER & 0.800 & 5.300 & 0.887 & 0.491 & 0.473 & 0.999 \\
		Similarity + Topics + NER & 0.800 & 5.212 & 0.887 & 0.491 & 0.473 & 0.999 \\
		Random & 0.200 & 20.000 & 0.130 & 0.720 & 0.080 & 0.799 \\
		Random & 0.400 & 40.000 & 0.110 & 0.540 & 0.060 & 0.600 \\
		Random & 0.600 & 60.000 & 0.090 & 0.359 & 0.040 & 0.399 \\
		Random & 0.800 & 80.000 & 0.070 & 0.180 & 0.020 & 0.200 \\
		\bottomrule
	\end{tabular}
\end{table}

\begin{table}[!htbp]
	\centering
	\caption{Comparison of deduplication and filtering strategies on webfaq for ClusterSemanticChunker (400, 0).}
	\label{tab:webfaq_ClusterSemanticChunker_400_0_main}
	\begin{tabular}{llrrrrr}
		\toprule
		Method & Threshold & Reduction (\%) & Precision & Recall & IoU & Oracle \\
		\midrule
		No filtering & - & 0.000 & 0.832 & 0.615 & 0.573 & 0.873 \\
		ExactNorm & - & 7.028 & 0.830 & 0.613 & 0.571 & 0.873 \\
		MinHash-LSH & 0.600 & 7.522 & 0.829 & 0.613 & 0.570 & 0.873 \\
		MinHash-LSH & 0.700 & 7.522 & 0.829 & 0.613 & 0.570 & 0.873 \\
		MinHash-LSH & 0.800 & 7.522 & 0.829 & 0.615 & 0.572 & 0.873 \\
		NER Exact & - & 21.825 & 0.840 & 0.621 & 0.575 & 0.999 \\
		NER Half & - & 21.578 & 0.727 & 0.459 & 0.419 & 0.999 \\
		Similarity & 0.800 & 16.276 & 0.812 & 0.582 & 0.534 & 0.999 \\
		Similarity + Topics & 0.800 & 16.030 & 0.815 & 0.586 & 0.539 & 0.999 \\
		Similarity + NER & 0.800 & 5.919 & 0.818 & 0.592 & 0.544 & 0.999 \\
		Similarity + Topics + NER & 0.800 & 5.672 & 0.822 & 0.599 & 0.550 & 0.999 \\
		Random & 0.200 & 20.000 & 0.130 & 0.719 & 0.080 & 0.799 \\
		Random & 0.400 & 40.000 & 0.110 & 0.539 & 0.060 & 0.599 \\
		Random & 0.600 & 60.000 & 0.090 & 0.360 & 0.040 & 0.400 \\
		Random & 0.800 & 80.000 & 0.070 & 0.180 & 0.020 & 0.200 \\
		\bottomrule
	\end{tabular}
\end{table}

\begin{table}[!htbp]
	\centering
	\caption{Comparison of deduplication and filtering strategies on webfaq for FixedTokenChunker (200, 0).}
	\label{tab:webfaq_FixedTokenChunker_200_0_main}
	\begin{tabular}{llrrrrr}
		\toprule
		Method & Threshold & Reduction (\%) & Precision & Recall & IoU & Oracle \\
		\midrule
		No filtering & - & 0.000 & 0.793 & 0.687 & 0.600 & 0.855 \\
		ExactNorm & - & 0.000 & 0.791 & 0.688 & 0.601 & 0.855 \\
		MinHash-LSH & 0.600 & 0.421 & 0.793 & 0.689 & 0.602 & 0.855 \\
		MinHash-LSH & 0.700 & 0.211 & 0.793 & 0.689 & 0.602 & 0.855 \\
		MinHash-LSH & 0.800 & 0.211 & 0.793 & 0.689 & 0.602 & 0.855 \\
		NER Exact & - & 0.842 & 0.792 & 0.690 & 0.600 & 0.990 \\
		NER Half & - & 14.737 & 0.731 & 0.643 & 0.537 & 0.990 \\
		Similarity & 0.800 & 20.632 & 0.739 & 0.654 & 0.545 & 0.990 \\
		Similarity + Topics & 0.800 & 20.632 & 0.739 & 0.654 & 0.545 & 0.990 \\
		Similarity + NER & 0.800 & 13.895 & 0.750 & 0.664 & 0.556 & 0.990 \\
		Similarity + Topics + NER & 0.800 & 13.895 & 0.750 & 0.664 & 0.556 & 0.990 \\
		Random & 0.200 & 20.000 & 0.130 & 0.720 & 0.080 & 0.800 \\
		Random & 0.400 & 40.000 & 0.110 & 0.540 & 0.060 & 0.600 \\
		Random & 0.600 & 60.000 & 0.090 & 0.360 & 0.040 & 0.400 \\
		Random & 0.800 & 80.000 & 0.070 & 0.178 & 0.020 & 0.198 \\
		\bottomrule
	\end{tabular}
\end{table}

\begin{table}[!htbp]
	\centering
	\caption{Comparison of deduplication and filtering strategies on webfaq for FixedTokenChunker (400, 0).}
	\label{tab:webfaq_FixedTokenChunker_400_0_main}
	\begin{tabular}{llrrrrr}
		\toprule
		Method & Threshold & Reduction (\%) & Precision & Recall & IoU & Oracle \\
		\midrule
		No filtering & - & 0.000 & 0.586 & 0.813 & 0.517 & 0.981 \\
		ExactNorm & - & 0.000 & 0.586 & 0.813 & 0.517 & 0.981 \\
		MinHash-LSH & 0.600 & 0.397 & 0.584 & 0.813 & 0.515 & 0.981 \\
		MinHash-LSH & 0.700 & 0.000 & 0.586 & 0.813 & 0.517 & 0.981 \\
		MinHash-LSH & 0.800 & 0.000 & 0.586 & 0.813 & 0.517 & 0.981 \\
		NER Exact & - & 0.794 & 0.577 & 0.817 & 0.514 & 0.995 \\
		NER Half & - & 7.143 & 0.551 & 0.770 & 0.479 & 0.995 \\
		Similarity & 0.800 & 14.286 & 0.554 & 0.786 & 0.481 & 0.995 \\
		Similarity + Topics & 0.800 & 13.889 & 0.557 & 0.789 & 0.484 & 0.995 \\
		Similarity + NER & 0.800 & 11.111 & 0.557 & 0.786 & 0.482 & 0.995 \\
		Similarity + Topics + NER & 0.800 & 10.317 & 0.563 & 0.793 & 0.489 & 0.995 \\
		Random & 0.200 & 20.000 & 0.130 & 0.718 & 0.080 & 0.798 \\
		Random & 0.400 & 40.000 & 0.110 & 0.539 & 0.060 & 0.599 \\
		Random & 0.600 & 60.000 & 0.090 & 0.357 & 0.040 & 0.397 \\
		Random & 0.800 & 80.000 & 0.070 & 0.179 & 0.020 & 0.198 \\
		\bottomrule
	\end{tabular}
\end{table}

\begin{table}[!htbp]
	\centering
	\caption{Comparison of deduplication and filtering strategies on webfaq for FixedTokenChunker (400, 200).}
	\label{tab:webfaq_FixedTokenChunker_400_200_main}
	\begin{tabular}{llrrrrr}
		\toprule
		Method & Threshold & Reduction (\%) & Precision & Recall & IoU & Oracle \\
		\midrule
		No filtering & - & 0.000 & 0.682 & 0.847 & 0.628 & 0.979 \\
		ExactNorm & - & 0.000 & 0.682 & 0.847 & 0.628 & 0.979 \\
		MinHash-LSH & 0.600 & 0.489 & 0.678 & 0.847 & 0.624 & 0.979 \\
		MinHash-LSH & 0.700 & 0.244 & 0.684 & 0.849 & 0.628 & 0.979 \\
		MinHash-LSH & 0.800 & 0.244 & 0.679 & 0.847 & 0.625 & 0.979 \\
		NER Exact & - & 0.244 & 0.686 & 0.850 & 0.632 & 0.994 \\
		NER Half & - & 36.919 & 0.532 & 0.814 & 0.484 & 0.994 \\
		Similarity & 0.800 & 25.672 & 0.590 & 0.824 & 0.540 & 0.995 \\
		Similarity + Topics & 0.800 & 25.428 & 0.593 & 0.824 & 0.543 & 0.995 \\
		Similarity + NER & 0.800 & 22.249 & 0.598 & 0.827 & 0.546 & 0.994 \\
		Similarity + Topics + NER & 0.800 & 22.005 & 0.602 & 0.829 & 0.550 & 0.994 \\
		Random & 0.200 & 20.000 & 0.130 & 0.720 & 0.080 & 0.800 \\
		Random & 0.400 & 40.000 & 0.110 & 0.539 & 0.060 & 0.599 \\
		Random & 0.600 & 60.000 & 0.090 & 0.359 & 0.040 & 0.399 \\
		Random & 0.800 & 80.000 & 0.070 & 0.178 & 0.020 & 0.198 \\
		\bottomrule
	\end{tabular}
\end{table}

\begin{table}[!htbp]
	\centering
	\caption{Comparison of deduplication and filtering strategies on webfaq for FixedTokenChunker (800, 400).}
	\label{tab:webfaq_FixedTokenChunker_800_400_main}
	\begin{tabular}{llrrrrr}
		\toprule
		Method & Threshold & Reduction (\%) & Precision & Recall & IoU & Oracle \\
		\midrule
		No filtering & - & 0.000 & 0.390 & 0.893 & 0.378 & 1.000 \\
		ExactNorm & - & 0.000 & 0.390 & 0.893 & 0.378 & 1.000 \\
		MinHash-LSH & 0.600 & 1.075 & 0.389 & 0.893 & 0.376 & 1.000 \\
		MinHash-LSH & 0.700 & 0.538 & 0.390 & 0.893 & 0.377 & 1.000 \\
		MinHash-LSH & 0.800 & 0.538 & 0.390 & 0.893 & 0.377 & 1.000 \\
		NER Exact & - & 0.000 & 0.395 & 0.895 & 0.382 & 0.997 \\
		NER Half & - & 34.409 & 0.324 & 0.845 & 0.311 & 0.998 \\
		Similarity & 0.800 & 15.054 & 0.364 & 0.881 & 0.351 & 0.998 \\
		Similarity + Topics & 0.800 & 15.054 & 0.364 & 0.881 & 0.351 & 0.998 \\
		Similarity + NER & 0.800 & 14.516 & 0.368 & 0.885 & 0.355 & 0.998 \\
		Similarity + Topics + NER & 0.800 & 14.516 & 0.368 & 0.885 & 0.355 & 0.998 \\
		Random & 0.200 & 20.000 & 0.130 & 0.716 & 0.080 & 0.796 \\
		Random & 0.400 & 40.000 & 0.110 & 0.537 & 0.060 & 0.597 \\
		Random & 0.600 & 60.000 & 0.090 & 0.358 & 0.040 & 0.398 \\
		Random & 0.800 & 80.000 & 0.070 & 0.179 & 0.020 & 0.199 \\
		\bottomrule
	\end{tabular}
\end{table}

\begin{table}[!htbp]
	\centering
	\caption{Comparison of deduplication and filtering strategies on webfaq for RecursiveTokenChunker (200, 0).}
	\label{tab:webfaq_RecursiveTokenChunker_200_0_main}
	\begin{tabular}{llrrrrr}
		\toprule
		Method & Threshold & Reduction (\%) & Precision & Recall & IoU & Oracle \\
		\midrule
		No filtering & - & 0.000 & 0.907 & 0.230 & 0.227 & 0.375 \\
		ExactNorm & - & 2.737 & 0.907 & 0.232 & 0.230 & 0.375 \\
		MinHash-LSH & 0.600 & 3.604 & 0.907 & 0.233 & 0.231 & 0.374 \\
		MinHash-LSH & 0.700 & 3.239 & 0.907 & 0.233 & 0.231 & 0.374 \\
		MinHash-LSH & 0.800 & 3.193 & 0.907 & 0.233 & 0.231 & 0.374 \\
		NER Exact & - & 20.438 & 0.901 & 0.248 & 0.245 & 1.000 \\
		NER Half & - & 25.958 & 0.863 & 0.230 & 0.225 & 1.000 \\
		Similarity & 0.800 & 10.538 & 0.897 & 0.240 & 0.237 & 1.000 \\
		Similarity + Topics & 0.800 & 10.036 & 0.895 & 0.240 & 0.236 & 1.000 \\
		Similarity + NER & 0.800 & 4.015 & 0.902 & 0.236 & 0.233 & 1.000 \\
		Similarity + Topics + NER & 0.800 & 3.969 & 0.902 & 0.236 & 0.233 & 1.000 \\
		Random & 0.200 & 20.000 & 0.130 & 0.720 & 0.080 & 0.800 \\
		Random & 0.400 & 40.000 & 0.110 & 0.540 & 0.060 & 0.600 \\
		Random & 0.600 & 60.000 & 0.090 & 0.360 & 0.040 & 0.400 \\
		Random & 0.800 & 80.000 & 0.070 & 0.180 & 0.020 & 0.200 \\
		\bottomrule
	\end{tabular}
\end{table}

\begin{table}[!htbp]
	\centering
	\caption{Comparison of deduplication and filtering strategies on webfaq for RecursiveTokenChunker (400, 0).}
	\label{tab:webfaq_RecursiveTokenChunker_400_0_main}
	\begin{tabular}{llrrrrr}
		\toprule
		Method & Threshold & Reduction (\%) & Precision & Recall & IoU & Oracle \\
		\midrule
		No filtering & - & 0.000 & 0.879 & 0.394 & 0.383 & 0.579 \\
		ExactNorm & - & 0.488 & 0.879 & 0.396 & 0.385 & 0.579 \\
		MinHash-LSH & 0.600 & 1.172 & 0.879 & 0.397 & 0.387 & 0.578 \\
		MinHash-LSH & 0.700 & 1.074 & 0.884 & 0.398 & 0.388 & 0.578 \\
		MinHash-LSH & 0.800 & 0.879 & 0.879 & 0.397 & 0.387 & 0.579 \\
		NER Exact & - & 7.129 & 0.882 & 0.408 & 0.396 & 0.999 \\
		NER Half & - & 26.855 & 0.803 & 0.369 & 0.350 & 0.999 \\
		Similarity & 0.800 & 13.770 & 0.867 & 0.399 & 0.385 & 1.000 \\
		Similarity + Topics & 0.800 & 13.574 & 0.863 & 0.397 & 0.384 & 1.000 \\
		Similarity + NER & 0.800 & 8.105 & 0.882 & 0.403 & 0.392 & 0.999 \\
		Similarity + Topics + NER & 0.800 & 8.008 & 0.882 & 0.403 & 0.392 & 0.999 \\
		Random & 0.200 & 20.000 & 0.130 & 0.720 & 0.080 & 0.800 \\
		Random & 0.400 & 40.000 & 0.110 & 0.540 & 0.060 & 0.600 \\
		Random & 0.600 & 60.000 & 0.090 & 0.359 & 0.040 & 0.399 \\
		Random & 0.800 & 80.000 & 0.070 & 0.179 & 0.020 & 0.199 \\
		\bottomrule
	\end{tabular}
\end{table}

\begin{table}[!htbp]
	\centering
	\caption{Comparison of deduplication and filtering strategies on webfaq for RecursiveTokenChunker (400, 200).}
	\label{tab:webfaq_RecursiveTokenChunker_400_200_main}
	\begin{tabular}{llrrrrr}
		\toprule
		Method & Threshold & Reduction (\%) & Precision & Recall & IoU & Oracle \\
		\midrule
		No filtering & - & 0.000 & 0.889 & 0.395 & 0.386 & 0.588 \\
		ExactNorm & - & 0.540 & 0.889 & 0.397 & 0.388 & 0.588 \\
		MinHash-LSH & 0.600 & 1.928 & 0.889 & 0.398 & 0.389 & 0.588 \\
		MinHash-LSH & 0.700 & 1.696 & 0.889 & 0.398 & 0.389 & 0.588 \\
		MinHash-LSH & 0.800 & 1.234 & 0.889 & 0.398 & 0.389 & 0.588 \\
		NER Exact & - & 6.631 & 0.895 & 0.407 & 0.399 & 1.000 \\
		NER Half & - & 39.167 & 0.830 & 0.390 & 0.372 & 1.000 \\
		Similarity & 0.800 & 27.602 & 0.873 & 0.417 & 0.407 & 1.000 \\
		Similarity + Topics & 0.800 & 26.831 & 0.877 & 0.417 & 0.407 & 1.000 \\
		Similarity + NER & 0.800 & 20.586 & 0.891 & 0.417 & 0.407 & 1.000 \\
		Similarity + Topics + NER & 0.800 & 20.200 & 0.891 & 0.417 & 0.407 & 1.000 \\
		Random & 0.200 & 20.000 & 0.130 & 0.720 & 0.080 & 0.800 \\
		Random & 0.400 & 40.000 & 0.110 & 0.540 & 0.060 & 0.600 \\
		Random & 0.600 & 60.000 & 0.090 & 0.359 & 0.040 & 0.399 \\
		Random & 0.800 & 80.000 & 0.070 & 0.180 & 0.020 & 0.200 \\
		\bottomrule
	\end{tabular}
\end{table}

\begin{table}[!htbp]
	\centering
	\caption{Comparison of deduplication and filtering strategies on webfaq for RecursiveTokenChunker (800, 400).}
	\label{tab:webfaq_RecursiveTokenChunker_800_400_main}
	\begin{tabular}{llrrrrr}
		\toprule
		Method & Threshold & Reduction (\%) & Precision & Recall & IoU & Oracle \\
		\midrule
		No filtering & - & 0.000 & 0.851 & 0.634 & 0.591 & 0.858 \\
		ExactNorm & - & 0.000 & 0.851 & 0.634 & 0.591 & 0.858 \\
		MinHash-LSH & 0.600 & 0.743 & 0.851 & 0.636 & 0.594 & 0.858 \\
		MinHash-LSH & 0.700 & 0.446 & 0.851 & 0.636 & 0.594 & 0.858 \\
		MinHash-LSH & 0.800 & 0.446 & 0.851 & 0.636 & 0.594 & 0.858 \\
		NER Exact & - & 1.337 & 0.854 & 0.641 & 0.598 & 1.000 \\
		NER Half & - & 38.187 & 0.760 & 0.613 & 0.529 & 1.000 \\
		Similarity & 0.800 & 38.039 & 0.767 & 0.646 & 0.563 & 1.000 \\
		Similarity + Topics & 0.800 & 37.593 & 0.767 & 0.646 & 0.563 & 1.000 \\
		Similarity + NER & 0.800 & 33.581 & 0.792 & 0.656 & 0.583 & 1.000 \\
		Similarity + Topics + NER & 0.800 & 32.987 & 0.795 & 0.657 & 0.586 & 1.000 \\
		Random & 0.200 & 20.000 & 0.130 & 0.719 & 0.080 & 0.799 \\
		Random & 0.400 & 40.000 & 0.110 & 0.539 & 0.060 & 0.599 \\
		Random & 0.600 & 60.000 & 0.090 & 0.360 & 0.040 & 0.400 \\
		Random & 0.800 & 80.000 & 0.070 & 0.179 & 0.020 & 0.199 \\
		\bottomrule
	\end{tabular}
\end{table}

\end{document}